\documentclass[10pt]{article} 
\usepackage[preprint]{tmlr}

\usepackage{amsmath,amssymb,amsfonts}
\usepackage{graphicx}
\usepackage{booktabs}
\usepackage{caption}
\usepackage{placeins}
\usepackage{subcaption}
\usepackage{hyperref}
\usepackage{xurl} 
\usepackage{tabularx}
\usepackage{tikz}
\usetikzlibrary{arrows.meta,positioning}

\newcommand{\Wq}{W_q}
\newcommand{\Wk}{W_k}
\newcommand{\Wv}{W_v}
\newcommand{\Wo}{W_o}
\newcommand{\Wqkv}{W_{\text{qkv}}}
\newcommand{\Wgate}{W_{\text{gate}}}
\newcommand{\Wup}{W_{\text{up}}}
\newcommand{\Wdown}{W_{\text{down}}}
\newcommand{\lamwd}{\lambda_{\text{wd}}}

\title{A Two-Parameter Weibull Framework for Diagnosing\\Transformer Weight Distributions}
\hypersetup{hidelinks,
  pdftitle={A Two-Parameter Weibull Framework for Diagnosing Transformer Weight Distributions},
  pdfauthor={Tiexin Ding}}

\author{\name Tiexin Ding\thanks{Large language model tools were used to assist with manuscript revision and figure preparation. The author is responsible for the final content.} \email tiexinding@gmail.com \\
  \addr Beijing Azure River Tech Co., Ltd.}

\def\month{11}
\def\year{2026}
\def\openreview{\url{https://openreview.net/forum?id=j7qGffCwDa}}

\begin{document}
\maketitle

\begin{abstract}
Transformer checkpoints contain billions of weights, yet how their element-wise distributions vary across components, models, and training is hard to characterize. We fit the magnitude distribution of each analyzed weight matrix with a two-parameter Weibull reference model under a fixed middle-80\% protocol. For i.i.d.\ Gaussian initialization, the corresponding half-Normal target has protocol-fitted shape $k_0 \approx 1.20$. Matched comparisons with gamma and lognormal favor Weibull for most component groups; support is less decisive for separately stored MHA query/key projections, and no tested single family cleanly describes matrices carrying a secondary weight population.
Across 12 model entries spanning seven families, terminal FFN shape medians fall in a narrow band, $k \in [1.186, 1.204]$ (CV $= 0.51\%$), whereas query/key projections show architecture-associated departures. This stability is macroscopic only: in the five Pythia models whose released step-0 checkpoints permit a paired comparison, FFN and attention-output matrices remain near the reference shape despite large displacement and reorientation, so terminal shape similarity is not explained by parameter proximity alone. The fitted shape $k$ is read as a reference-relative central-body shape indicator and the scale $\lambda$ as a magnitude coordinate; neither is presented as a predictor of downstream performance. In a Qwen case study, the screen localizes a shallow-FFN secondary small-magnitude population, and direct raw-weight analysis of a flagged block traces much of that population to coherently suppressed intermediate channels. The method requires only released checkpoints and complements behavioral and structural analyses. The code and reference database are available at\newline
\url{https://github.com/tiexinding/NPM-Weibull-public}.
\end{abstract}

\section{Introduction}

\subsection{Problem and Motivation}

Understanding what happens during transformer training requires quantitative tools for interrogating learned weight distributions. Existing approaches operate on complementary observables: WeightWatcher \citep{martin2019traditional} and HT-SR \citep{martin2020heavy} analyze eigenvalue spectra; AlphaDecay \citep{he2025alphadecay} assigns module-wise weight decay from heavy-tailed spectral metrics; massive activation analysis \citep{sun2024massive} measures activation magnitudes. None of these methods directly characterizes the element-wise distribution of weight magnitudes $|W_{ij}|$ --- the finest-grained observable of the stored weights themselves.

One reason this gap matters is resolution: a statistic reported at an aggregate level --- spectral or element-wise alike --- averages across component types. If different functional components within the same model, say FFN modules versus attention projections, follow systematically different distributions, an aggregate summary averages the distinction away. We therefore fit each analyzed projection matrix separately, and read the result against an analytically known initialization reference.

The reference family is selected empirically rather than assumed: among the two-parameter candidates in common use for positive-valued data --- Weibull \citep{weibull1951statistical}, gamma, and lognormal --- a matched-protocol comparison favors Weibull for most component groups, with the exceptions flagged and reported alongside (Appendix~\ref{sec:appendix-dist-compare}). Each matrix's $|W_{ij}|$ is fitted over the middle 80\% of magnitudes (the protocol is specified in Section~\ref{sec:theory}), yielding one pair per matrix: a shape parameter $k$ quantifying tailedness and a scale parameter $\lambda$ quantifying magnitude. The initialization reference is analytic --- i.i.d.\ Gaussian weights give $|w| \sim \text{HalfNormal}$, which under the same protocol fits to $k_0 \approx 1.20$ --- so departures from $k_0$ can be read as training-induced, conditional on the matched protocol and i.i.d.\ Gaussian initialization.

These ingredients are verified before use: the anchor directly on Pythia step-0 checkpoints (within $0.13\%$ of the analytic value), the family choice by body fit and held-out tail prediction, and the protocol dependence of the read-out across trim choices (Appendices~\ref{sec:appendix-init} and \ref{sec:appendix-dist-compare}, Section~\ref{sec:bodytail}).

The two fitted parameters are read in a fixed order: fit adequacy is checked first; for an adequately fitted matrix, $k$ is read as a calibrated reference screen and $\lambda$ as its fitted magnitude --- cross-entry $\lambda$ comparisons additionally require matched protocol, component type, initialization recipe and training configuration (Section~\ref{sec:limitations}); if $k$ falls outside the reference band, $\lambda$ is not interpreted until the matrix is disaggregated and inspected. The claim category of this paper is state diagnostics rather than performance prediction: the framework flags and localizes structural atypicality relative to a calibrated reference population.

Read across the cohort, these coordinates reveal heterogeneous, depth-dependent evolution: different layers of the same model specialize at different rates, and within each layer, different projection matrices --- $\Wq$, $\Wk$, $\Wo$, the FFN matrices --- evolve along different trajectories (Sections~\ref{sec:ffn} and \ref{sec:qk}).

\begin{figure}[t]
\centering
\includegraphics[width=0.95\linewidth]{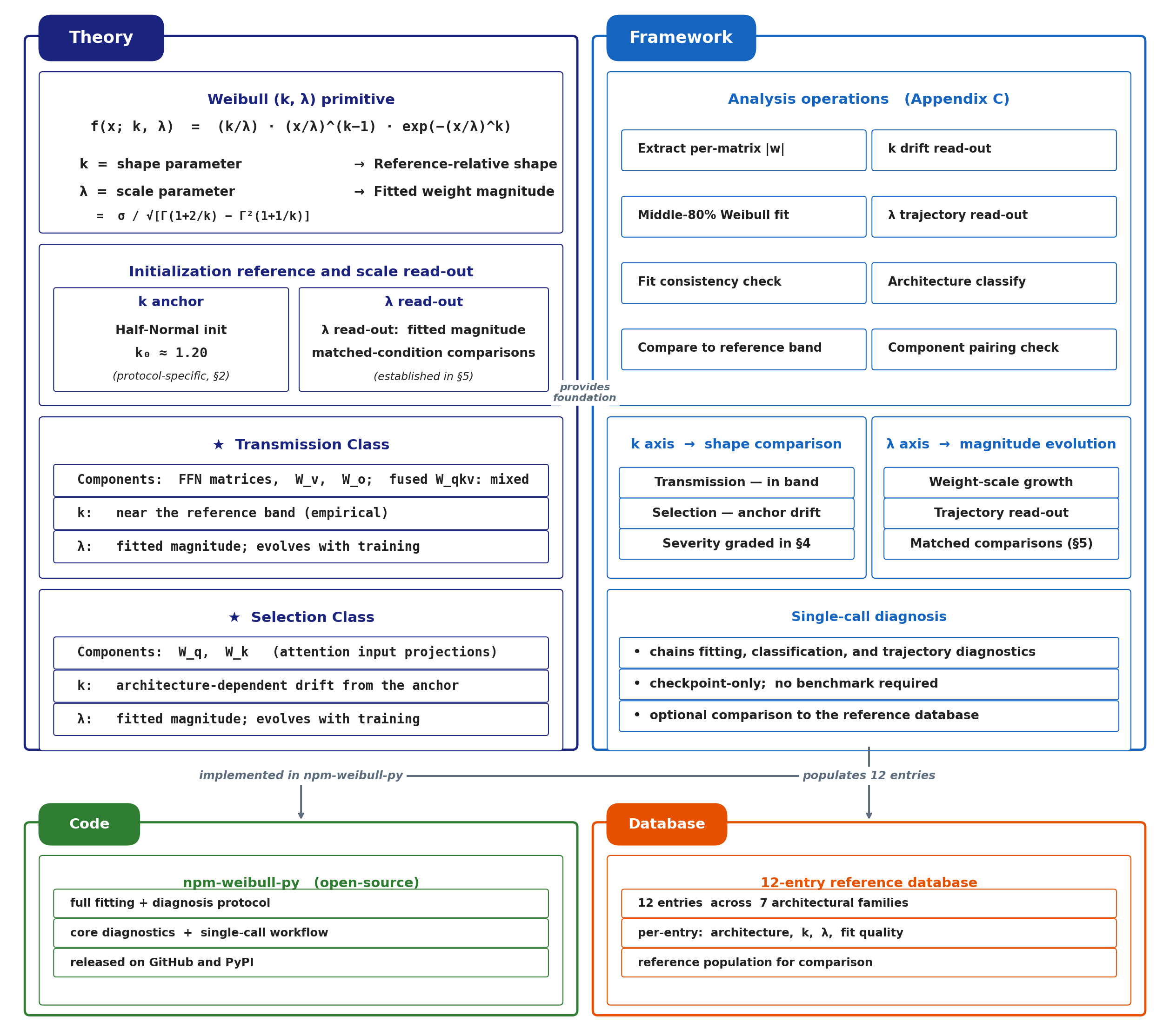}
\caption{\textbf{Diagnostic framework architecture.} The framework links four components: theory, diagnostic read-outs, software, and reference data. The theory module defines the two-parameter Weibull fit and the protocol-specific Half-Normal initialization reference, $k_0 \approx 1.20$. Component roles are assigned from the architecture before fitting. The framework module checks fit adequacy and then reads $k$ as reference-relative shape and $\lambda$ as fitted magnitude under matched comparisons; poor fits or marked departures prompt component-level localization and raw-weight inspection. The accompanying library implements the fitting and read-out workflow, while the 12-entry database (7 architectural families) supplies an optional cohort reference.}
\label{fig:f165}
\end{figure}

Together, the analytic reference, the fitting protocol with its fixed reading order, the verification battery, and the released tools and reference database assemble into the two-parameter Weibull diagnostic framework of the paper's title. Figure~\ref{fig:f165} summarizes this architecture; the remainder of the paper develops each component and then demonstrates the framework in use.

\subsection{Contributions}

We make three contributions. (i) A \emph{diagnostic framework}: per-matrix two-parameter Weibull fits read against an analytically specified initialization anchor, with the anti-interference property this construction confers (Section~\ref{sec:framework}). (ii) \emph{Cross-family empirical evidence}: across 12 model entries from 7 architectural families --- diverse in scale, activation, and normalization --- the FFN Transmission band (Section~\ref{sec:ffn}) and the architecture-associated Selection drift (Section~\ref{sec:qk}) are reproduced with no per-entry tuning. (iii) \emph{Open-source tools}: \texttt{npm-weibull-py v0.4} (seven exported core diagnostics; Appendix~\ref{sec:appendix-api}) and the companion benchmark database \texttt{DATABASE\_v9\_1} (12 entries). Beyond the measurements themselves, the paper demonstrates one application example of the framework: screening a released cohort, localizing a flagged deviation, and verifying the associated channel-level structure directly in the raw weights (Section~\ref{sec:within-entry-qwen}, Appendix~\ref{sec:appendix:qwen-diagnostic:channels}). The computation operates on released weights alone and requires neither forward passes nor eigendecomposition.

\subsection{Related Work}
\label{sec:related}

\paragraph{Element-wise weight distributions.} The question of what distribution
trained weights follow is old: \citet{bellido1993normal} asked whether
backpropagation-trained networks have normally distributed weights. Modern
treatments engage the trained weight distribution for specific purposes:
topology-based diagnostics whose read-outs are governed by weight variance and
the concentration of large weights \citep{girrbach2023persistence},
generalized-gamma fits for quantizer design \citep{kim2018generalized}, and
generalized-Gaussian modeling for initialization and training
\citep{wu2025generalized}. More recently, \citet{si2025gaussian} argue that
the weights of large foundation models remain approximately Gaussian throughout
training, and \citet{fortuin2022bayesian} measure element-wise weight marginals
of SGD-trained networks and report heavy-tailed marginals in fully connected
architectures --- evidence that the marginal is an informative, structured
object. Our work builds on this line, adding calibration --- a
protocol-fixed estimand, an analytic initialization anchor, and a cross-family
reference band; on our corpus this provides a component-resolved comparison to
the half-Normal initialization reference: in-band
Transmission matrices stay within about $1\%$ of the
(Gaussian-magnitude) anchor, while separately stored MHA query/key projections
depart from it measurably (Section~\ref{sec:qk}).

\paragraph{Heavy-tail emergence.} Theoretical analyses of stochastic optimization show that training can move
parameter distributions away from Gaussian: in specific SGD models the iterates
converge to heavy-tailed stationary distributions
\citep{gurbuzbalaban2021heavy}, and related analyses report heavy-tailed
coordinate-wise behaviour in network parameters \citep{barsbey2021heavy}. This literature predicts the
departure; the class-dependent \emph{invariance} documented here may offer a
further constraint for such accounts.

\paragraph{Spectral diagnostics.} WeightWatcher \citep{martin2019traditional}
and HT-SR \citep{martin2020heavy} analyze eigenvalue spectra of weight
matrices; the program extends to data-free quality prediction at scale
\citep{martin2021predicting} and to per-layer, in-training diagnostics
\citep{zhou2023tempbalance}; AlphaDecay \citep{he2025alphadecay} assigns module-wise weight decay from heavy-tailed spectral metrics. Random-matrix analyses find the singular-value bulk of
trained matrices close to random-matrix universality, with deviations
concentrated at the spectral edges \citep{thamm2022random}, and shape-type and
scale-type statistics of the same weights have been shown to succeed on
complementary data partitions \citep{martin2021simpson}. We read these
statistics as complementary rather than competitive: the element-wise shape is
a bulk property whose behaviour under training is component-dependent, and it
admits the initialization reference introduced above --- a reference specific
to the element-wise statistic, since under i.i.d.\ Gaussian initialization the
spectrum follows the Marchenko--Pastur law, whose bounded support provides no
power-law tail to anchor a shape exponent to. We do not claim that
spectral quantities cannot capture the phenomena documented in this paper;
Appendix~\ref{sec:appendix:qwen-diagnostic:channels} demonstrates what the element-wise route
provides --- verification at the granularity of named channel indices in the
raw weights.

\paragraph{Distributional assumptions in practice.} Downstream methods already
rest on assumptions about the element-wise marginal. The NF4 quantization
format is derived as information-theoretically optimal for normally
distributed weights \citep{dettmers2023qlora}; subsequent measurement across
thirty networks corrected this assumption toward heavier-tailed families
\citep{dotzel2024students}. Assumptions about the shape of the weight marginal thus already enter
practical decisions, which motivates a calibrated per-matrix diagnostic of it.

Adjacent weight- and activation-level phenomena --- super-weights
\citep{yu2024superweight}, massive activations \citep{sun2024massive},
outlier activations under adaptive optimizers \citep{kaul2025attention}, and
dead neurons \citep{voita2023dead} --- have points of contact with our
observations, noted where they arise
(Sections~\ref{sec:superweight} and \ref{sec:within-entry-qwen}, Appendix~\ref{sec:appendix:qwen-diagnostic}). To our
knowledge, no prior work uses the Weibull shape parameter of the element-wise
$|W|$ distribution as a per-matrix training diagnostic with an analytically specified,
numerically evaluated initialization anchor.

\begin{figure}[t]
\centering
\begin{tikzpicture}[
  font=\footnotesize,
  box/.style={rectangle, rounded corners=2pt, draw=black!65, align=center, inner sep=4pt, fill=black!3},
  bandlabel/.style={font=\footnotesize\itshape, anchor=west},
  arr/.style={->, >=stealth, black!65, semithick}
]
\node[bandlabel] at (-6.8,3.5) {Main text --- what the framework does};

\node[box] (theory) at (-3.7,2.55) {\textbf{Theory (\S\ref{sec:theory})}\\ per-matrix $|W| \to$ Weibull $(k,\lambda)$\\ analytic initialization anchor $k_0$};
\node[box] (classes) at (3.1,2.55) {\textbf{Architectural roles (\S\ref{sec:classes})}\\ Transmission (FFN $+$ $\Wv$/$\Wo$) $\cdot$ Selection ($\Wq$/$\Wk$)\\ $\cdot$ fused $\Wqkv$: mixed};
\draw[arr] (theory) -- (classes);

\node[box] (gate) at (0,1.15) {\textbf{Fit-adequacy gate}: read in fixed order, fit $\to$ $k$ $\to$ $\lambda$};
\draw[arr] (classes.south) -- (gate.north);

\node[box, text width=4.1cm] (s3) at (-4.65,-0.25) {\textbf{$k$ $\times$ Transmission (\S\ref{sec:ffn})}\\ FFN band, robustness, body--tail; documented $\Wv$/$\Wo$ exceptions};
\node[box, text width=4.1cm] (s4) at (0,-0.25) {\textbf{$k$ $\times$ Selection (\S\ref{sec:qk})}\\ departures: depth \&\\ architecture associations;\\ candidate mechanisms};
\node[box, text width=4.1cm] (s5) at (4.65,-0.25) {\textbf{$\lambda$ per stored kind (\S\ref{sec:lambda})}\\ separate scale coordinate,\\ matched conditions;\\ trajectories under AdamW};
\draw[arr] (gate.south) -- (s3.north);
\draw[arr] (gate.south) -- (s4.north);
\draw[arr] (gate.south) -- (s5.north);

\node[box] (case) at (0,-1.95) {\textbf{Application: Qwen case (\S\ref{sec:within-entry-qwen})}\\ screen $\to$ localize $\to$ raw-weight verify};
\draw[arr] (s3.south) -- (case.north west);
\draw[arr] (s5.south) -- (case.north east);

\node[bandlabel] at (-6.8,-2.75) {Appendices --- why it is justified};
\node[box, text width=13.1cm, align=left] (app) at (0,-4.25) {%
\renewcommand{\arraystretch}{1.25}%
\begin{tabular}{@{}p{6.5cm}p{6.0cm}@{}}
anchor derivation; protocol sensitivity (\ref{sec:appendix-init}) $\to$ \S\ref{sec:theory}, \S\ref{sec:bodytail}, \S\ref{sec:lambda-init} & reference-family adjudication (\ref{sec:appendix-dist-compare}) $\to$ \S\ref{sec:theory}\\
initialization recipes (\ref{sec:appendix-init-pythia}) $\to$ \S\ref{sec:theory}, \S\ref{sec:lambda} & candidate mechanisms (\ref{sec:driving-forces}--\ref{sec:appendix-khead}) $\to$ \S\ref{sec:classes}, \S\ref{sec:qk}\\
displacement measurement (\ref{sec:appendix-displacement}) $\to$ \S\ref{sec:ffn} & experiment summary (\ref{sec:appendix-experiments}) $\to$ \S\ref{sec:ffn}--\S\ref{sec:lambda}\\
software \& database (\ref{sec:appendix-api}): workflow,\newline cohort reference & full diagnostic chain (\ref{sec:appendix:qwen-diagnostic}) $\to$ \S\ref{sec:within-entry-qwen}\\
\end{tabular}};
\draw[arr] (app.north) -- ++(0,0.45);
\end{tikzpicture}
\caption{\textbf{Paper overview and evidence map.} Figure~\ref{fig:f165} charts the components of the diagnostic framework; this figure charts the units of the paper. The main text proceeds from a protocol-specific reference through a fit-adequacy gate to class-conditional $k$ diagnostics, matched-condition $\lambda$ read-outs, and the Qwen application chain (\S\ref{sec:within-entry-qwen}). Appendix~\ref{sec:appendix-theory} validates the anchor, protocol, and reference-family choice, and develops candidate mechanisms; Appendices \ref{sec:appendix-experiments}--\ref{sec:appendix-api} document experiments, software, and data; Appendix~\ref{sec:appendix:qwen-diagnostic} completes the case study at raw-weight granularity.}
\label{fig:reading-map}
\end{figure}

\section{Framework}
\label{sec:framework}

This section specifies the framework as it is used --- the initialization anchor, the fitting protocol, the cohort it is read on, and the two component classes; the derivations, numerical evaluations, and verification experiments that justify these choices are collected in Appendix~\ref{sec:appendix-theory}. Figure~\ref{fig:reading-map} charts how the units of the paper fit together.

\subsection{Theoretical Basis}
\label{sec:theory}

\paragraph{Initialization anchor.} The framework begins with an analytical result at initialization. The analyzed weight matrices of every cohort entry are initialized with i.i.d.\ Gaussian weight elements, $w \sim \mathcal{N}(0, \sigma_{\text{init}}^2)$, under per-component recipes (Appendix~\ref{sec:appendix-init-pythia}); the anchor below is stated for this initialization family. The absolute values $|w|$ therefore follow a half-Normal distribution, which is not a closed-form special case of Weibull. Fitting $|w|$ to a Weibull distribution via middle-80\% probability-plot least-squares --- the protocol used throughout this work --- gives shape parameter $k_0 \approx 1.20$ and scale parameter $\lambda_0 \approx 0.8875\,\sigma_{\text{init}}$, analytically specified by the protocol and numerically evaluated via deterministic special-function integrals (Appendix~\ref{sec:appendix-init}). The shape $k_0$ is independent of $\sigma_{\text{init}}$ and therefore invariant to the initialization scale within the i.i.d.\ Gaussian family; the scale $\lambda_0$ is proportional to $\sigma_{\text{init}}$ and varies by initialization recipe and by component (Appendix~\ref{sec:appendix-init-pythia}). We treat $k_0 \approx 1.20$ as the protocol-specific initialization reference: departures from it indicate training-induced structural change. Departures of $\lambda$ from $\lambda_0$ reflect both training-induced scaling and the per-component initialization split discussed in Section~\ref{sec:lambda}.

\paragraph{Fitting protocol.} To measure $k$ in a trained model, we extract each analyzed weight matrix separately, flatten its absolute entries, and fit one Weibull distribution per matrix using least-squares on the Weibull probability plot. A critical design choice is the \textbf{middle 80\% trim}: before fitting, we discard the smallest and largest 10\% of $|W_{ij}|$ values.

The trim fraction is motivated by the sampling-noise profile of the probability-plot coordinate rather than chosen ad hoc: the variance of the rank coordinate diverges at both endpoints and reaches its minimum near $p \approx 0.80$ (Appendix~\ref{sec:appendix-init}, Figure~\ref{fig:f167}). Empirically, re-fitting on the full data (no trim) shifts the fitted $k$ down by approximately $4\%$, a shift that matches the analytic movement of the anchor itself between the two protocols (Section~\ref{sec:bodytail}). The choice of Weibull over alternative two-parameter families is examined in Appendix~\ref{sec:appendix-dist-compare}, which supports Weibull in every component group while reporting separately stored MHA query/key projections and flagged secondary-population matrices as the exceptions. The anchor $k_0(P)$ is analytically specified for any trim fraction (Table~\ref{tab:k0-protocol}), and the framework reading $d_P = k_P - k_0(P)$ is protocol-stable (Appendix~\ref{sec:appendix-init}).

This gives the framework its \textbf{anti-interference property}: the trim makes fitted $k$ robust to isolated outliers and the heavy-tail phenomena that dominate the extreme percentiles of trained weight distributions; robustness to the checked precision and centering perturbations is established empirically in Appendix~\ref{sec:appendix-init}.

\paragraph{Cohort and experimental setup.} All measurements in this paper are taken on publicly released checkpoints; Table~\ref{tab:cohort-setup} states the cohort in full --- model identity, training corpus, and the training stage at which each entry is analyzed --- before any result is reported. Two points about alignment are worth stating explicitly. First, the cohort is aligned at \emph{end of pretraining as released}: for every non-Pythia entry we analyze the final pretrained (base) checkpoint, excluding instruction-tuned or chat variants. For Pythia, step 143{,}000 is likewise the final released checkpoint --- one epoch over the Pile, about 300B tokens --- and is therefore a terminal, not an early, state. Second, this alignment is by release stage rather than by a fixed token count or compute budget: token counts differ by two orders of magnitude across the cohort (300B to 36T), and for several entries the corpus is not publicly documented. The entries are thus comparable as \emph{released artifacts}, which is the setting the framework is designed for --- it reads the weights a practitioner actually receives.

\begin{table}[t]
\centering
\small
\caption{\textbf{Cohort and experimental setup: 12 entries across 7 architectural families.} ``final base\,'' denotes the final pretrained (base) checkpoint as released, excluding instruction-tuned and chat variants. Corpus and token counts are as documented by each release; ``not disclosed'' records that the publisher does not document the corpus, and is stated as a fact about the release rather than an inference. Hugging Face repositories for every entry are listed in Appendix~\ref{sec:appendix-api}; optimizer settings ($\eta_{\text{peak}}$, $\lamwd$) used in the $\lambda$ analysis are given in Table~\ref{tab:lambda-crossfamily}. $^{\dagger}$Pythia releases every checkpoint from step 0 to step 143{,}000; step 143{,}000 is the final one, and trajectory analyses additionally use 14 log-spaced revisions of the same runs. $^{\S}$OLMo-2 pretraining runs in two stages (OLMo-Mix-1124, then Dolmino-Mix-1124).}
\label{tab:cohort-setup}
\begin{tabular}{lccll}
\toprule
Entry & Layers & Tokens & Training corpus & Checkpoint \\
\midrule
Pythia-70m   & 6  & 300B & the Pile & step 143{,}000$^{\dagger}$ \\
Pythia-160m  & 12 & 300B & the Pile & step 143{,}000$^{\dagger}$ \\
Pythia-410m  & 24 & 300B & the Pile & step 143{,}000$^{\dagger}$ \\
Pythia-1B    & 16 & 300B & the Pile & step 143{,}000$^{\dagger}$ \\
Pythia-6.9B  & 32 & 300B & the Pile & step 143{,}000$^{\dagger}$ \\
\midrule
OLMo-1-7B    & 32 & 2.5T & Dolma & final base \\
OLMo-2-7B    & 32 & $\sim$4T & OLMo-Mix $\to$ Dolmino-Mix$^{\S}$ & final base \\
LLaMA-3-8B   & 32 & $>$15T & not disclosed & final base \\
Mistral-7B   & 32 & not disclosed & not disclosed & final base \\
Qwen2.5-7B   & 28 & up to 18T & not disclosed & final base \\
Qwen2.5-14B  & 48 & up to 18T & not disclosed & final base \\
Qwen3-8B     & 36 & $\sim$36T & not disclosed & final base \\
\bottomrule
\end{tabular}

\end{table}

\paragraph{What $k$ measures.} The shape parameter $k$ of a Weibull distribution summarizes the shape of the fitted middle-80\% body of $|W|$: smaller $k$ corresponds to a more spread-out body with a heavier right shoulder, while larger $k$ corresponds to a narrower, more uniform body. The extreme percentiles are excluded by the protocol, so $k$ is not a tail exponent of the full distribution; where the fit is inadequate it should be read as an effective summary and anomaly score rather than a shape law. Because $k$ is dimensionless it is not tied to the scale of a matrix, so values are comparable across weight matrices of different shapes and sizes --- provided they are obtained under the same trim protocol and refer to the same component definition. Figure~\ref{fig:f47} illustrates, for Transmission-class components, how the $|W|$ distribution evolves through training: the body $k$ stays close to the initialization anchor while isolated elements detach in the extreme tail. The contrasting behaviour of the attention Q/K projections is reported in Section~\ref{sec:qk}. Together these motivate the two component classes --- \textbf{Transmission} (FFN $+$ $\Wv$/$\Wo$) and \textbf{Selection} (Q/K) --- developed in Section~\ref{sec:classes}.

\subsection{Two Component Classes}
\label{sec:classes}

Having established the initialization anchor and the measurement protocol, we now apply this diagnostic tool to trained transformer weights. Across 12 model entries spanning 7 architectural families and all training stages examined, the separately stored components organize into two classes with statistically distinct Weibull $k$ trajectories; fused storage is treated as mixed from the outset. We call these \textbf{Transmission} and \textbf{Selection}. Transmission and Selection are mnemonic labels. Class membership is defined architecturally: among the analyzed attention projections and FFN matrices, $\Wq$/$\Wk$ parameterize the softmax and the remaining ones act outside it. Fused $\Wqkv$, which stores query, key, and value together, is labelled \emph{mixed} in advance and is not treated as a clean member of either class. The rule can be stated before any statistic is computed. The labels name this architectural role; the distinct statistical behavior is an empirical finding, not a functional claim.

\begin{figure}[t]
\centering
\includegraphics[width=0.95\linewidth]{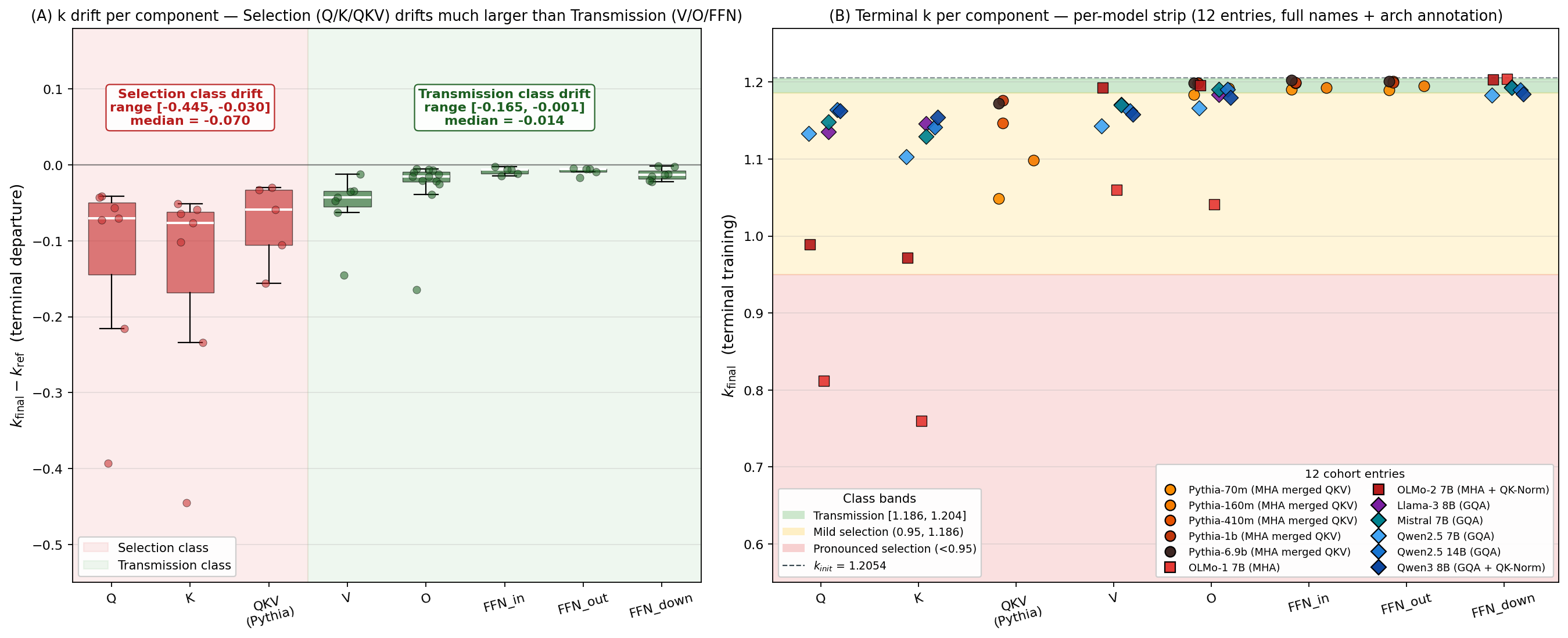}
\caption{\textbf{$k$ drift dichotomy across the 12-entry cohort.}
\textbf{(A)} Terminal departure of $k$ from the initialization reference, using
the measured step-0 $k$ for the Pythia entries and the analytic anchor
$k_0 = 1.2054$ otherwise (medians: Transmission $-0.014$; Selection $-0.070$).
\textbf{(B)} Terminal $k$ relative to the Transmission band $[1.186, 1.204]$
(green). Fused $\Wqkv$ is shown adjacent to Selection and treated as a mixed
case; the cascade FFN labels alias per Section~\ref{sec:classes}. The FFN
components cluster in the band, while $\Wv$ and $\Wo$ show documented exceptions
(notably the OLMo $\Wo$ points). Model family and training recipe are not
controlled across the architecture groups.}
\label{fig:f158}
\end{figure}

\paragraph{Transmission Class.} The Transmission Class consists of all FFN modules together with the attention value and output projections $\Wv$ and $\Wo$. We retain the architecture-specific FFN naming convention of each model family rather than imposing a uniform notation: for 3-matrix SwiGLU FFNs (LLaMA-3, Mistral, Qwen2.5/3, OLMo-1, OLMo-2), the gate projection $\Wgate$, the up projection $\Wup$, and the down projection $\Wdown$; for the 2-matrix GeLU FFNs of Pythia, the input projection $W_{\text{FFN\_in}}$ and the output projection $W_{\text{FFN\_out}}$. Cohort-wide figures (notably Figure~\ref{fig:f158}) adopt the generic cascade labels \texttt{FFN\_in}, \texttt{FFN\_out}, \texttt{FFN\_down} as an alias over both schemes. For these components, $k$ remains remarkably stable throughout training, close to the initialization anchor $k_0 \approx 1.20$ --- most tightly for the FFN modules, while $\Wv$/$\Wo$ show documented exceptions in the most strongly drifted MHA families (see the Selection paragraph below and Figure~\ref{fig:f158}). The scale parameter $\lambda$ varies by more than an order of magnitude across families and layers while $k$ does not follow it; we call this empirical stability the Transmission Class signature; the quantitative evidence is presented in Section~\ref{sec:ffn}.

The QK/OV circuit decomposition of prior work \citep{elhage2021circuits} motivates the mnemonic labels: $\Wq$/$\Wk$ form the circuit that computes attention patterns, while $\Wv$/$\Wo$ and the FFN modules carry the transformed content. We do not claim this decomposition explains the element-wise shape behaviour reported here. The body-tail decomposition (Section~\ref{sec:ffn}) shows that the Weibull body of Transmission Class components remains intact even as isolated super-weights emerge in the tail --- within this cohort, training-induced change in the fitted body shape is small for in-band FFN matrices, while scale, matrix direction, and extreme-tail outliers evolve separately (Section~\ref{sec:lambda}, Appendix~\ref{sec:appendix-displacement}).

\paragraph{Selection Class.} The Selection Class consists of the attention input projections $\Wq$ and $\Wk$. In contrast to Transmission components, $\Wq$ and $\Wk$ depart systematically from the initialization anchor $k_0 \approx 1.20$. Three candidate mechanisms may collectively contribute to this departure: functional necessity (sparse attention requires selective $\Wq, \Wk$), AdamW sign-descent dynamics, and softmax saturation feedback (no-op heads back-propagating extreme logits into $\Wk$); two further candidate forces (residual-stream coupling and training-budget accumulation) are discussed in Appendix~\ref{sec:driving-forces}. These candidates are not isolated by this cohort. Empirically $\Wk$ shows the larger departure of the Q/K pair in most entries (LLaMA-3 is the exception), and the Transmission-class $\Wv$ and $\Wo$ show intermediate departures in some families, co-occurring with the severity of Q/K departure in the same entry (Figure~\ref{fig:f158}; in OLMo-1-7B, where MHA Q/K departure is most severe, mid-layer $\Wo$ has median $k \approx 1.04$ rather than the Transmission band $\sim 1.20$). Severity and structure differ by architecture; quantitative evidence is presented in Section~\ref{sec:qk}.

\section{FFN Transmission Band}
\label{sec:ffn}

Section~\ref{sec:classes} identified the FFN modules and attention output projection $\Wo$ as the Transmission Class. This section documents their empirical Weibull signature.

\subsection{The $k$ Band}

\begin{figure}[!htbp]
\centering
\includegraphics[width=0.95\linewidth]{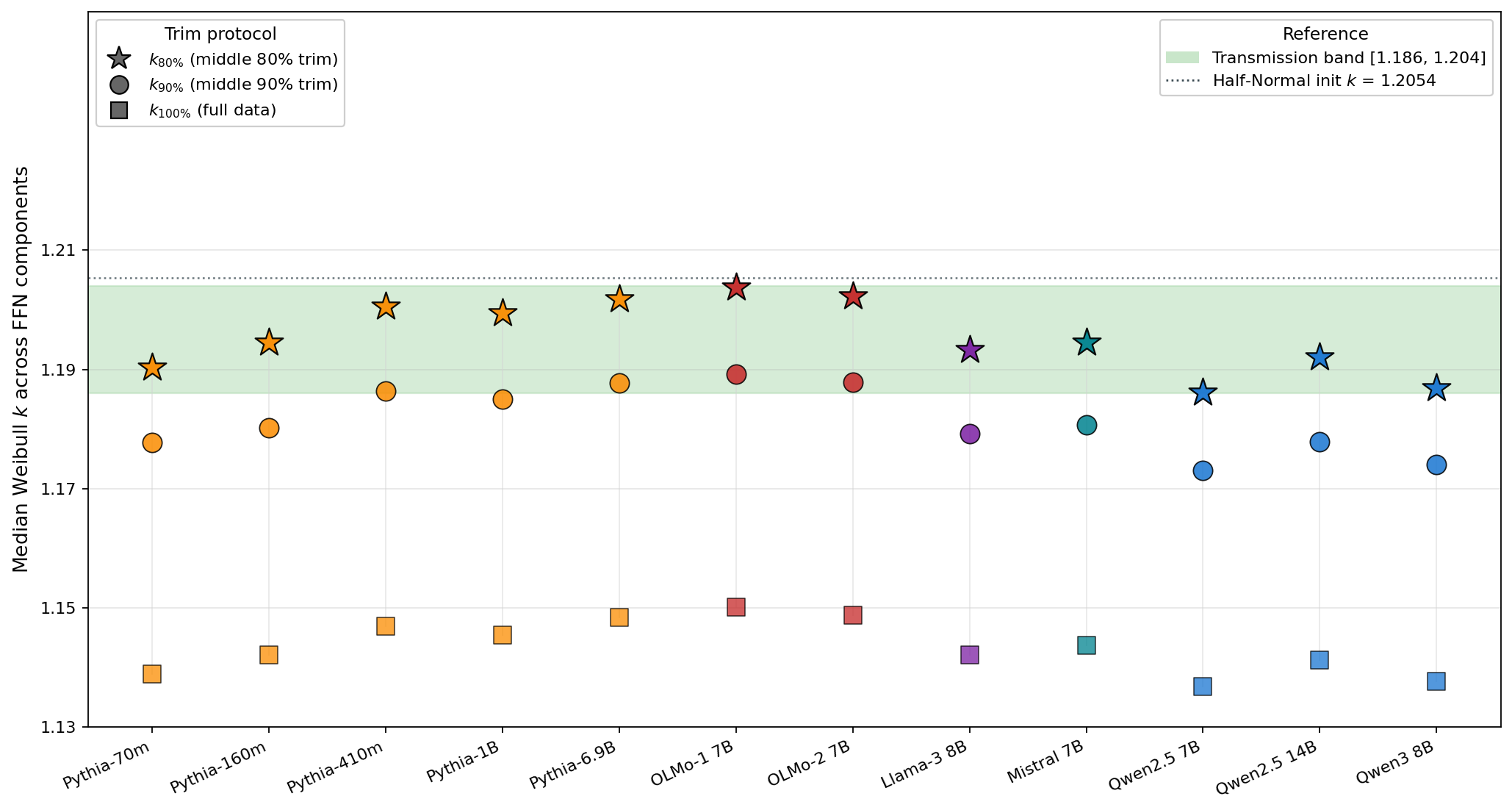}
\caption{\textbf{Body--tail ablation across the 12-entry cohort.} Three trim protocols ($k_{80\%}$, $k_{90\%}$, $k_{100\%}$) applied to FFN components across 12 entries spanning 7 architectural families. The middle 80\% protocol places 12/12 entries inside the Transmission band $[1.186, 1.204]$; full-data fit ($k_{100\%}$) places 0/12 inside; the body--tail gap $k_{80\%} - k_{100\%} = 0.0519 \pm 0.0017$ (CV $= 3.3\%$) matches the analytic anchor shift $k_0(80\%) - k_0(100\%) = 0.0543$ (Table~\ref{tab:k0-protocol}), identifying the gap as protocol geometry; the residual after anchoring is an order of magnitude smaller (Appendix~\ref{sec:appendix-init}).}
\label{fig:f160}
\end{figure}

Across 12 model entries spanning 7 architectural families --- Pythia, OLMo-1, OLMo-2, LLaMA-3, Mistral, Qwen2.5, Qwen3 --- across two orders of magnitude (70M--14B, factor of $200\times$) in parameter count, three initialization schemes, two activation patterns (GeLU 2-matrix in Pythia, SwiGLU 3-matrix in modern frontier models), and four norm placements (Pre-LN, Pre-LN $+$ RMSNorm, Peri-LN, QK-Norm), the FFN modules fit a Weibull distribution with median terminal $k \in [1.186, 1.204]$ (Figure~\ref{fig:f160}; per-component drift dichotomy in Figure~\ref{fig:f158}). The coefficient of variation is CV $= 0.51\%$ across 12 entries (family-level CV $= 0.53\%$ with each of the 7 families contributing one vote; leave-one-family-out range of the per-entry CV $[0.47\%, 0.57\%]$); excluding the pre-identified Regime-B shallow matrices, every FFN fit achieves $R^2 \geq 0.99$, and the $\Gamma$ closure check is evaluated on all 900 FFN fits (median first-moment relative error $4.0\%$, 90\% of fits within $4.4\%$, systematically positive; Appendix~\ref{sec:appendix-init}). Fit quality by component group is reported separately in Table~\ref{tab:fit-quality}: the FFN fits that do fall below $R^2 = 0.99$ are exactly those at the flagged Regime-B site, so a reduced $R^2$ co-occurs with the shape departure and is read as part of the signal the framework is designed to surface rather than as a failure of the protocol. The band is reproducible with no per-entry tuning: while the scale parameter $\lambda$ varies by nearly an order of magnitude across families, the shape $k$ stays within $[1.186, 1.204]$ regardless of activation pattern, normalization placement, or initialization scheme. This invariance is macroscopic only: over the same training runs, these matrices undergo large matrix-level displacement from initialization, with low cosine similarity between the flattened initial and terminal weights (Appendix~\ref{sec:appendix-displacement}).

\subsection{Robustness: Body--Tail Ablation}
\label{sec:bodytail}

The transmission band is established using the middle 80\% trim fitting protocol. The theoretical justification (sampling-noise divergence at distribution tails) is given in Section~\ref{sec:theory}. Re-fitting the same FFN components with three truncation protocols --- $k_{80\%}$, $k_{90\%}$, and $k_{100\%}$ --- across 12 entries gives:

\begin{table}[h]
\centering
\caption{Body--tail ablation across the 12-entry cohort. Raw fitted values are protocol-specific: the downward shift under wider windows matches the analytic anchor shift $k_0(P)$ (Table~\ref{tab:k0-protocol}), and band membership is defined at the 80\% protocol.}
\label{tab:bodytail}
\begin{tabular}{lcc}
\toprule
Protocol & Median $k$ & In band $[1.186, 1.204]$ \\
\midrule
$k_{80\%}$ & 1.195 & $\mathbf{12\,/\,12}$ \\
$k_{90\%}$ & 1.182 & 0 / 12 \\
$k_{100\%}$ & 1.143 & 0 / 12 \\
\bottomrule
\end{tabular}

\end{table}

The body--tail gap $k_{80\%} - k_{100\%} = 0.0519 \pm 0.0017$ is remarkably stable across all 12 entries spanning 7 families (CV $= 3.3\%$). Its origin is protocol geometry rather than the trained-model tails: the analytic half-Normal anchor moves by $k_0(80\%) - k_0(100\%) = 0.0543$ between the same two protocols (Table~\ref{tab:k0-protocol}), so a distribution with no training-induced tail at all exhibits nearly the same gap --- widening the window re-admits the endpoint regions where the reference family cannot match a half-Normal shape, and the fitted $k$ becomes a window-dependent compromise. The tight cross-family stability of the gap follows: it is a near-constant of the protocol, not a measure of each family's tail weight; net of this geometry, the remaining anchored discrepancy is an order of magnitude smaller (median anchored residual $d_{80\%} - d_{100\%} = -0.0021$; Appendix~\ref{sec:appendix-init}). The ablation is therefore a consistency requirement --- fitted values are read against the anchor of their own protocol --- rather than a tail measurement; the trained-model outliers themselves are the subject of Section~\ref{sec:superweight}.

\subsection{Cross-Family Super-Weight Observation}
\label{sec:superweight}

The body--tail ablation (Section~\ref{sec:bodytail}) establishes that the fitted body shape is read consistently across protocols. We now turn to the tail phenomenon that the trim excludes from the fit --- isolated heavy-tailed weight outliers, or \emph{super-weights}, observed in every entry for which we have per-block weight access (8 entries spanning 4 of the 7 families).

\begin{figure}[t]
\centering
\includegraphics[width=0.95\linewidth]{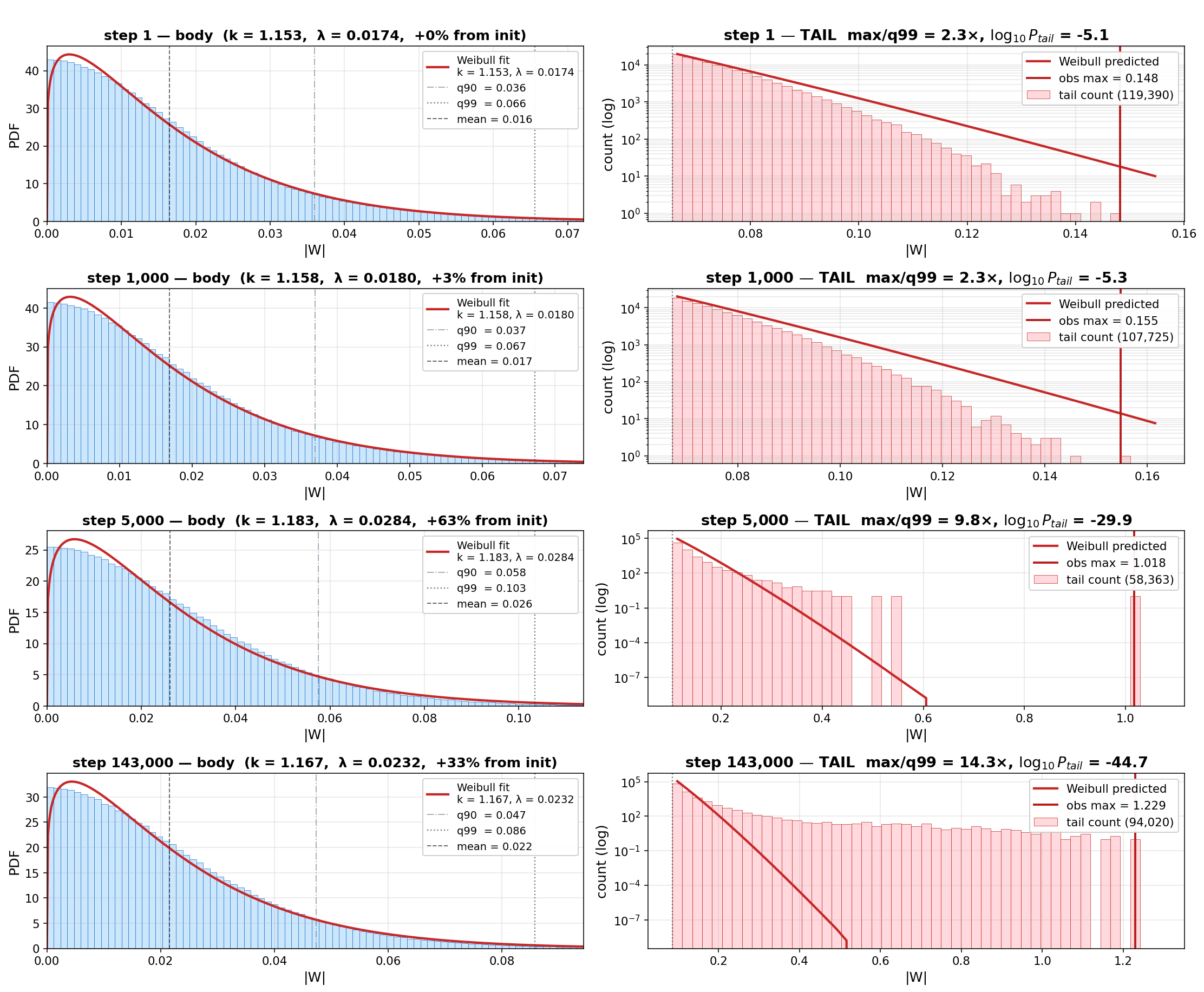}
\caption{\textbf{Pythia-70m $|W|$ Weibull body fit (left column) + tail evolution (right column) across 4 representative training steps (1, 1000, 5000, 143000) --- Transmission Class only ($\Wo$ $+$ FFN, $\Wqkv$ excluded).} The left column shows the middle-80\% Weibull fit on the Transmission Class bulk weight distribution ($n = 14{,}155{,}776$ samples per checkpoint). The shape parameter $k$ stays near initialization throughout training (aggregate-fit $k = 1.153 \to 1.158 \to 1.183 \to 1.167$ across the four steps shown) and the aggregate scale parameter $\lambda$ grows from $0.0174$ at step 1 to $0.0232$ at step 143000 ($+33\%$). Note that the aggregate-fit $k = 1.153$ at step 1 lies below the per-block anchor $k_0 \approx 1.20$ because pooling $\Wo + W_{\text{FFN\_in}} + W_{\text{FFN\_out}}$ mixes components with different per-component $\sigma_{\text{init}}$ (Appendix~\ref{sec:appendix-init-pythia} Recipe~A: $\sigma_{\text{in}}/\sigma_{\text{out}} = L/\sqrt{10}$); per-block fits at step 0 yield $k = 1.205 \pm 0.001$ across all kinds (Appendix~\ref{sec:appendix-init}, with the per-component initialization recipes in Appendix~\ref{sec:appendix-init-pythia}, Table~\ref{tab:lambda-init-verify}). The right column shows the tail region beyond q99. Bars are observed counts; the red line is the Weibull-predicted count using the body fit extrapolated. By step 5000, an isolated super-weight emerges at $|w| \approx 1.0$ (max/q99 $= 9.8\times$, $\log_{10} P_{\text{tail}} = -29.9$); at step 143000 the super-weight reaches $|w| = 1.2$ (max/q99 $= 14.3\times$, $\log_{10} P_{\text{tail}} = -44.7$). The body remains Weibull-conforming throughout; only a small number of extreme elements per matrix detach from the bulk. Isolated upper-tail outliers are observed across the eight entries with per-block weight access; see Table~\ref{tab:superweight} for per-block quantification and Section~\ref{sec:superweight} for discussion.}
\label{fig:f47}
\end{figure}

\paragraph{Super-weight emergence within a single training trajectory.} Figure~\ref{fig:f47} (right column) shows the tail evolution of Pythia-70m Transmission Class weights across four training checkpoints. At step 1, the right tail beyond q99 matches the Weibull body prediction within sampling noise. By step 5{,}000 an isolated outlier emerges at $|w| \approx 1.0$ ($\max / q99 = 9.8\times$, $\log_{10} P_{\text{tail}} = -29.9$). By step 143{,}000 this outlier reaches $|w| = 1.2$ ($\max / q99 = 14.3\times$, $\log_{10} P_{\text{tail}} = -44.7$). The body of the distribution remains Weibull-conforming throughout; only one --- or a small handful of --- extreme elements per matrix detach from the bulk.

\paragraph{Cross-family quantification.} Re-applying the same per-block tail diagnostic across our 8 measured terminal checkpoints (the 12-entry cohort restricted to checkpoints with per-block weight access; the remaining 4 entries enter the cohort only at the aggregate level used in Section~\ref{sec:ffn}) reveals that the Pythia-70m pattern generalizes: every one of these 8 entries contains isolated outliers (Table~\ref{tab:superweight}). The per-block max-to-q99 ratio has cohort medians of approximately $7$--$28\times$ (range $6.6\times$ at Pythia-160m to $27.7\times$ at OLMo-1-7B; see Table~\ref{tab:superweight}) and family-level maxima reaching $107\times$ (OLMo-1-7B). The per-block kurtosis distribution likewise concentrates near $\sim 4$--$10$ but exhibits per-family extreme values reaching $446$ (OLMo-1-7B) and $257$ (Pythia-410m).

\begin{table}[h]
\centering
\small
\caption{Per-block super-weight signatures across the 8 entries with per-block weight access --- Pythia (5 entries), OLMo-1, Qwen2.5-14B and Qwen3, i.e.\ 4 of the 7 architectural families; the full 12-entry cohort is reported in the FFN-band analysis in text (Section~\ref{sec:ffn}). ``$\max / q99$'' is the ratio of the largest $|w_{ij}|$ in a block to the 99th-percentile $|w_{ij}|$ of the same block. Kurtosis is the empirical excess kurtosis, computed from raw moments as $E[|W|^4]/E[|W|^2]^2 - 3$ under the zero-mean convention (as in Table~\ref{tab:moment-compare}). Median values report the typical block in each family; extreme values report the single most outlier-laden block.}
\label{tab:superweight}
\begin{tabular}{lcccc}
\toprule
Entry & \# blocks & max$/q99$ median & max$/q99$ extreme & kurtosis extreme \\
\midrule
Pythia-70m   & 6  & 7.2  & 15.7  & 196.9 \\
Pythia-160m  & 12 & 6.6  & 13.4  & 104.2 \\
Pythia-410m  & 24 & 8.0  & 19.6  & 257.1 \\
Pythia-1B    & 16 & 11.3 & 21.5  & 21.8 \\
Pythia-6.9B  & 32 & 8.6  & 31.4  & 27.7 \\
OLMo-1-7B    & 32 & 27.7 & \textbf{107.2} & \textbf{445.9} \\
Qwen2.5-14B  & 48 & 17.3 & 23.7  & 46.9 \\
Qwen3-8B     & 36 & 17.3 & 40.4  & 14.5 \\
\bottomrule
\end{tabular}

\end{table}

These outliers are weight-side counterparts of the \emph{super-weight} \citep{yu2024superweight} and \emph{massive activation} \citep{sun2024massive} phenomena documented in adjacent literature. Our contribution is per-block quantification across these entries: every one contains isolated outliers, and the per-entry extreme magnitude varies by nearly an order of magnitude. For the framework here, the relevant observation is that the body Weibull fit is robust to these outliers (by design of the middle 80\% protocol).

\subsection{Diagnostic case study: family-specific anomaly detection}
\label{sec:within-entry-qwen}

We apply the framework to a within-entry deviation flagged by our 12-entry cohort audit. The cohort audit surfaces three Qwen entries (Qwen2.5-7B/14B, Qwen3-8B) whose FFN $k$ falls below the Transmission band with a paired $\lambda$ collapse; per-block disaggregation localizes the deviation to shallow blocks ($\ell \leq 6$; the exact range varies by entry). Extending the cascade to eight additional Qwen-family entries (1.5B/3B/7B variants and Math-CPT pairs) yields an 11-entry Qwen cohort that partitions cleanly: all four 1.5B variants stay in-band; all seven 3B-and-above variants (3B--14B) exhibit the shallow deviation (with depth varying across 28L/36L/48L). Direct per-block $(k, \lambda)$ trajectories and 1D weight-magnitude histograms (Figures~\ref{fig:qwen-perblock-k}, \ref{fig:qwen-multilayer-hist}) confirm the deviation as a secondary small-magnitude population --- explicitly bimodal in 4 of the 9 flagged matrices, the rest showing an unseparated shoulder (Appendix~\ref{sec:appendix:qwen-diagnostic:detection}) --- structurally distinct from super-weights \citep{yu2024superweight}. The framework diagnoses this as a characteristic of the surveyed Qwen cohort; the full diagnostic chain is in Appendix~\ref{sec:appendix:qwen-diagnostic}.

\begin{figure}[!p]
\centering
\includegraphics[width=1.0\linewidth]{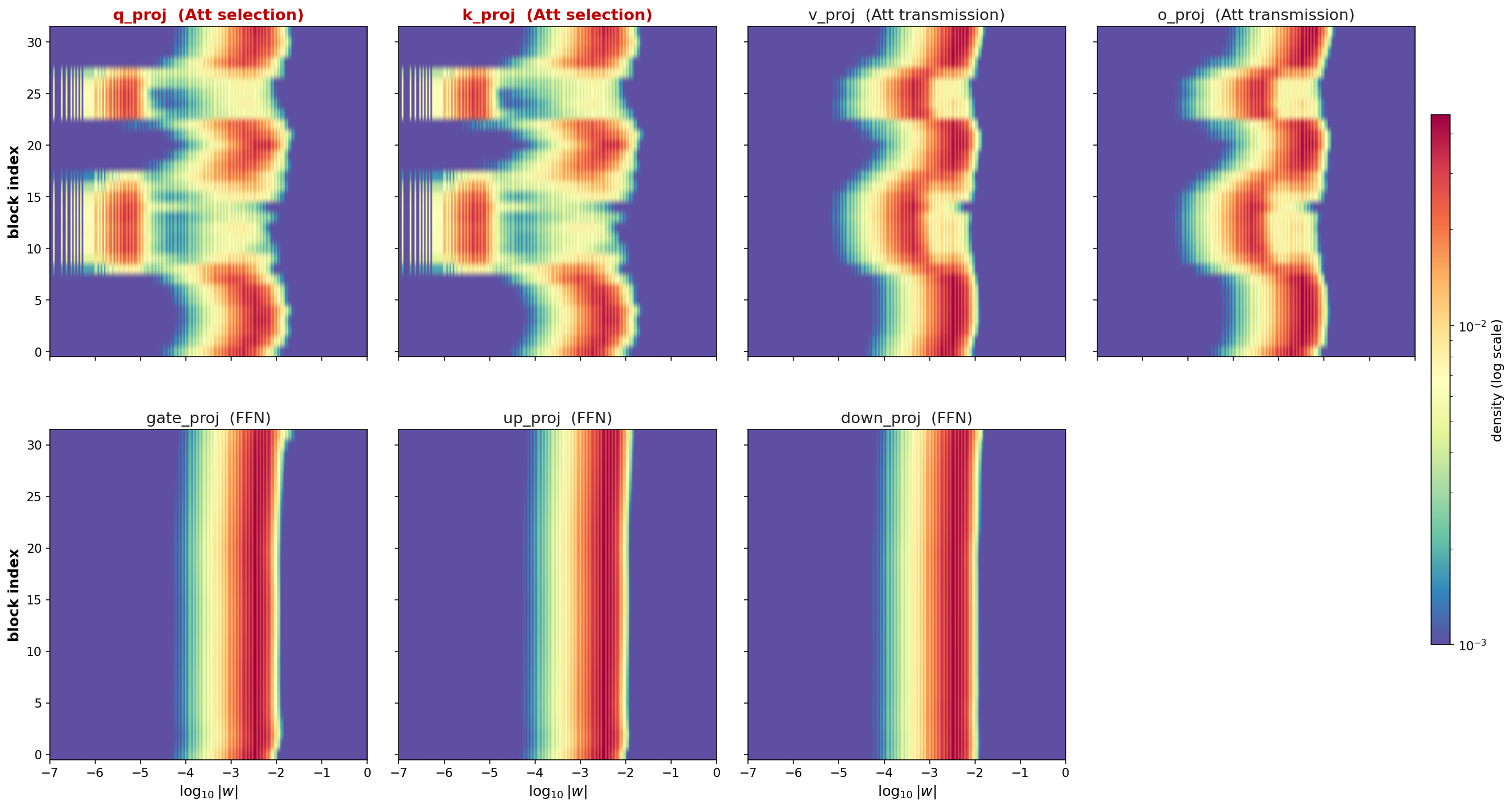}
\caption{\textbf{OLMo-1 (7B, terminal) per-layer $|w|$ distribution heatmap --- 7 components $\times$ 32 blocks.} Q/K (Selection Class) show an extended small-magnitude tail reaching $\log_{10}|w| \approx -5$ in blocks 8--17 and 23--27; V/O (Transmission Class) and FFN gate/up/down show tight distributions across blocks.}
\label{fig:f151}
\end{figure}

\section{Q/K Selection Evolution}
\label{sec:qk}

Section~\ref{sec:classes} established that $\Wq$ and $\Wk$ constitute the Selection Class and introduced three candidate mechanisms that may contribute to their departure from the initialization anchor. This section documents the empirical evidence along three dimensions: spatial localization across layer depth, temporal evolution with training budget, and architectural modulation of drift severity.

\subsection{Spatial Localization}
\label{sec:qk-spatial}

Figure~\ref{fig:f151} shows the per-layer per-component $|W|$ distribution heatmap for OLMo-1-7B. Q/K distributions start near the initialization body and progressively develop heavy tails --- lower $k$ --- concentrated in the \textbf{mid-to-deep blocks} (blocks 7--28 of 32 have $k_q$ or $k_k$ below $1.0$). This layer-wise pattern is consistent with prior findings that attention heads develop specialized roles \citep{voita2019analyzing}, with induction heads and other middle-layer-concentrated patterns characterized by \citet{elhage2021circuits} and \citet{olsson2022induction}. These works motivate inspecting layer localization; they do not establish a link between head specialization and the observed $k$ departure. The heavy-tail signal in $\Wq/\Wk$ is not uniform across depth but forms clusters in specific layer ranges, indicating that Selection is a spatially localized distributional departure rather than a global property of attention.

\begin{figure}[!p]
\centering
\includegraphics[width=0.90\linewidth]{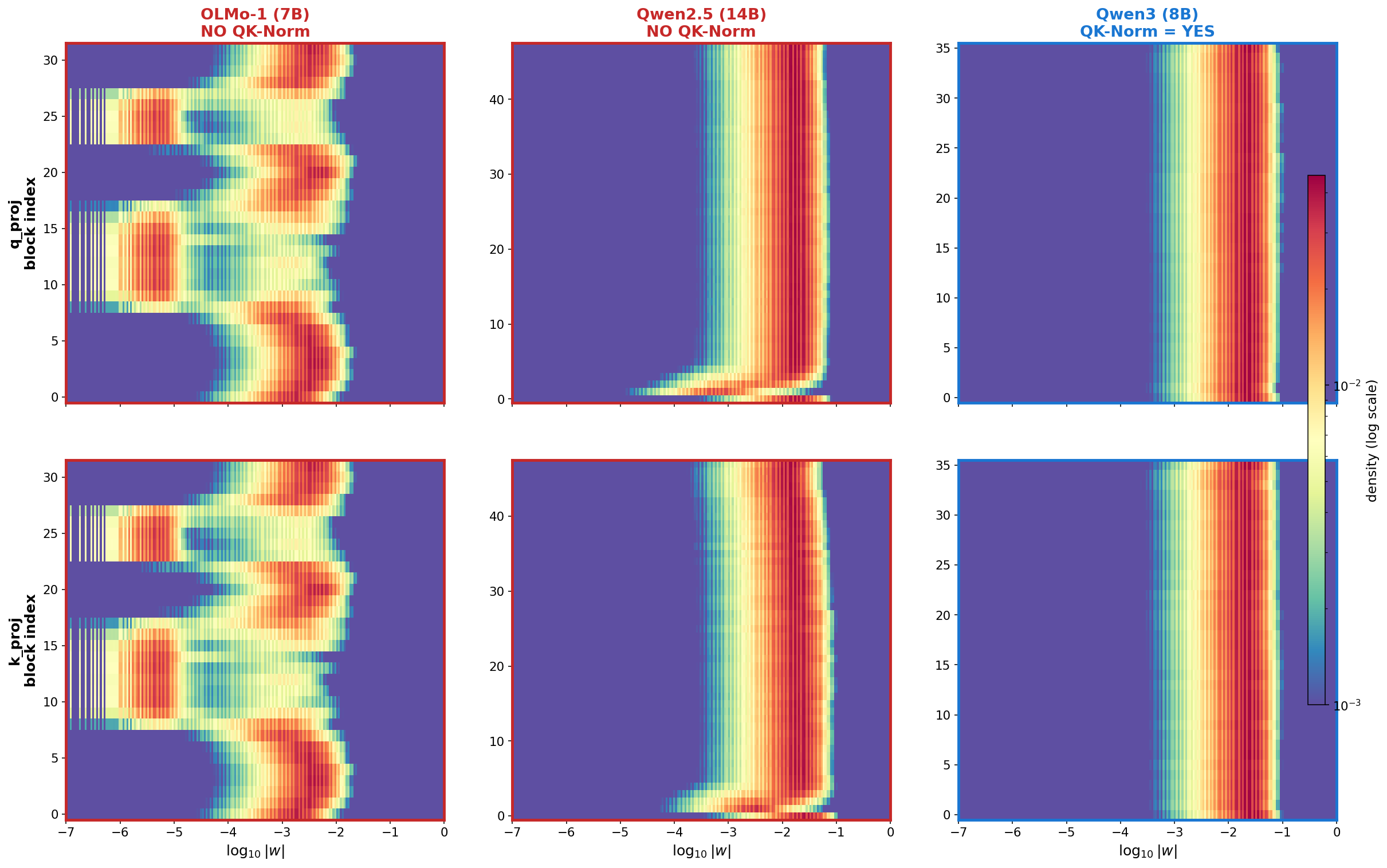}
\caption{\textbf{QK-Norm contrast: cross-family terminal Q/K per-layer heatmap.} OLMo-1 7B (red, no QK-Norm), Qwen2.5 14B (red, no QK-Norm), Qwen3 8B (blue, QK-Norm). NO-QK-Norm models show tail extending to $\log_{10}|w| \approx -5$ in shallow-to-mid blocks; Qwen3 (QK-Norm) tail compressed and confined to fewer blocks.}
\label{fig:f152}
\end{figure}

Qwen3-8B (with QK-Norm) shows visibly tighter Q/K distributions in shallow-to-mid blocks than Qwen2.5-14B (no QK-Norm; Figure~\ref{fig:f152}). The two models also differ in size, training data, and other recipe details, so QK-Norm can only be noted as one architectural difference among several; a controlled QK-Norm ablation on otherwise-identical training would be required to isolate the effect.

\subsection{Temporal Evolution}
\label{sec:qk-temporal}

\begin{figure}[!t]
\centering
\includegraphics[width=1.0\linewidth]{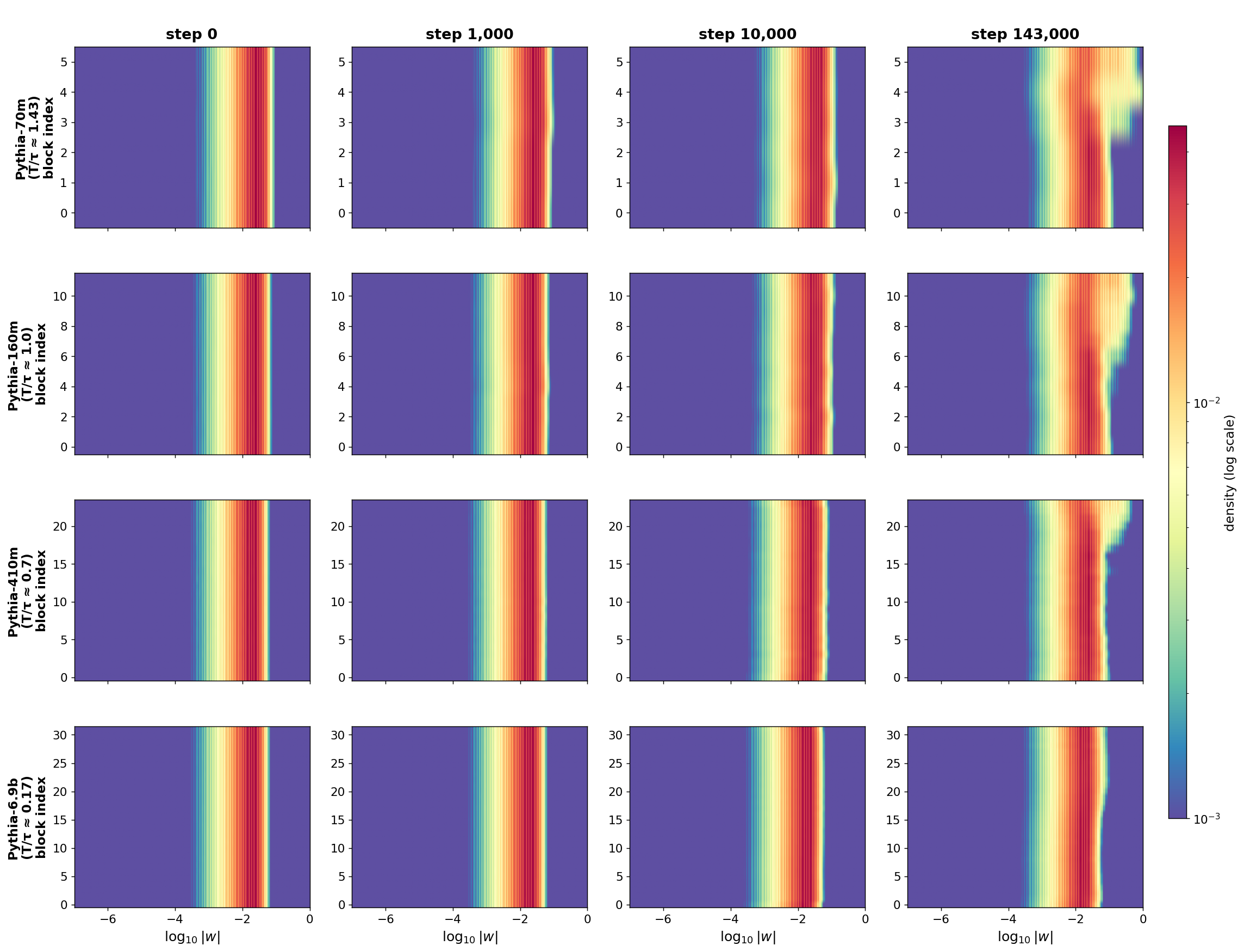}
\caption{\textbf{Pythia merged $\Wqkv$ per-layer $|w|$ distribution --- 4 sizes $\times$ 4 training steps.} Heatmap density of $\log_{10}|w|$ across blocks, with columns $=$ training step and rows $=$ Pythia size. The bulk distribution stays narrow; the left tail extends progressively, consistent with Selection-class specialization within the merged $\Wqkv$ tensor (fused storage prevents attribution to Q/K/V separately). The dimensionless $T/\tau = T \cdot \eta \cdot \lamwd$ partitions the Pythia sizes into Physical States (the four sizes displayed here; 1B is not shown and shares the same $T/\tau = 0.43$ as 410m): Pythia-70m ($T/\tau = 1.43$, Saturated), Pythia-160m ($T/\tau = 0.86$, Near-saturated), Pythia-410m ($T/\tau = 0.43$, Approaching), Pythia-6.9B ($T/\tau = 0.17$, Transition). Selection drift severity is approximately monotonic in $T/\tau$: 410M and 1B share the same $T/\tau$, and the 6.9B entry is level with 1B.}
\label{fig:f149}
\end{figure}

The dimensionless training budget is $T/\tau = T \cdot \eta \cdot \lamwd$, where $T$ is total training steps, $\eta$ is the peak learning rate, and $\lamwd$ is the weight-decay coefficient; equivalently $T/\tau = T / \tau_{\text{iter}}$ with $\tau_{\text{iter}} = 1/(\eta \lamwd)$ the AdamW characteristic timescale \citep{fan2025robust, wang2024adamw}. This quantity measures how many optimizer timescales the nominal peak-learning-rate budget corresponds to (under warmup and cosine decay the realized budget $\sum_t \eta_t \lamwd$ is smaller) during training. Within the Pythia family (5 sizes), the severity of Selection drift (read on the merged $\Wqkv$) is approximately monotonic in $T/\tau$: median $k_{\text{qkv}}$ rises from $1.05$ at the largest $T/\tau$ to $1.18$ at the smallest (Figure~\ref{fig:f149}), with two departures from strict ordering --- 410M and 1B share the same $T/\tau$ yet differ by $0.03$ in $k_{\text{qkv}}$, and the 6.9B entry is level with 1B rather than above it. All five Pythia sizes used identical 143k-step training budgets, but the 6.9B model uses a lower $\eta_{\text{peak}} = 1.2 \times 10^{-4}$ versus $\eta \in [3.0 \times 10^{-4}, 1.0 \times 10^{-3}]$ for the others, yielding $T/\tau = 0.17$ (Transition regime) instead of $0.43$--$1.43$. This is consistent with Selection drift being associated with the cumulative training signal $\eta \cdot \lamwd \cdot T$; size, width, depth, learning rate and (for 6.9B) the initialization recipe co-vary across these entries, so this observation does not isolate any one of them. This observation is consistent with candidate mechanism D5 (Appendix~\ref{sec:driving-forces}: cumulative training signal $T/\tau$).

\subsection{Architecture Associations and Candidate Mechanisms}
\label{sec:qk-mech}

\begin{figure}[!t]
\centering
\includegraphics[width=0.95\linewidth]{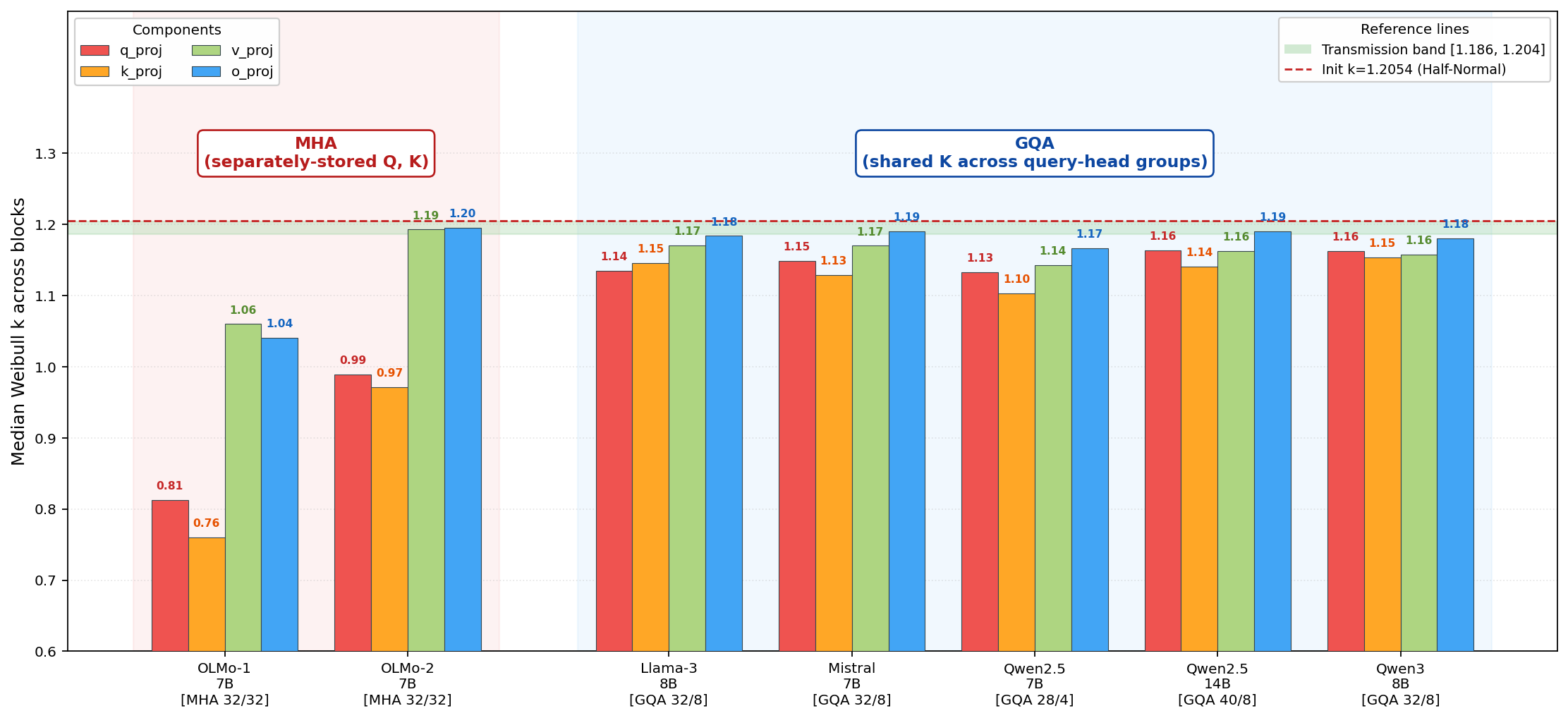}
\caption{\textbf{MHA vs GQA dichotomy: terminal $k_q$ and $k_k$ for the seven entries with separately stored Q/K projections.} Two architectural groups are shown: separately-stored MHA (OLMo-1, OLMo-2; deep Selection, $k \in [0.76, 0.99]$) and grouped-query attention (LLaMA-3, Mistral, Qwen2.5-7B/14B, Qwen3; mild Selection, $k \in [1.10, 1.16]$). The five Pythia entries store $\Wq$, $\Wk$, $\Wv$ in a single fused $\Wqkv$ tensor and therefore have no separately fitted $k_q$, $k_k$; their fused values ($k_{\text{qkv}} \in [1.05, 1.18]$, approximately monotonic in $T/\tau$, with the departures noted in Figure~\ref{fig:f149}) are reported there. Across this cohort, departure severity is associated with the storage architecture; K-head sharing in GQA is a candidate explanation, not one isolated by these observations.}
\label{fig:f159}
\end{figure}

The three candidate mechanisms introduced in Section~\ref{sec:classes} --- functional necessity, AdamW sign-descent, and softmax saturation feedback --- are consistent with Selection drift; their formal statement and supporting references are given in Appendix~\ref{sec:driving-forces} (D1, D2, D3). The observed outcome is a downward departure of $k$ below the Transmission band; the individual contributions of these candidates are not isolated here.

The severity of this drift is associated with attention architecture (Figure~\ref{fig:f159}). Models with separately-stored Q/K projections (OLMo-1, OLMo-2; multi-head attention) show the most severe drift: median terminal $k_q, k_k \in [0.76, 0.99]$, with individual blocks as low as $k = 0.25$ (OLMo-2-7B, block 2, $\Wk$). Grouped-query attention models (LLaMA-3, Mistral, Qwen2.5, Qwen3) show substantially less drift: median $k_q, k_k \in [1.10, 1.16]$, consistently below the Transmission band but far less extreme. The five Pythia checkpoints, using a merged $\Wqkv$ tensor, occupy a transitional zone: median $k_{\text{qkv}} \in [1.05, 1.18]$, approximately monotonic in $T/\tau$ (with the departures from strict ordering noted above). The architectural constraint of K-head sharing in GQA may limit the specialization freedom available to individual K-head projections, consistent with the observed MHA/GQA dichotomy --- an observational correlation; causal derivation requires controlled ablation (Section~\ref{sec:limitations}).

\FloatBarrier
\section{$\lambda$ Evolution}
\label{sec:lambda}

The Weibull scale parameter $\lambda$ measures the magnitude of weight matrices. Unlike the shape $k$, which is nearly invariant for Transmission Class components, $\lambda$ increases substantially during training.

\subsection{Initialization Reference}
\label{sec:lambda-init}

At initialization, $\lambda$ is fully determined by the per-component initializer: the four stored weight kinds ($\Wqkv$, $\Wo$, $W_{\text{FFN\_in}}$, $W_{\text{FFN\_out}}$) do not share a single $\sigma_{\text{init}}$ but split into input-side and output-side groups with distinct recipes. Within Pythia, the input/output ratio $\lambda_{\text{in}}/\lambda_{\text{out}}$ follows $L/\sqrt{10}$ for sizes 70m through 1B and $\sqrt{2L}$ for the 6.9B (Appendix~\ref{sec:appendix-init-pythia}, Table~\ref{tab:lambda-init-verify}; the analytical derivation (deterministic special-function integrals) in Appendix~\ref{sec:appendix-init} and 5-size verification give agreement within 0.13\% across all kinds). Training then collapses this recipe-specific initial ratio toward $\sim 1.2\times$ through the paired growth described below (for example, 6.9B terminal $\lambda_{\Wqkv}/\lambda_O \approx 1.29$ and $\lambda_{W_{\text{FFN\_in}}}/\lambda_{W_{\text{FFN\_out}}} \approx 1.14$, down from the initial $8\times$).

\subsection{Per-Component-Type Trajectories}
\label{sec:lambda-per-component}

\begin{figure}[p]
\centering
\begin{subfigure}{\linewidth}
  \centering
  \includegraphics[width=0.90\linewidth]{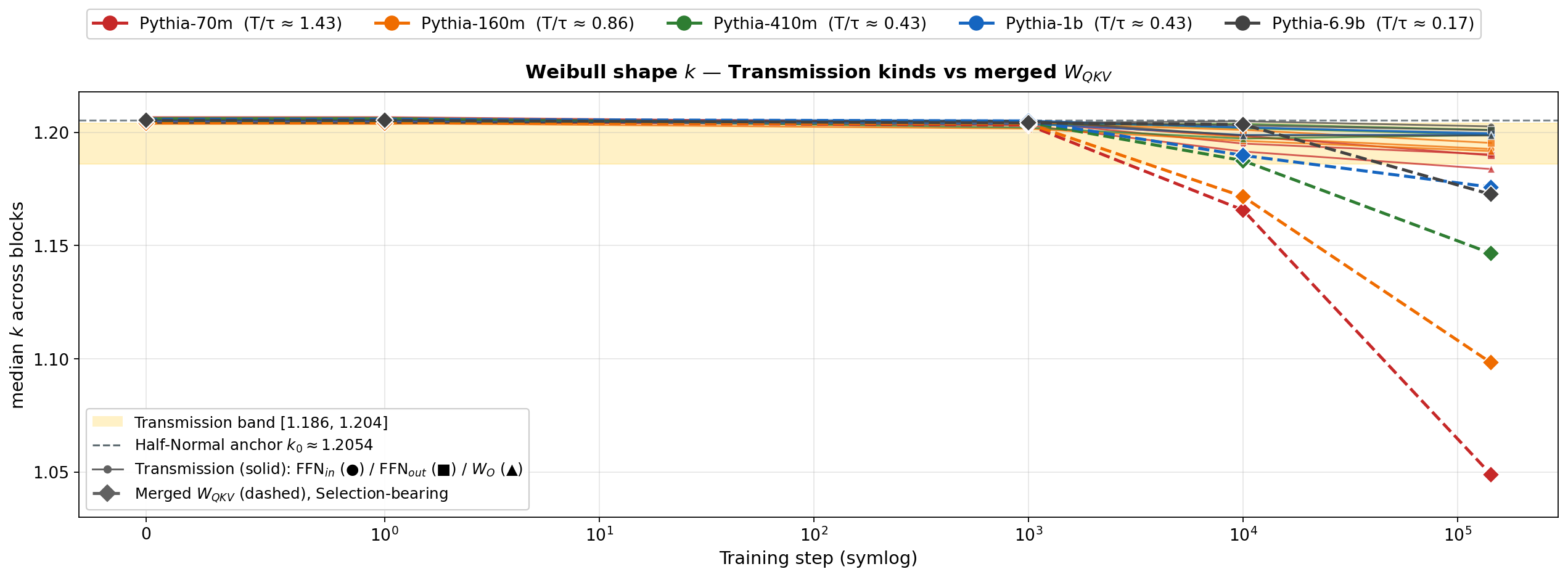}
  \caption{}\label{fig:f156a}
\end{subfigure}\\[2pt]
\begin{subfigure}{\linewidth}
  \centering
  \includegraphics[width=0.90\linewidth]{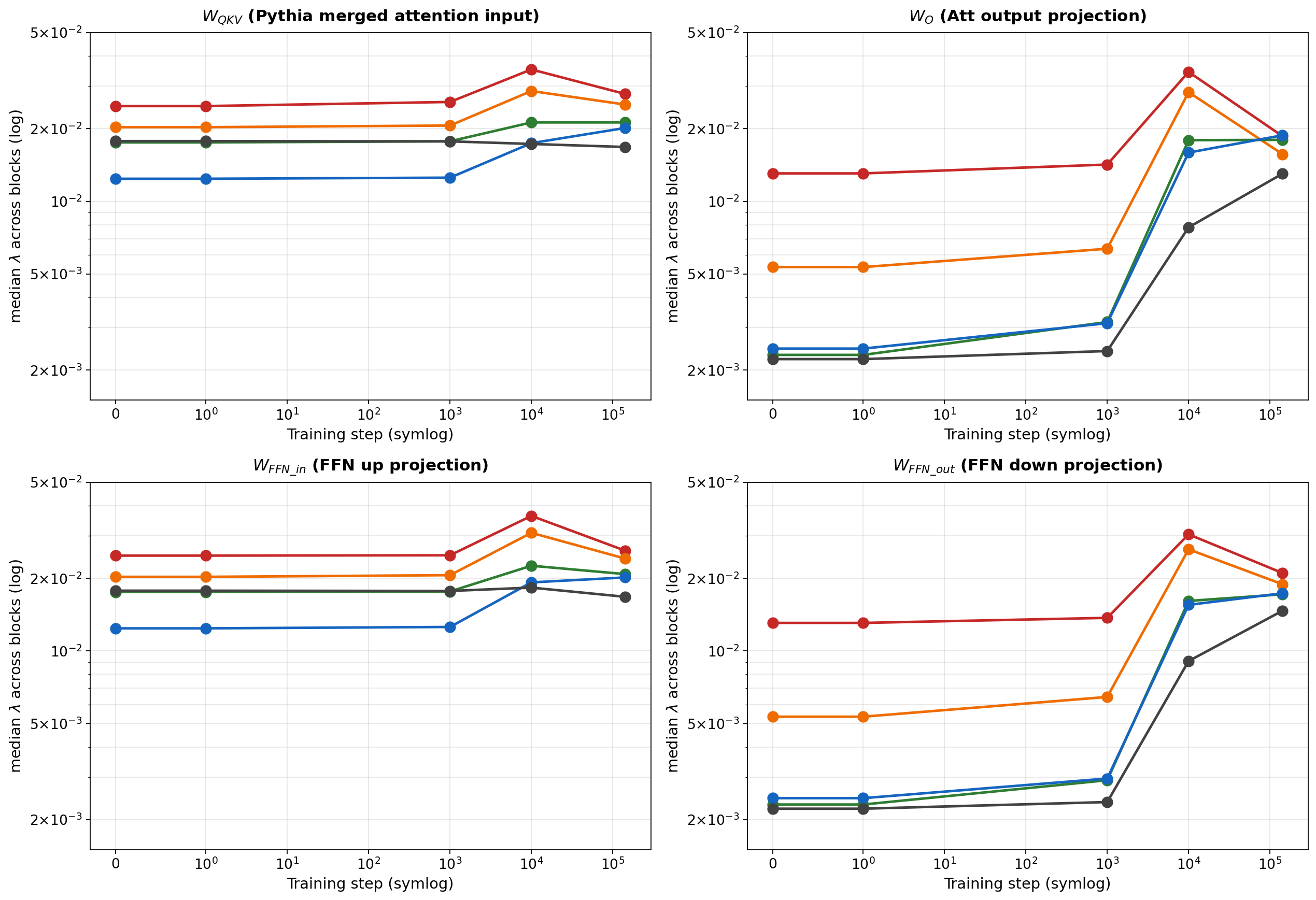}
  \caption{}\label{fig:f156b}
\end{subfigure}
\caption{\textbf{Weibull parameter trajectories from the same per-revision fits ---
median per-block values, 5 Pythia sizes $\times$ 5 released checkpoints
(steps 0, 1, 1000, 10{,}000, 143{,}000), four stored weight kinds.}
\textbf{(a) Gate view (shape $k$).} The three Transmission kinds (solid:
$W_{\text{FFN\_in}}$, $W_{\text{FFN\_out}}$, $\Wo$; 15 series) remain pinned near
the half-Normal anchor $k_0 \approx 1.2054$ throughout training, ending in 1.184--1.203
against the Transmission band $[1.186, 1.204]$ (the single marginal value, 1.184,
is $\Wo$ of the most-saturated 70m). The merged $\Wqkv$ (dashed), labelled mixed under the rule of
Section~\ref{sec:classes} as it stores $\Wq$/$\Wk$ together with $\Wv$, drifts far below the band within the same runs
(terminal 1.049--1.176, deepest for the most-saturated sizes). The invariance is
selective, not generic.
\textbf{(b) Readout view (scale $\lambda$).} Median $\lambda$ per block, each panel
overlaying the 5 Pythia sizes (70m--6.9B); shared color key in panel (a). The
paired growth across $\Wo$ and $W_{\text{FFN\_out}}$ (Pearson $r = 0.9967$ on 25
size--step combinations) is reported in Section~\ref{sec:lambda-coupling}.
\textbf{Reading order is (a) then (b):} fit adequacy is checked first;
$k$ then gates reference-regime interpretation before $\lambda$ is read. $\lambda_{QKV}$ in (b) is the raw fitted
scale; the gate in (a) flags its departure, which is why the $\lambda$ scaling
analyses exclude $\Wqkv$ (Table~\ref{tab:lambda}).}
\label{fig:f156}
\end{figure}

Figure~\ref{fig:f156} shows the median $\lambda$ trajectory for each of the four stored weight kinds ($\Wqkv$, $\Wo$, $W_{\text{FFN\_in}}$, $W_{\text{FFN\_out}}$) across the 5 Pythia sizes. Each component type exhibits a characteristic trajectory shape. The $\lambda_O$ trajectory is non-monotonic: it rises through learning-rate warmup, peaks near step 10k (warmup completion), then retreats during cosine decay. The degree of post-peak retreat depends on the Physical State of the model. Models with $T/\tau > 1$ (Pythia-70m, $T/\tau = 1.43$, Saturated) show strong post-peak retreat ($\lambda_O$ retreats $46\%$ from its peak); models with $T/\tau \approx 0.4$--$0.9$ (160m--1B) show moderate retreat or near-flat trajectories; models with $T/\tau < 0.2$ (6.9B, Transition) are still rising at the terminal checkpoint, indicating unsaturated training. The growth itself reflects increased weight magnitude under AdamW, directionally consistent with the steady-state scaling analysis of \citet{fan2025robust}. Table~\ref{tab:lambda} reports initialization-to-terminal growth across all five sizes.

\begin{table}[h]
\centering
\small
\caption{$\lambda_O$ growth across the Pythia family. The ``Mean $\lambda$ (3 Trans. kinds, terminal)'' column reports the terminal-checkpoint mean $\lambda$ across the three Transmission Class kinds $\Wo$, $W_{\text{FFN\_in}}$, $W_{\text{FFN\_out}}$ (mean per kind across blocks, then mean over the three kinds; $\Wqkv$ excluded as mixed (fused storage) per Section~\ref{sec:classes}). This is the quantity used in the within-Pythia scaling fit of Section~\ref{sec:lambda-scaling} and reported again for Pythia in Table~\ref{tab:lambda-crossfamily}.}
\label{tab:lambda}
\begin{tabular}{lccccccc}
\toprule
Size & $T/\tau$ & $\lambda_O$ init & $\lambda_O$ terminal & $\Delta\lambda_O$ & Mean $\lambda$ (3 Trans. kinds, terminal) & Growth \\
\midrule
70m & 1.43 (Saturated) & 0.0131 & 0.0186 & 0.0056 & 0.0224 & $+43\%$ \\
160m & 0.86 (Near-sat.) & 0.0053 & 0.0157 & 0.0103 & 0.0200 & $+194\%$ \\
410m & 0.43 (Approaching) & 0.0023 & 0.0180 & 0.0157 & 0.0190 & $+678\%$ \\
1B & 0.43 (Approaching) & 0.0025 & 0.0188 & 0.0163 & 0.0191 & $+665\%$ \\
6.9B & 0.17 (Transition) & 0.0022 & 0.0130 & 0.0108 & 0.0149 & $+487\%$ \\
\bottomrule
\end{tabular}

\end{table}

\subsection{Per-Component-Type Coupling}
\label{sec:lambda-coupling}

The scale increase is not localized to the attention output. Across all 25 size--step combinations, $\log(\lambda_O)$ vs.\ $\log(\lambda_{\text{FFN\_out}})$ yields Pearson $r = 0.9967$ (log--log, $n = 25$). This near-perfect correlation indicates that both component types scale co-vary strongly across the 25 pooled size--step combinations: as $\Wo$ scales, $\lambda_{W_{\text{FFN\_out}}}$ tracks it, consistent with their shared output-side role in the residual stream. This paired behavior contrasts with the non-monotonic single-trajectory of $\lambda_O$ (peaking at warmup completion, then cosine-decaying, Section~\ref{sec:lambda-per-component}).

\begin{figure}[!htbp]
\centering
\includegraphics[width=0.95\linewidth]{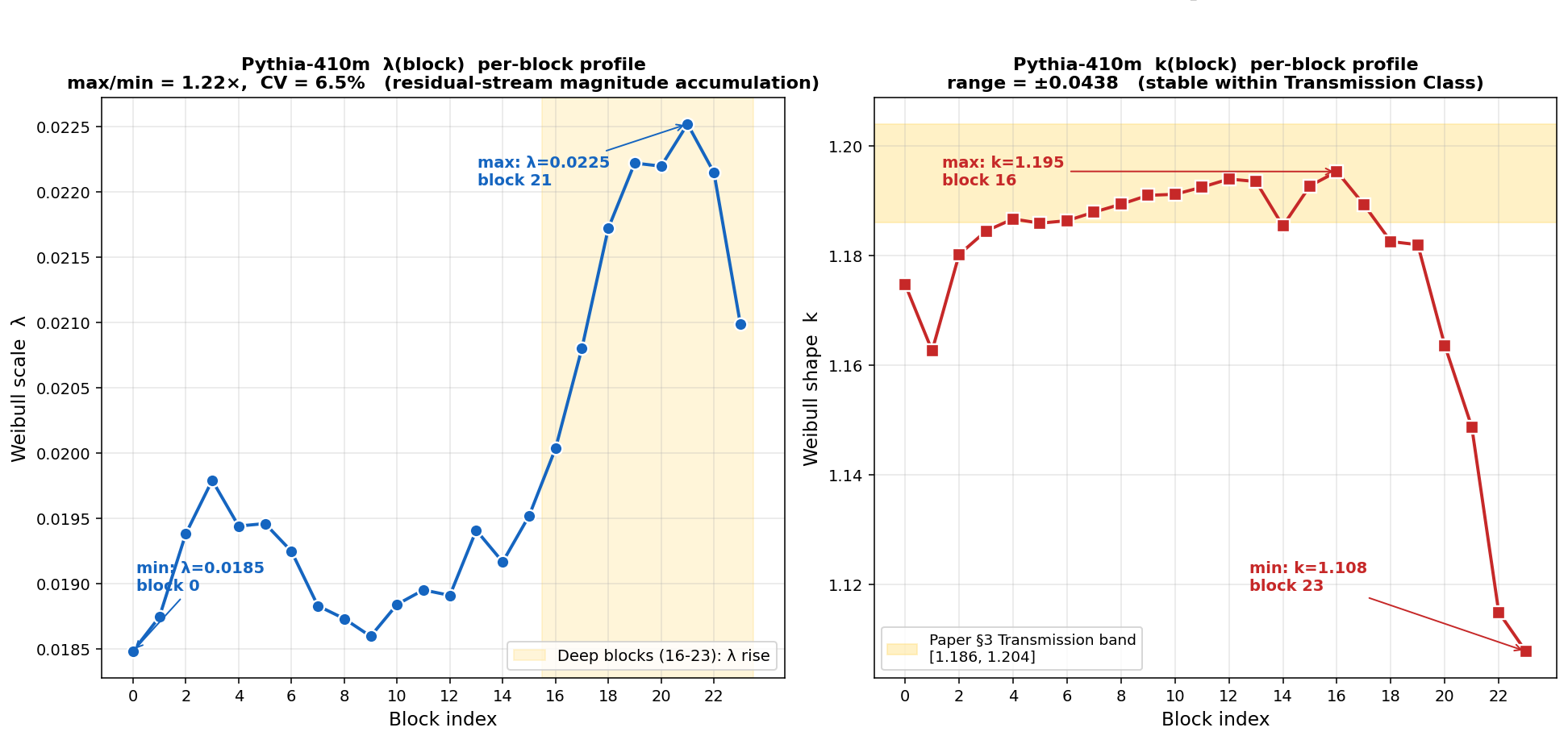}
\caption{\textbf{Pythia-410m terminal per-block $(k, \lambda)$ profile (24 blocks, aggregate fit per block).} $\lambda$ (left) shows depth-dependent rise in deep blocks (16--23, max $\sim 1.22\times$ shallow); $k$ (right) is a block-aggregate fit and declines with depth, reaching $1.10$ at block~23; aggregate fits pool components with different $\sigma_{\text{init}}$ and are therefore not directly comparable to the per-component band $[1.186, 1.204]$ (cf.\ Figure~\ref{fig:f47}). The cause of the depth decline is not established here. Per-block depth-heterogeneity complements component-paired uniformity (Pearson $r = 0.9967$ between $\lambda_O$ and $\lambda_{\text{FFN\_out}}$): the two phenomena coexist.}
\label{fig:f156c}
\end{figure}

The component-paired uniformity does not, however, extend to per-block uniformity within a model. Figure~\ref{fig:f156c} shows the per-block $(k, \lambda)$ profile for Pythia-410m terminal across 24 blocks: $\lambda$ exhibits a modest depth-dependent rise (max/min $\approx 1.22\times$, CV $\approx 6.5\%$), with deep blocks (16--23) systematically larger --- a pattern for which residual-stream magnitude accumulation is a candidate account, not established here. Block-aggregate $k$ declines with depth, reaching $1.10$ at block~23. Because a block-aggregate fit pools components with different $\sigma_{\text{init}}$, its value is not directly comparable to the per-component Transmission band $[1.186, 1.204]$ (the same pooling effect documented in Figure~\ref{fig:f47}); we therefore report the depth profile itself rather than band membership. The two phenomena --- component-paired uniformity and per-block depth heterogeneity --- coexist.

\subsection{Cross-Size Scaling}
\label{sec:lambda-scaling}

\begin{figure}[!htbp]
\centering
\includegraphics[width=0.85\linewidth]{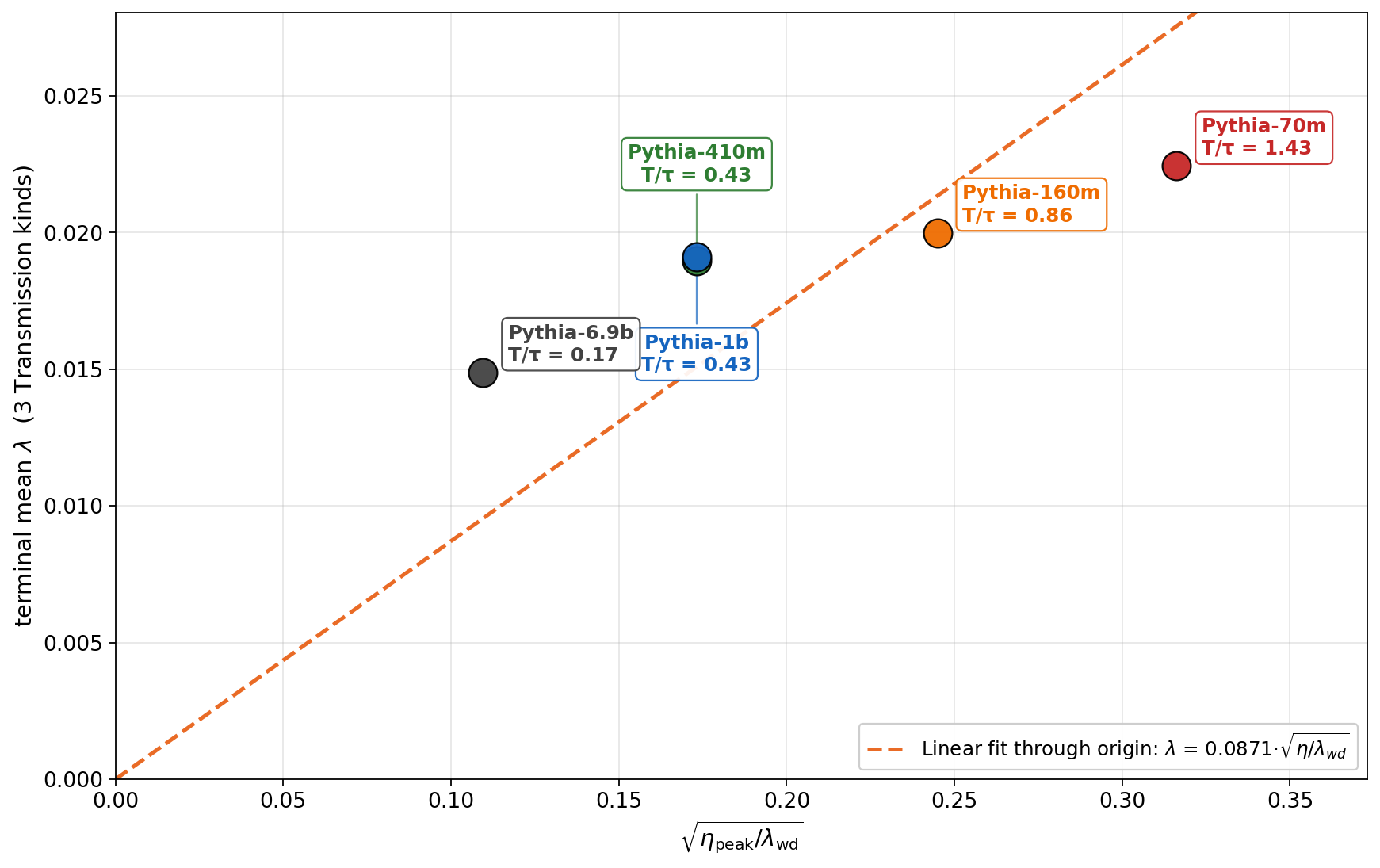}
\caption{\textbf{Within-Pythia $\lambda$ scaling vs.\ $\sqrt{\eta_{\text{peak}}/\lamwd}$.} Terminal mean $\lambda$ at step 143{,}000, averaged across the three Transmission Class kinds ($\Wo$, $W_{\text{FFN\_in}}$, $W_{\text{FFN\_out}}$; $\Wqkv$ excluded as mixed per Section~\ref{sec:classes}) plotted against $\sqrt{\eta_{\text{peak}}/\lamwd}$ for the 5 Pythia sizes. Linear fit through origin gives slope $0.087$, Pearson $r = 0.94$. Per-size deviations of 7--36\% indicate a directional rather than quantitative match with the scaling law of \citet{fan2025robust}.}
\label{fig:f170}
\end{figure}

Within the Pythia training family --- varying model size, with the 6.9B entry differing from the 70M--1B entries in both initialization recipe (Appendix~\ref{sec:appendix-init-pythia}) and learning rate --- the terminal mean $\lambda$ across the three Transmission Class kinds ($\Wo$, $W_{\text{FFN\_in}}$, $W_{\text{FFN\_out}}$; $\Wqkv$ excluded) shows a within-family scaling trend tracking $\sqrt{\eta_{\text{peak}}/\lamwd}$ ($n = 5$ sizes spanning \textbf{4 distinct} $(\eta, \lamwd)$ settings --- 410M and 1B share the same configuration and hence the same predicted value --- linear fit through origin: $\lambda = 0.087 \cdot \sqrt{\eta/\lamwd}$, Pearson $r = 0.94$; leave-one-size-out $r$ ranges over $[0.90, 0.97]$, and $r = 0.97$ with the 6.9B entry --- the one differing in initialization recipe --- excluded; Figure~\ref{fig:f170}). The scaling direction --- larger learning rate or lower weight decay produces larger terminal $\lambda$ --- shows \textbf{directional consistency} with the AdamW steady-state $\sqrt{\eta/\lamwd}$ scaling analysis of \citet{fan2025robust}, whose direct validation covers LLaMA models with $d \leq 2048$; four of the five Pythia sizes fall within this width range, while Pythia-6.9B ($d = 4096$) lies outside it. Per-size deviations from the linear fit range from 7\% to 36\%, indicating directional consistency rather than quantitative magnitude match.

For completeness, Table~\ref{tab:lambda-crossfamily} reports the terminal mean $\lambda$ (across Transmission Class kinds) for all 12 cohort entries alongside the published $\eta_{\text{peak}}$ and $\lamwd$ used to compute $\sqrt{\eta/\lamwd}$. The cross-family slope ratio $\lambda / \sqrt{\eta/\lamwd}$ is shown in the final column and is the quantity discussed in Section~\ref{sec:limitations} (cross-family scaling limitation).

\begin{table}[h]
\centering
\small
\caption{Cross-family terminal mean $\lambda$ across the Transmission Class kinds (for Pythia: $\Wo$, $W_{\text{FFN\_in}}$, $W_{\text{FFN\_out}}$; for the SwiGLU families LLaMA-3/Mistral/Qwen2.5/Qwen3/OLMo-1/OLMo-2: $\Wo$, $W_{\text{gate}}$, $W_{\text{up}}$, $W_{\text{down}}$; $\Wq$, $\Wk$ excluded as Selection class and $\Wqkv$ as mixed per Section~\ref{sec:classes}) and the slope ratio $\lambda/\sqrt{\eta/\lamwd}$ across the 12 cohort entries. Aggregation is mean per kind across blocks then mean over kinds. Within Pythia (top block, same family, differing recipe at 6.9B), the ratio spans $\sim 1.9\times$ (0.071--0.136). Across the 7 non-Pythia 7B--14B entries (bottom block), the ratio spans $\sim 6.6\times$ (Mistral-7B at 0.048 to Qwen3-8B at 0.319). The non-Pythia $\eta_{\text{peak}}$ and $\lamwd$ values are taken from published training configurations and may differ from end-of-training schedules; this is an additional source of uncertainty discussed in Section~\ref{sec:limitations}.}
\label{tab:lambda-crossfamily}
\begin{tabular}{lccccc}
\toprule
Family / size & $\eta_{\text{peak}}$ & $\lamwd$ & $\sqrt{\eta/\lamwd}$ & Terminal mean $\lambda$ & $\lambda / \sqrt{\eta/\lamwd}$ \\
\midrule
Pythia-70m   & $1.0 \times 10^{-3}$ & $0.01$ & $0.316$ & $0.0224$ & $0.071$ \\
Pythia-160m  & $6.0 \times 10^{-4}$ & $0.01$ & $0.245$ & $0.0200$ & $0.082$ \\
Pythia-410m  & $3.0 \times 10^{-4}$ & $0.01$ & $0.173$ & $0.0190$ & $0.110$ \\
Pythia-1B    & $3.0 \times 10^{-4}$ & $0.01$ & $0.173$ & $0.0191$ & $0.110$ \\
Pythia-6.9B  & $1.2 \times 10^{-4}$ & $0.01$ & $0.110$ & $0.0149$ & $0.136$ \\
\midrule
OLMo-1-7B    & $3.0 \times 10^{-4}$ & $0.10$ & $0.055$ & $0.0032$ & $0.058$ \\
OLMo-2-7B    & $3.0 \times 10^{-4}$ & $0.10$ & $0.055$ & $0.0167$ & $0.304$ \\
LLaMA-3-8B   & $3.0 \times 10^{-4}$ & $0.10$ & $0.055$ & $0.0097$ & $0.178$ \\
Mistral-7B   & $3.0 \times 10^{-4}$ & $0.10$ & $0.055$ & $0.0026$ & $0.048$ \\
Qwen2.5-7B   & $3.0 \times 10^{-4}$ & $0.10$ & $0.055$ & $0.0138$ & $0.253$ \\
Qwen2.5-14B  & $3.0 \times 10^{-4}$ & $0.10$ & $0.055$ & $0.0171$ & $0.312$ \\
Qwen3-8B     & $5.0 \times 10^{-4}$ & $0.10$ & $0.071$ & $0.0225$ & $0.319$ \\
\bottomrule
\end{tabular}

\end{table}

\subsection{Relationship to $k$}
\label{sec:lambda-vs-k}

While $\lambda_O$ grows by factors of $1.4\times$--$7.8\times$ across the Pythia family (Table~\ref{tab:lambda}), Transmission Class $k$ stays in or near $[1.186, 1.204]$ (Figure~\ref{fig:f156}a). Within the Pythia Transmission kinds, $\lambda$ grows while $k$ stays in or near the reference band; Selection and mixed projections show reference-relative shape departures; and Transmission body fits remain adequate outside the flagged Regime-B matrices, with Selection fits showing a wider spread of fit quality (Table~\ref{tab:fit-quality}). The two parameters therefore carry complementary information: within this scope, shape is preserved while magnitude scales.

\FloatBarrier
\section{Using the Framework Across Resolution, Time and References}
\label{sec:practical-use}

Use the framework to describe a checkpoint, locate differences across its matrices, and follow
changes during training. Figure~\ref{fig:practical-workflow} links the measurements to three views
of the same weights: structural resolution, time and reference comparison.

\paragraph{Measurements.}
Apply the fixed fitting protocol to the uncentered magnitudes of each stored matrix
(Section~\ref{sec:framework}). Keep the model, checkpoint, parameter name, layer, component,
tensor shape and stored precision with the results. Check fit adequacy first, then read $k$ as
reference-relative body shape and $\lambda$ as fitted magnitude under matched conditions.
A marked departure or inadequate fit prompts raw-weight inspection before scale interpretation.
Choose supplementary statistics for the question: empirical quantiles describe the magnitude
profile, direct tail measurements locate extremes (Section~\ref{sec:superweight}), and row/column
RMS profiles can describe magnitude differences across channels.

\paragraph{Resolution.}
Construct a checkpoint profile from the per-matrix results, preserving component and layer
identity. This avoids mixing component scales in a single pooled fit (Figure~\ref{fig:f47}).
Compare FFN, separately stored Q/K, V/O and mixed fused QKV in their architectural context, then
use depth profiles to locate the affected matrices. Component roles are defined by the
architecture (Section~\ref{sec:classes}). If a secondary weight population appears---a
distinguishable group in the magnitude distribution---inspect rows and columns to identify its
carrier. In the Qwen case (Appendix~\ref{sec:appendix:qwen-diagnostic:channels}), this step
locates a shared suppressed-channel set across gate/up rows and down columns. Describe the
groups separately and retain their channel indices as targets for subsequent activation
measurements or controlled ablations. Raw channel statistics provide this finer description;
departures whose structure remains unclear can be reported with empirical distributions.

\paragraph{Time.}
Repeat the measurements on the same matrices across checkpoints and record their learning-rate
schedule. Joint trajectories of $k$, $\lambda$ and fit quality locate magnitude growth, peaks,
recession and body-shape changes (Section~\ref{sec:lambda}). Read these trajectories alongside
warmup and decay: intermediate checkpoints show when a change occurs and which layers contribute.
Compare matched intervals to select stages for closer study. With nearly stable $k$, rescaled
quantile profiles can check whether the measured body changes mainly in magnitude. If the
question concerns directional reorganization, add the displacement measurements of
Appendix~\ref{sec:appendix-displacement}. Questions about the source of a transition additionally
require the relevant data, schedule, gradient or optimizer-state records.

\paragraph{References.}
The database in Appendix~\ref{sec:appendix-api} provides per-component summaries and training
metadata. Read its numerical entries together with the worked patterns in this paper: the FFN
band, architecture-associated Q/K differences, shallow-FFN secondary populations and isolated
tail outliers. Select references by component, storage architecture, fitting protocol and
checkpoint stage. Record the initialization, training data, learning rate, weight decay and
training budget where available; retain unknown settings as missing information. A matched
initial or earlier checkpoint gives a direct baseline, while the released cohort positions the
measurement among published models. Use the FFN band at its reported entry-median resolution
and inspect individual matrices against an appropriate component reference. These comparisons
help choose a baseline, a structural level and a training interval for further investigation.

\begingroup
\makeatletter
\setlength{\@fptop}{0pt}
\makeatother
\begin{figure}[!htbp]
\centering
\includegraphics[width=\linewidth]{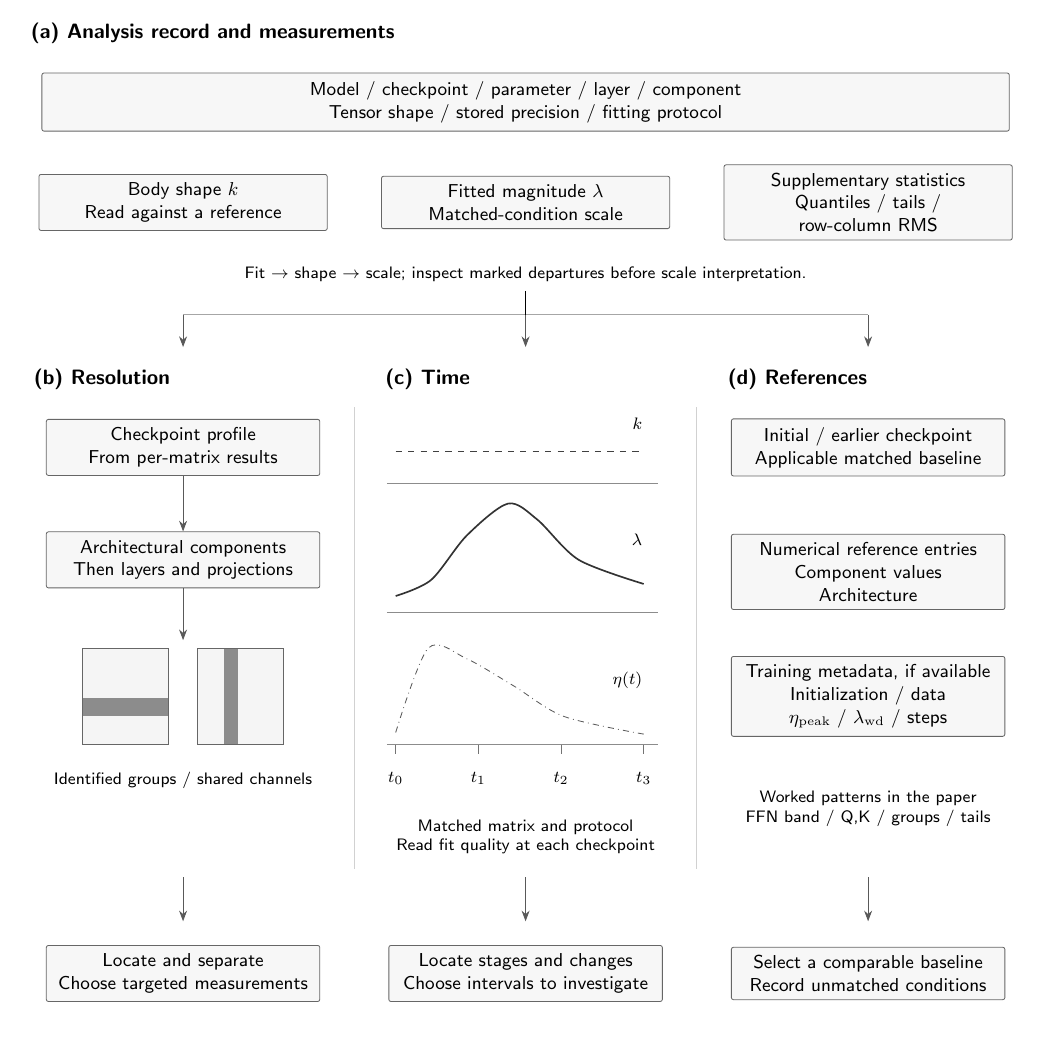}
\caption{\textbf{Measurements and three views of checkpoint structure.}
\textbf{(a)} Keep the analysis object and fitting context with each measurement; read fit adequacy,
shape and scale in the paper's prescribed order.
\textbf{(b)} Resolve per-matrix summaries into components, layers and identified groups or channels.
\textbf{(c)} Track matched matrices over time and read their measurements alongside the learning-rate
schedule. Curves are schematic, with independent vertical scales, and show a possible trajectory;
their alignment does not establish causation.
\textbf{(d)} Select comparison context from numerical entries, available configuration metadata and
worked patterns. Combine the three views for the question at hand and refine
where needed. For functional decisions, pair the weight analysis with task-relevant evaluation;
magnitude alone does not prescribe a model edit.}
\label{fig:practical-workflow}
\end{figure}
\clearpage
\endgroup

\FloatBarrier
\section{Limitations and Future Work}
\label{sec:limitations}

Three limitations of the current work deserve explicit acknowledgment.

\paragraph{Interpretation and mechanism.} We document that $\Wq$ and $\Wk$ show
training-associated departures from the calibrated reference regime, and identify five candidate
driving forces (Appendix~\ref{sec:driving-forces}, D1--D5); we do not provide controlled ablation
experiments isolating the contribution of each. Two interpretive boundaries should be stated
explicitly. First, \emph{Transmission} and \emph{Selection} are descriptive labels for groups defined
by architectural role (Section~\ref{sec:classes}); $k$ measures the shape of an element-wise magnitude
distribution and does not measure whether a component performs transmission or selection. Second, the
element-wise marginal discards entry correlations, singular directions, cross-layer alignment and
interactions with activations --- two matrices with the same marginal can implement different
functions --- so this readout complements, rather than replaces, spectral, activation-level and
behavioral analyses. Within the present cohort, architecture co-varies with model family, training
data and recipe, normalization placement, and size/width/depth, so the association between the
MHA/GQA distinction and Selection severity (Section~\ref{sec:qk}) cannot be attributed to any one of
them; K-head sharing remains one candidate modulator. A definitive causal claim requires an ablation
--- training an otherwise-identical model in MHA versus GQA configuration from scratch --- which we
leave to future work.

\paragraph{$\lambda$ scaling.} The within-Pythia $\lambda \propto \sqrt{\eta/\lamwd}$ trend
(Section~\ref{sec:lambda-scaling}) rests on a small sample: $n = 5$ sizes spanning only four distinct
$(\eta, \lamwd)$ settings, with per-size deviations of 7--36\%, and the 6.9B entry additionally
changing initialization recipe. It also transfers a result across objects: \citet{fan2025robust}
analyze the scaling of matrix spectral norms, whereas we track an element-wise Weibull scale
parameter, so the correspondence is an empirical analogy rather than a direct test of the same law.
Cross-family generalization --- applying the same formula to the 7B--14B modern cohort
($d \in [3584, 5120]$) --- is observational only. Cross-family comparison of
$\lambda/\sqrt{\eta/\lamwd}$ is affected by uncontrolled recipe and model factors beyond $\eta$ and
$\lamwd$, including initialization scheme (small\_init vs Kaiming vs scaled\_normal, which sets
$\sigma_{\text{init}}$ and hence $\lambda_{\text{init}}$;
Appendix~\ref{sec:appendix-init-pythia}), normalization placement (Pre-LN vs Post-LN vs QK-Norm),
training data and schedule, and width/depth; these are not captured by $\eta/\lamwd$ alone. Across the
7 non-Pythia entries, the per-entry ratio $\lambda/\sqrt{\eta/\lamwd}$ spans $\sim 6.6\times$
(Mistral-7B at $0.048$ to Qwen3-8B at $0.319$), substantially larger than the $\sim 1.9\times$
Pythia-internal range. OLMo-1-7B and OLMo-2-7B make the point concretely: the two share the same
published $\eta_{\text{peak}}$ and $\lamwd$, and therefore the same predicted $\sqrt{\eta/\lamwd}$,
yet their terminal mean $\lambda$ differs by roughly fivefold ($0.0032$ versus $0.0167$;
Table~\ref{tab:lambda-crossfamily}). This shows that the scaling relation alone does not account for
the observed spread; the source of the gap is not isolated by our data. (The non-Pythia $\eta$ and
$\lamwd$ values are taken from published training configurations; actual end-of-training schedules may
differ, an additional source of uncertainty.) The cross-family scatter is therefore reported as a
descriptive observation, not as a cross-family scaling law.

\paragraph{Downstream performance.} The framework characterizes the statistical structure of weight
distributions. What we demonstrate is a weight-structure diagnostic: a shape alarm computed from
released weights alone is localized to specific layers and verified as concrete channel-level
structure in the raw weights (Section~\ref{sec:within-entry-qwen},
Appendix~\ref{sec:appendix:qwen-diagnostic:channels}). What we do not establish is a relation between
that structure and downstream task performance or behavior. Such a correlation, if it exists, would
make $k$ and $\lambda$ useful as early-stopping or checkpoint-selection proxies --- a direction we
list here rather than a claim we make, and one we leave to future work.

\paragraph{Ongoing work.} Two questions follow directly from the readout established here, and we are
pursuing both outside the scope of this paper: how the $\lambda$ dynamics couple quantitatively to the
optimizer's control variables, and how the same readout responds to properties of the training data. We
keep these questions out of this paper's claims until they can be supported at the same standard as the
results reported above.

\FloatBarrier
\subsubsection*{Acknowledgments}
The author thanks Chuan Fang for helpful discussions and constructive feedback on the manuscript, and Beijing Azure River Tech Co., Ltd. for its generous support and provision of computational resources. The author also thanks the Action Editor and reviewers for their constructive suggestions, which enriched and strengthened the paper and helped clarify the practical uses of the framework.

\bibliographystyle{tmlr}
\bibliography{references}

\begin{thebibliography}{34}
\providecommand{\natexlab}[1]{#1}
\providecommand{\url}[1]{\texttt{#1}}
\expandafter\ifx\csname urlstyle\endcsname\relax
  \providecommand{\doi}[1]{doi: #1}\else
  \providecommand{\doi}{doi: \begingroup \urlstyle{rm}\Url}\fi

\bibitem[Ainslie et~al.(2023)Ainslie, Lee-Thorp, de~Jong, Zemlyanskiy,
  Lebr{\'o}n, and Sanghai]{ainslie2023gqa}
Joshua Ainslie, James Lee-Thorp, Michiel de~Jong, Yury Zemlyanskiy, Federico
  Lebr{\'o}n, and Sumit Sanghai.
\newblock {GQA}: Training generalized multi-query transformer models from
  multi-head checkpoints.
\newblock In \emph{Proceedings of the 2023 Conference on Empirical Methods in
  Natural Language Processing ({EMNLP})}, pp.\  4895--4901, Singapore, 2023.
  Association for Computational Linguistics.

\bibitem[Barsbey et~al.(2021)Barsbey, Sefidgaran, Erdogdu, Richard, and
  {\c{S}}im{\c{s}}ekli]{barsbey2021heavy}
Melih Barsbey, Milad Sefidgaran, Murat~A. Erdogdu, Ga{\"e}l Richard, and Umut
  {\c{S}}im{\c{s}}ekli.
\newblock Heavy tails in sgd and compressibility of overparametrized neural
  networks.
\newblock In \emph{NeurIPS}, 2021.
\newblock arXiv:2106.03795.

\bibitem[Bellido \& Fiesler(1993)Bellido and Fiesler]{bellido1993normal}
Ivan Bellido and Emile Fiesler.
\newblock Do backpropagation trained neural networks have normal weight
  distributions?
\newblock In \emph{ICANN'93}, pp.\  772--775. Springer, 1993.

\bibitem[Black et~al.(2022)Black, Biderman, Hallahan, Anthony, Gao, Golding,
  He, Leahy, McDonell, Phang, Pieler, Prashanth, Purohit, Reynolds, Tow, Wang,
  and Weinbach]{black2022gptneox}
Sid Black, Stella Biderman, Eric Hallahan, Quentin Anthony, Leo Gao, Laurence
  Golding, Horace He, Connor Leahy, Kyle McDonell, Jason Phang, Michael Pieler,
  USVSN~Sai Prashanth, Shivanshu Purohit, Laria Reynolds, Jonathan Tow, Ben
  Wang, and Samuel Weinbach.
\newblock {GPT-NeoX-20B}: An open-source autoregressive language model.
\newblock In \emph{Proceedings of {BigScience} Episode \#5 -- Workshop on
  Challenges \& Perspectives in Creating Large Language Models}, 2022.

\bibitem[Bondarenko et~al.(2023)Bondarenko, Nagel, and
  Blankevoort]{bondarenko2023quantizable}
Yelysei Bondarenko, Markus Nagel, and Tijmen Blankevoort.
\newblock Quantizable transformers: Removing outliers by helping attention
  heads do nothing.
\newblock In \emph{Advances in Neural Information Processing Systems
  ({NeurIPS})}, volume~36, 2023.

\bibitem[Dettmers et~al.(2023)Dettmers, Pagnoni, Holtzman, and
  Zettlemoyer]{dettmers2023qlora}
Tim Dettmers, Artidoro Pagnoni, Ari Holtzman, and Luke Zettlemoyer.
\newblock Qlora: Efficient finetuning of quantized llms.
\newblock In \emph{NeurIPS}, 2023.
\newblock arXiv:2305.14314.

\bibitem[Dotzel et~al.(2024)Dotzel, Chen, Kotb, Prasad, Wu, Li, Abdelfattah,
  and Zhang]{dotzel2024students}
Jordan Dotzel, Yuzong Chen, Bahaa Kotb, Sushma Prasad, Gang Wu, Sheng Li,
  Mohamed~S. Abdelfattah, and Zhiru Zhang.
\newblock Learning from students: Applying t-distributions to explore accurate
  and efficient formats for llms.
\newblock In \emph{ICML}, 2024.
\newblock arXiv:2405.03103.

\bibitem[Elhage et~al.(2021)Elhage, Nanda, Olsson, Henighan, Joseph, Mann,
  Askell, Bai, Chen, Conerly, DasSarma, Drain, Ganguli, Hatfield-Dodds,
  Hernandez, Jones, Kernion, Lovitt, Ndousse, Amodei, Brown, Clark, Kaplan,
  McCandlish, and Olah]{elhage2021circuits}
Nelson Elhage, Neel Nanda, Catherine Olsson, Tom Henighan, Nicholas Joseph, Ben
  Mann, Amanda Askell, Yuntao Bai, Anna Chen, Tom Conerly, Nova DasSarma, Dawn
  Drain, Deep Ganguli, Zac Hatfield-Dodds, Danny Hernandez, Andy Jones, Jackson
  Kernion, Liane Lovitt, Kamal Ndousse, Dario Amodei, Tom Brown, Jack Clark,
  Jared Kaplan, Sam McCandlish, and Chris Olah.
\newblock A mathematical framework for transformer circuits.
\newblock \emph{Transformer Circuits Thread}, 2021.
\newblock Published Dec 22, 2021.
  \url{https://transformer-circuits.pub/2021/framework/index.html}.

\bibitem[Fan et~al.(2025)Fan, Liu, Zhao, Yuan, and Gu]{fan2025robust}
Zhiyuan Fan, Yifeng Liu, Qingyue Zhao, Angela Yuan, and Quanquan Gu.
\newblock Robust layerwise scaling rules by proper weight decay tuning.
\newblock \emph{arXiv preprint arXiv:2510.15262}, 2025.

\bibitem[Fortuin et~al.(2022)Fortuin, Garriga-Alonso, Ober, Wenzel, R{\"a}tsch,
  Turner, van~der Wilk, and Aitchison]{fortuin2022bayesian}
Vincent Fortuin, Adri{\`a} Garriga-Alonso, Sebastian~W. Ober, Florian Wenzel,
  Gunnar R{\"a}tsch, Richard~E. Turner, Mark van~der Wilk, and Laurence
  Aitchison.
\newblock Bayesian neural network priors revisited.
\newblock In \emph{ICLR}, 2022.
\newblock arXiv:2102.06571.

\bibitem[Girrbach et~al.(2023)Girrbach, Christensen, Winther, Akata, and
  Koepke]{girrbach2023persistence}
Leander Girrbach, Anders Christensen, Ole Winther, Zeynep Akata, and A.~Sophia
  Koepke.
\newblock Addressing caveats of neural persistence with deep graph persistence.
\newblock \emph{Transactions on Machine Learning Research}, 2023.
\newblock arXiv:2307.10865.

\bibitem[G{\"u}rb{\"u}zbalaban et~al.(2021)G{\"u}rb{\"u}zbalaban,
  {\c{S}}im{\c{s}}ekli, and Zhu]{gurbuzbalaban2021heavy}
Mert G{\"u}rb{\"u}zbalaban, Umut {\c{S}}im{\c{s}}ekli, and Lingjiong Zhu.
\newblock The heavy-tail phenomenon in sgd.
\newblock In \emph{ICML}, 2021.

\bibitem[He et~al.(2025)He, Tu, Jaiswal, Shen, Yuan, Liu, and
  Yin]{he2025alphadecay}
Di~He, Songjun Tu, Ajay Jaiswal, Li~Shen, Ganzhao Yuan, Shiwei Liu, and Lu~Yin.
\newblock {AlphaDecay}: Module-wise weight decay for heavy-tailed balancing in
  {LLMs}.
\newblock In \emph{Advances in Neural Information Processing Systems
  ({NeurIPS})}, 2025.

\bibitem[Kaul et~al.(2025)Kaul, Ma, Elezi, and Deng]{kaul2025attention}
Prannay Kaul, Chengcheng Ma, Ismail Elezi, and Jiankang Deng.
\newblock From attention to activation: Unravelling the enigmas of large
  language models.
\newblock In \emph{International Conference on Learning Representations
  ({ICLR})}, 2025.

\bibitem[Kim et~al.(2018)Kim, Yim, Ha, Lee, and Kang]{kim2018generalized}
Doyun Kim, Han~Young Yim, Sanghyuck Ha, Changgwun Lee, and Inyup Kang.
\newblock Convolutional neural network quantization using generalized gamma
  distribution.
\newblock \emph{arXiv:1810.13329}, 2018.

\bibitem[Kunstner et~al.(2023)Kunstner, Chen, Lavington, and
  Schmidt]{kunstner2023noise}
Frederik Kunstner, Jacques Chen, J.~Wilder Lavington, and Mark Schmidt.
\newblock Noise is not the main factor behind the gap between {SGD} and {Adam}
  on transformers, but sign descent might be.
\newblock In \emph{International Conference on Learning Representations
  ({ICLR})}, 2023.

\bibitem[Martin \& Mahoney(2019)Martin and Mahoney]{martin2019traditional}
Charles~H. Martin and Michael~W. Mahoney.
\newblock Traditional and heavy-tailed self regularization in neural network
  models.
\newblock In \emph{Proceedings of the 36th International Conference on Machine
  Learning ({ICML})}, volume~97 of \emph{Proceedings of Machine Learning
  Research}, pp.\  4284--4293. PMLR, 2019.

\bibitem[Martin \& Mahoney(2020)Martin and Mahoney]{martin2020heavy}
Charles~H. Martin and Michael~W. Mahoney.
\newblock Heavy-tailed universality predicts trends in test accuracies for very
  large pre-trained deep neural networks.
\newblock In \emph{Proceedings of the 2020 {SIAM} International Conference on
  Data Mining ({SDM})}, pp.\  505--513. SIAM, 2020.

\bibitem[Martin \& Mahoney(2021)Martin and Mahoney]{martin2021simpson}
Charles~H. Martin and Michael~W. Mahoney.
\newblock Post-mortem on a deep learning contest: a simpson's paradox and the
  complementary roles of scale metrics versus shape metrics.
\newblock \emph{arXiv:2106.00734}, 2021.

\bibitem[Martin et~al.(2021)Martin, Peng, and Mahoney]{martin2021predicting}
Charles~H. Martin, Tongsu Peng, and Michael~W. Mahoney.
\newblock Predicting trends in the quality of state-of-the-art neural networks
  without access to training or testing data.
\newblock \emph{Nature Communications}, 12:\penalty0 4122, 2021.

\bibitem[Nguyen \& Salazar(2019)Nguyen and Salazar]{nguyen2019transformers}
Toan~Q. Nguyen and Julian Salazar.
\newblock Transformers without tears: Improving the normalization of
  self-attention.
\newblock In \emph{International Conference on Spoken Language Translation
  ({IWSLT})}, 2019.

\bibitem[Olsson et~al.(2022)Olsson, Elhage, Nanda, Joseph, DasSarma, Henighan,
  Mann, Askell, Bai, Chen, Conerly, Drain, Ganguli, Hatfield-Dodds, Hernandez,
  Johnston, Jones, Kernion, Lovitt, Ndousse, Amodei, Brown, Clark, Kaplan,
  McCandlish, and Olah]{olsson2022induction}
Catherine Olsson, Nelson Elhage, Neel Nanda, Nicholas Joseph, Nova DasSarma,
  Tom Henighan, Ben Mann, Amanda Askell, Yuntao Bai, Anna Chen, Tom Conerly,
  Dawn Drain, Deep Ganguli, Zac Hatfield-Dodds, Danny Hernandez, Scott
  Johnston, Andy Jones, Jackson Kernion, Liane Lovitt, Kamal Ndousse, Dario
  Amodei, Tom Brown, Jack Clark, Jared Kaplan, Sam McCandlish, and Chris Olah.
\newblock In-context learning and induction heads.
\newblock \emph{Transformer Circuits Thread, arXiv preprint arXiv:2209.11895},
  2022.
\newblock Published Mar 8, 2022.

\bibitem[Si et~al.(2025)Si, Jiang, and Shen]{si2025gaussian}
Chongjie Si, Jingjing Jiang, and Wei Shen.
\newblock Unveiling the mystery of weight in large foundation models: Gaussian
  distribution never fades.
\newblock \emph{arXiv:2501.10661}, 2025.

\bibitem[Sun et~al.(2024)Sun, Chen, Kolter, and Liu]{sun2024massive}
Mingjie Sun, Xinlei Chen, J.~Zico Kolter, and Zhuang Liu.
\newblock Massive activations in large language models.
\newblock \emph{arXiv preprint arXiv:2402.17762}, 2024.
\newblock COLM 2024.

\bibitem[Thamm et~al.(2022)Thamm, Staats, and Rosenow]{thamm2022random}
Matthias Thamm, Max Staats, and Bernd Rosenow.
\newblock Random matrix analysis of deep neural network weight matrices.
\newblock \emph{Physical Review E}, 106:\penalty0 054124, 2022.

\bibitem[Voita et~al.(2019)Voita, Talbot, Moiseev, Sennrich, and
  Titov]{voita2019analyzing}
Elena Voita, David Talbot, Fedor Moiseev, Rico Sennrich, and Ivan Titov.
\newblock Analyzing multi-head self-attention: Specialized heads do the heavy
  lifting, the rest can be pruned.
\newblock In \emph{Proceedings of the 57th Annual Meeting of the Association
  for Computational Linguistics ({ACL})}, pp.\  5797--5808, Florence, Italy,
  2019. Association for Computational Linguistics.

\bibitem[Voita et~al.(2023)Voita, Ferrando, and Nalmpantis]{voita2023dead}
Elena Voita, Javier Ferrando, and Christoforos Nalmpantis.
\newblock Neurons in large language models: Dead, $n$-gram, positional.
\newblock \emph{arXiv preprint arXiv:2309.04827}, 2023.

\bibitem[Wang \& Aitchison(2024)Wang and Aitchison]{wang2024adamw}
Xi~Wang and Laurence Aitchison.
\newblock How to set {AdamW}'s weight decay as you scale model and dataset
  size.
\newblock \emph{arXiv preprint arXiv:2405.13698}, 2024.
\newblock Preprint; v3 released 1 Jun 2025.

\bibitem[Weibull(1951)]{weibull1951statistical}
Waloddi Weibull.
\newblock A statistical distribution function of wide applicability.
\newblock \emph{Journal of Applied Mechanics}, 18\penalty0 (3):\penalty0
  293--297, 1951.
\newblock \doi{10.1115/1.4010337}.

\bibitem[Wong et~al.(2025)Wong, Zhang, Mahon, Luk, Isopoussu, and
  Zhao]{wong2025secondarysinks}
Jeffrey T.~H. Wong, Cheng Zhang, Louis Mahon, Wayne Luk, Anton Isopoussu, and
  Yiren Zhao.
\newblock On the existence and behavior of secondary attention sinks.
\newblock \emph{arXiv preprint arXiv:2512.22213}, 2025.

\bibitem[Wu et~al.(2025)Wu, Huang, Wen, and Han]{wu2025generalized}
Jun Wu, Patrick Huang, Jiangtao Wen, and Yuxing Han.
\newblock It takes a good model to train a good model: Generalized gaussian
  priors for optimized llms.
\newblock \emph{arXiv:2506.00486}, 2025.

\bibitem[Yu et~al.(2024)Yu, Wang, Shan, Reed, and Wan]{yu2024superweight}
Mengxia Yu, De~Wang, Qi~Shan, Colorado~J. Reed, and Alvin Wan.
\newblock The super weight in large language models.
\newblock \emph{arXiv preprint arXiv:2411.07191}, 2024.

\bibitem[Zhang et~al.(2026)Zhang, He, Fang, Tan, Chen, and
  Zhuang]{zhang2026domlora}
Suoxin Zhang, Run He, Di~Fang, Xiang Tan, Kaixuan Chen, and Huiping Zhuang.
\newblock Rethinking adapter placement: A dominant adaptation module
  perspective.
\newblock \emph{arXiv preprint arXiv:2605.06183}, 2026.

\bibitem[Zhou et~al.(2023)Zhou, Pang, Liu, Martin, Mahoney, and
  Yang]{zhou2023tempbalance}
Yefan Zhou, Tianyu Pang, Keqin Liu, Charles~H. Martin, Michael~W. Mahoney, and
  Yaoqing Yang.
\newblock Temperature balancing, layer-wise weight analysis, and neural network
  training.
\newblock In \emph{NeurIPS}, 2023.
\newblock arXiv:2312.00359.

\end{thebibliography}

\clearpage
\appendix

\section{Theory Details}
\label{sec:appendix-theory}

\subsection{Half-Normal Initialization Anchor}
\label{sec:appendix-init}

The models in our cohort initialize weights from i.i.d.\ Gaussian distributions: $w_{ij} \sim \mathcal{N}(0, \sigma_{\text{init}}^2)$. The absolute value $|w_{ij}|$ therefore follows a half-Normal distribution. Fitting half-Normal samples to a Weibull distribution via least-squares on the Weibull probability plot with middle-80\% trim --- the protocol used throughout this work --- yields a fit pair $(k_0, \lambda_0)$ that is fully determined by the fit protocol and $\sigma_{\text{init}}$.

\begin{figure}[!htbp]
\centering
\includegraphics[width=0.65\linewidth]{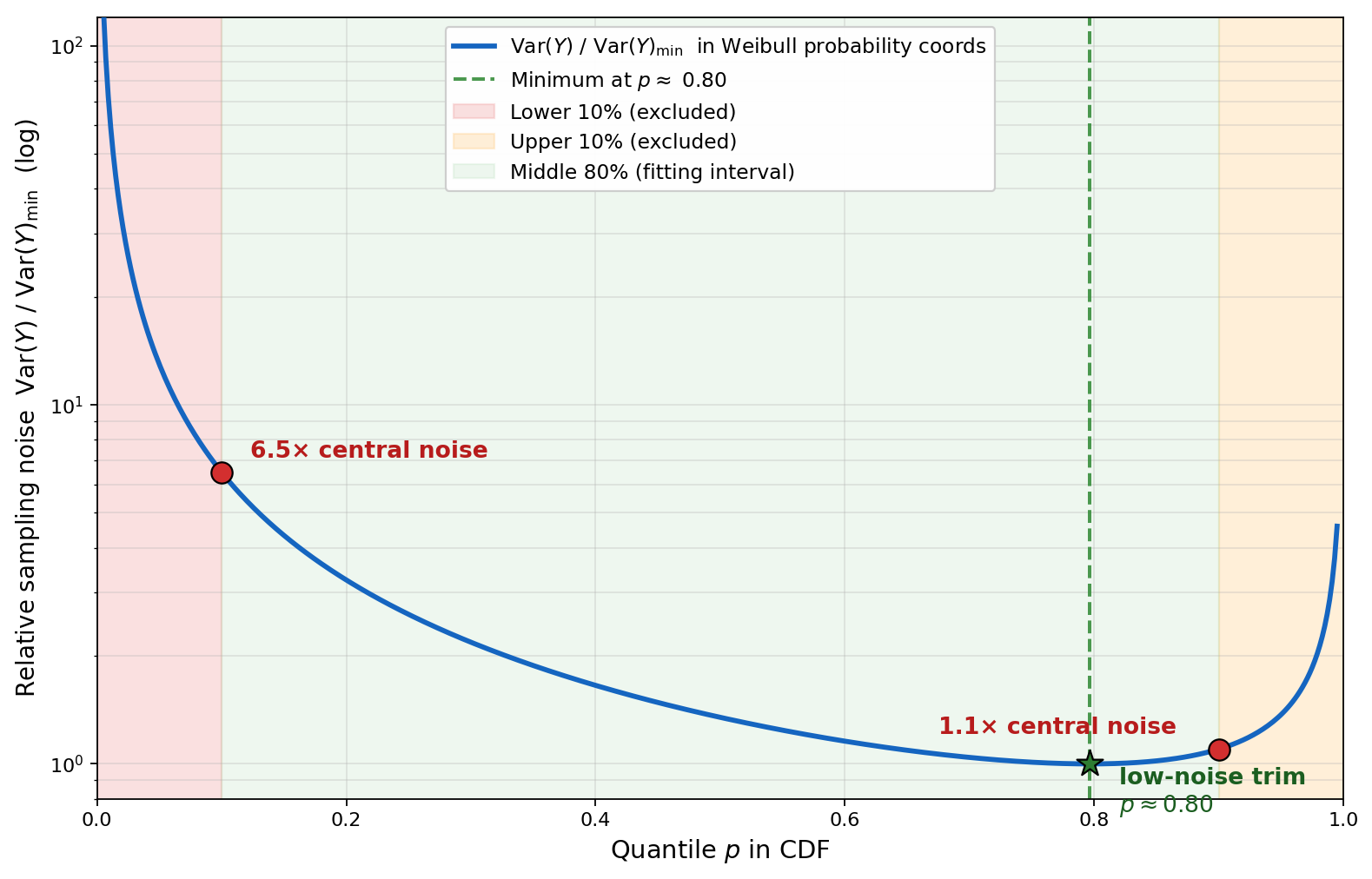}
\caption{\textbf{Sampling-noise profile of the probability-plot coordinate.} Theoretical curve of $\mathrm{Var}(Y)/\mathrm{Var}(Y)_{\min}$ as a function of quantile $p$, where $Y = \ln(-\ln(1-F))$ is the Weibull probability transform. Three shaded regions: bottom 10\% (red, $p < 0.10$), middle 80\% (green, $0.10 \leq p \leq 0.90$, the fitting interval used for the $k_{80\%}$ fit), top 10\% (orange, $p > 0.90$). The variance diverges at both endpoints and reaches its minimum at $p^* \approx 0.797$; the bottom 10\% region carries $\sim 6.5\times$ more noise than that minimum, while the $p=0.90$ endpoint carries about $1.1\times$. The middle-80\% protocol is a robust convention for recovering the body Weibull shape, excluding both endpoint regions where this coordinate is noisiest.}
\label{fig:f167}
\end{figure}

\paragraph{Trim fraction: sampling-noise motivation.} In Weibull probability coordinates $Y = \ln(-\ln(1 - F))$, the empirical rank $Y_i$ of the $i$-th order statistic has sampling-noise variance $\text{Var}(Y_i) \propto p / [(1-p) \cdot (\ln(1-p))^2]$, where $p = i/N$ is the cumulative fraction. This variance diverges at both ends ($p \to 0$ and $p \to 1$) and reaches a minimum near $p \approx 0.80$. The bottom 10\% of the distribution carries approximately $6.5\times$ more measurement noise than the minimum-variance point (Figure~\ref{fig:f167}). The middle 80\% trim is therefore adopted as our standard protocol: a robust convention motivated by this variance profile.

\paragraph{Analytic derivation of the fit constants.} The half-Normal CDF $F(x) = \text{erf}\!\big(x/(\sigma_{\text{init}}\sqrt{2})\big)$ has inverse $x(F) = \sigma_{\text{init}} \cdot \sqrt{2} \cdot \text{erf}^{-1}(F)$. Writing $g(F) = \ln(\sqrt{2}\,\text{erf}^{-1}(F))$ and $Y(F) = \ln(-\ln(1-F))$, the probability-plot coordinates become $X = \ln\sigma_{\text{init}} + g(F)$ and $Y = Y(F)$. Least-squares regression of $Y$ on $X$ over $F \in [0.1, 0.9]$ gives
\begin{equation}
k_0 = \frac{\text{Cov}_{[0.1, 0.9]}(g, Y)}{\text{Var}_{[0.1, 0.9]}(g)}, \qquad
\frac{\lambda_0}{\sigma_{\text{init}}} = \exp\!\left(\overline{g}_{[0.1, 0.9]} - \overline{Y}_{[0.1, 0.9]} / k_0\right).
\end{equation}
The integrals are deterministic numerical integrals of special functions over the trim interval. They evaluate to
\begin{equation}
k_0 \approx 1.2054, \qquad \lambda_0 \approx 0.8875\,\sigma_{\text{init}}.
\label{eq:lambda-init-anchor}
\end{equation}
The constant $0.8875$ is specific to this fit protocol and is not interchangeable with moment-matching values, which would yield different conversion factors. We confirm $0.8875$ by direct measurement at the step-0 checkpoint across all 5 Pythia sizes and 4 stored-weight kinds, which yield $\lambda_{\text{init}} / \sigma_{\text{init}} \in [0.887, 0.889]$, agreeing with the deterministic analytical derivation above to within $0.13\%$.

\paragraph{Empirical zero-mean verification.} The half-Normal anchor presumes symmetric initialization. We verify this assumption directly on Pythia-70m raw weight matrices: at step 0 the per-matrix means satisfy $|\mu|/\sigma < 0.3\%$ across all 4 stored-weight kinds, and at the terminal checkpoint (step 143{,}000) the ratio remains below $0.6\%$ --- an observed near-centering consistent with the zero-mean symmetric initialization, an approximate preservation of marginal symmetry under training, and the averaging effect of large matrices; we do not attribute it to any specific architectural mechanism. The estimand throughout this work --- in the released pipeline (Appendix~\ref{sec:appendix-api}) as in every result reported here --- is the uncentered magnitude $|W|$: no per-matrix mean is subtracted before histogramming. A location-offset sensitivity check shows that centering would not change the conclusions: we verified directly on raw \texttt{bf16} weights that the centered convention $|W - \bar{W}|$ differs from $|W|$ only in the region $|w| \lesssim |\mu|$ (relative difference $\leq 0.26\%$ on body statistics), and the secondary population reported in Section~\ref{sec:within-entry-qwen} lies entirely at $|w| \geq 10 \cdot \max|\mu|$, where the two conventions are numerically indistinguishable.

\paragraph{Per-matrix zero-mean verification (cross-architecture).} Beyond the Pythia-70m verification above, we cross-validate the zero-mean assumption on Qwen2.5-14B: across 48 layers $\times$ 4 Transmission Class kinds (192 matrices), the median $|{\mu}|/\sigma$ per kind ranges from 0.008\% to 0.059\%, with a maximum of 0.319\% ($W_\text{gate}$). Both Pythia-70m (70\,M, MHA-merged, small\_init) and Qwen2.5-14B (14\,B, GQA-7:1, scaled\_normal) --- spanning 200$\times$ parameter scale, two architecture generations, and two initialization recipes --- satisfy $|{\mu}|/\sigma < 1\%$: neither checked model shows a material location offset at the checkpoint examined. For Qwen2.5-14B the step-0 state is not itself observed; the zero-mean premise of the anchor there rests on the documented Gaussian initialization recipe, and the terminal measurement establishes that training has not introduced a location offset material to the $|W|$ diagnostic.

\paragraph{Weibull fit on $|w|$ robustness.} To verify that fitting Weibull on $|w|$ (combined positive and negative weights) does not introduce asymmetry bias, we fit positive-$W$ and $|-W|$ separately on 6 representative Qwen2.5-14B matrices (layers 0, 24, 47 $\times$ $W_\text{down}$, $W_q$). The shape parameter difference $\Delta k$ between $+W$ and $|-W|$ fits is below 0.5\% relative across all 6 matrices, and $\Delta k$ between $|W|$ (combined) and $+W$ (positive only) is below 1\% relative. Exact zero counts are 0.0001\%--0.0002\% (fewer than 200 exact zeros per matrix), confirming $|w|$ histogramming has negligible numerical impact.

\paragraph{$\Gamma$ closure: full-moment extrapolation check.} For a Weibull$(k, \lambda)$ variate the full-support moments are $E|W| = \lambda \Gamma(1+1/k)$ and $E[W^2] = \lambda^2 \Gamma(1+2/k)$. Any fitted pair implies these moments; comparing them with the full-sample moments therefore measures how far the middle-80\% body fit extrapolates to moments of the full distribution, tails included --- exact agreement is not expected, and the magnitude of the residual is itself informative. Across the 900 FFN fits in our cohort, the first-moment relation holds to a median relative error of $4.0\%$, with 90\% of fits within $4.4\%$, and the deviation is systematically positive across the cohort --- the body fit overpredicts $E|W|$. The second-moment relation, which weights the extreme percentiles more heavily, deviates by a median of $15.7\%$. The check is thus an extrapolation-error bound for moments implied by the body fit, reported as measured; it is not a goodness-of-fit test (the latter is reported as $R^2$ in Table~\ref{tab:fit-quality}).

\paragraph{The anchor is protocol-specific, and the reading is protocol-stable.} The value $k_0 \approx 1.2054$ quoted above is specific to the middle-80\% convention. Table~\ref{tab:k0-protocol} gives the anchor for a range of trim fractions, each computed analytically by the same procedure. This matters for interpreting protocol sensitivity. The raw fitted shape parameter does move with the trim choice --- across the cohort the per-entry median $k_{80\%}-k_{100\%}$ is 0.052 (range of per-entry medians 0.050--0.054; individual matrices spread more widely) --- but almost all of that movement is anchor geometry rather than a change in what is being measured. Once each fit is referenced to its own anchor, the residual is an order of magnitude smaller: the median $d_{80\%}-d_{100\%}$ is -0.0021 (range $-0.0041$ to $-0.0008$). For in-band FFN matrices the anchored reading stays between -0.70\% and -0.52\% of the anchor across $P \in \{80\%, 90\%, 100\%\}$, with a per-matrix spread of about 0.60\%.

\begin{table}[t]
\centering
\small
\caption{\textbf{Protocol-specific initialization anchor $k_0(P)$ for the half-Normal marginal.} Each row applies the paper's probability-plot procedure to exact half-Normal quantiles over the central fraction $P$ (deterministic, no sampling; $2\times10^5$ grid points). The anchor is a property of the protocol rather than a tunable parameter: every trim choice has an analytically specified reference value, so a fitted $k_P$ is read as the offset $d_P = k_P - k_0(P)$. $^{\dagger}$The middle-80\% convention used throughout this paper.}
\label{tab:k0-protocol}
\begin{tabular}{cccc}
\toprule
Trim fraction $P$ & $k_0(P)$ & $\lambda_0(P)/\sigma_{\text{init}}$ & $R^2$ of probability-plot fit \\
\midrule
100\% & 1.1511 & 0.8748 & 0.9937 \\
95\% & 1.1793 & 0.8770 & 0.9958 \\
90\% & 1.1909 & 0.8809 & 0.9967 \\
85\% & 1.1991 & 0.8844 & 0.9973 \\
80\%$^{\dagger}$ & 1.2054 & 0.8875 & 0.9977 \\
70\% & 1.2149 & 0.8927 & 0.9984 \\
60\% & 1.2218 & 0.8969 & 0.9989 \\
50\% & 1.2269 & 0.9002 & 0.9993 \\
\bottomrule
\end{tabular}

\end{table}

\subsection{Comparison with Alternative Reference Families}
\label{sec:appendix-dist-compare}

We select the reference family empirically, by comparing the candidates in
common use for positive-valued data --- Weibull, gamma, and lognormal ---
under a protocol that matches the estimand used throughout this work. For each of the 1{,}220 weight matrices in this six-model battery (the cohort subset characterized in Table~\ref{tab:fit-quality}) we fit all three
families by truncated maximum likelihood restricted to the rank-$[10, 90]$ body
with the CDF renormalized to the truncation window (the same body targeted by
the middle-80\% protocol of Section~\ref{sec:framework}), and we score each fit in two ways:
(i) in-window truncated log-likelihood per observation, and (ii) held-out
quantile error --- the body fit is extrapolated to the quantiles excluded from
fitting, and we report $|\log_{10}(\text{predicted}/\text{observed})|$ at $q99$
and $q99.9$. The second criterion tests what the diagnostic actually uses:
whether the fitted body predicts the tail.

\paragraph{Body fit.} Table~\ref{tab:dist-body} reports the per-matrix win
shares. Weibull attains the best truncated likelihood in
$\geq 92\%$ of matrices in every component class outside the previously-flagged
Regime-B set and the separately-stored MHA $\Wq$/$\Wk$ group (treated
separately below): 100\% of training-state FFN matrices (475 matrices, after setting
aside the 9 flagged Regime-B matrices, which are reported separately below),
98\% of $\Wv$/$\Wo$ (304), 92\% of GQA $\Wq$/$\Wk$ (184), 93\% of the
terminal fused mixed $\Wqkv$ matrices (56; Section~\ref{sec:classes}), and 100\% of
initialization checkpoints (128, which include the 32 step-1 fused
$\Wqkv$ matrices). Lognormal is essentially never preferred in
the body. Gamma runs close to Weibull in-window (median likelihood gap
$\sim 0.2 \times 10^{-3}$ per observation); the two are separated by the tail
criterion below. The one partial exception is separately-stored MHA $\Wq$/$\Wk$
(64 matrices, 62\% Weibull / 34\% gamma) --- the most strongly drifted Selection
components, where no two-parameter family fits cleanly (consistent with
Section~\ref{sec:qk}'s reading of deep Selection drift as departure from the reference
family).

\begin{figure}[!t]
\centering
\includegraphics[width=0.95\linewidth]{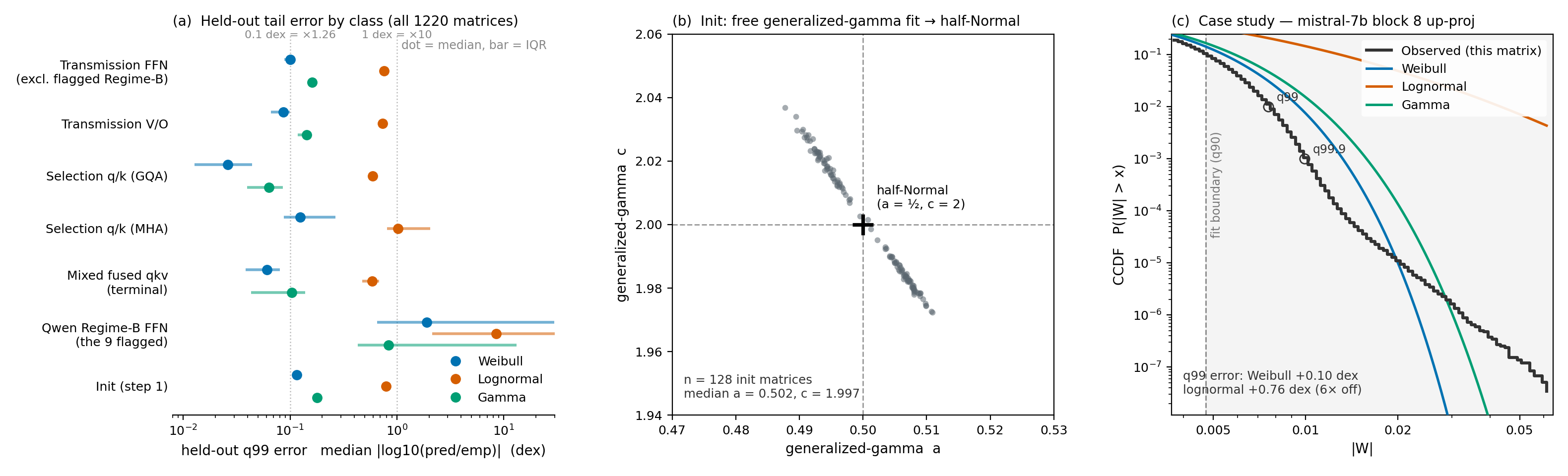}
\caption{\textbf{Reference-family comparison across 1{,}220 matrices (six-model subset of the cohort).}
(a)~Held-out $q99$ extrapolation error by component class (median and IQR,
log axis): Weibull (blue) leads in every class outside the flagged Regime-B
set; lognormal (orange) mispredicts $q99$ by $4$--$10\times$ throughout. (b)~Generalized
gamma fits to all 128 initialization matrices concentrate on the half-Normal
point $(a, c) = (0.5, 2)$ --- the population-level confirmation of the
initialization anchor. (c)~A single-matrix example (selected automatically as
the matrix whose Weibull and lognormal $q99$ errors are jointly closest to the
class medians in panel~(a)): observed CCDF versus the three body-fit
extrapolations; the shaded band beyond the $q90$ fit boundary is held out
from the fit; Weibull tracks the observed tail through $q99.9$, lognormal
overshoots by an order of magnitude, and all three families undershoot the
isolated super-weight beyond (cf.\ Section~\ref{sec:superweight}).}
\label{fig:dist-compare}
\end{figure}

\paragraph{Tail prediction.} Table~\ref{tab:dist-tail} and
Figure~\ref{fig:dist-compare}a report the held-out quantile errors. Weibull is
first in every class outside Regime-B: median $q99$ error 0.03--0.13~dex
($\sim$7--35\%), versus 0.6--1.0~dex for lognormal (its predicted $q99$ off by $4$--$10\times$) and
0.06--0.18~dex for gamma. All three families share a small
systematic \emph{positive} bias at $q99$: the ratio $\log_{10}(q_{\text{pred}}/q_{\text{emp}})$
has a positive median throughout, so the body fits \emph{overpredict} the
observed $q99$ (Weibull median $+0.100$~dex for FFN, $+0.087$ for $\Wv$/$\Wo$,
$+0.034$ for Selection). The sign is shared by all three families and is
consistent across component classes; we report it as an empirical regularity
of body-to-tail extrapolation under this protocol. At the extreme lower tail ($q1$) of MHA
Selection matrices, all three families fail comparably (1.8--2.3~dex) ---
near-zero weight accumulation that no two-parameter family captures.

\paragraph{Initialization anchor, population-level.} Fitting the more general
three-parameter generalized gamma family to the 128 initialization matrices
recovers the half-Normal special case: median $a = 0.502$ (IQR
$[0.494, 0.507]$), $c = 1.997$ ($[1.983, 2.019]$)
(Figure~\ref{fig:dist-compare}b), where $(a, c) = (0.5, 2)$ is half-Normal.
The initialization data thus independently select the distribution that the
initialization argument of Appendix~\ref{sec:appendix-init} derives; the
generalized-gamma fit is run on the initialization matrices only, and we do
not track the exponent $c$ through training. Within the generalized
gamma plane, Weibull is the two-parameter subfamily that keeps the
\emph{exponent} $c$ free --- the parameter that carries the
initialization tail ($c = 2$) and, in the sense probed by the comparison
below, the training-induced drift toward heavier tails --- whereas gamma
fixes $c = 1$ and frees only the power-law prefactor;
the anchor $(k_0, \lambda_0)$ of Appendix~\ref{sec:appendix-init} is the
protocol-determined Weibull coordinate of this half-Normal point.

\paragraph{Reading the comparison: why the exponent family wins.} This
geometry makes the outcome of the battery interpretable rather than
fortuitous. Initialization pins the shape at the $c = 2$ point, and a
two-parameter slice can track departures from that point only along its own
free axis. Weibull frees exactly the decay exponent, and wins wherever a
single family describes the data; gamma, frozen at exponential decay
($c = 1$), imitates the body within the fitting window (the small in-window
likelihood gaps of Table~\ref{tab:dist-body}) but is exposed by the held-out
tail criterion, which probes the decay regime itself; lognormal, outside the
plane with an asymptotically heavier tail than any stretched exponential,
fails both criteria. Had the training-induced deformation run along the
prefactor axis instead, the same battery would have favored gamma; that it
favors Weibull throughout the single-family regime --- and stops doing so
exactly where the framework independently diagnoses departure from any
single family (the deep-drifted MHA $\Wq$/$\Wk$ group and the Regime-B
mixtures) --- indicates that, at the resolution of this comparison, the
trained shapes are better captured along the exponent axis than along the
prefactor axis. The family choice is thus supported by which deformation
direction the fits favor, not by convenience.

\paragraph{Regime-B.} For the 9 flagged Regime-B matrices no family fits: the
truncated KS statistic rises to $\approx 0.02$ for all three (normal classes:
$\approx 0.004$), and tail errors reach 0.8--10~dex. The failure of every
single-family candidate is consistent with the secondary-population mixture structure
established in Appendix~\ref{sec:appendix:qwen-diagnostic} --- these matrices have left the single-family
regime altogether, which is precisely the condition the $k$-based screen is
designed to flag.

In summary, for the weight-matrix cohort studied here, and among the
two-parameter families compared, Weibull gives the best body fit and by far
the best body-to-tail extrapolation, and its initialization anchor
takes an analytic form under the measurement protocol
(Appendix~\ref{sec:appendix-init}). Within this cohort and comparison, the
reference-family choice is therefore not a convenience but the
best-supported choice among the three compared two-parameter families.

\begin{figure}[!t]
\centering
\includegraphics[width=0.95\linewidth]{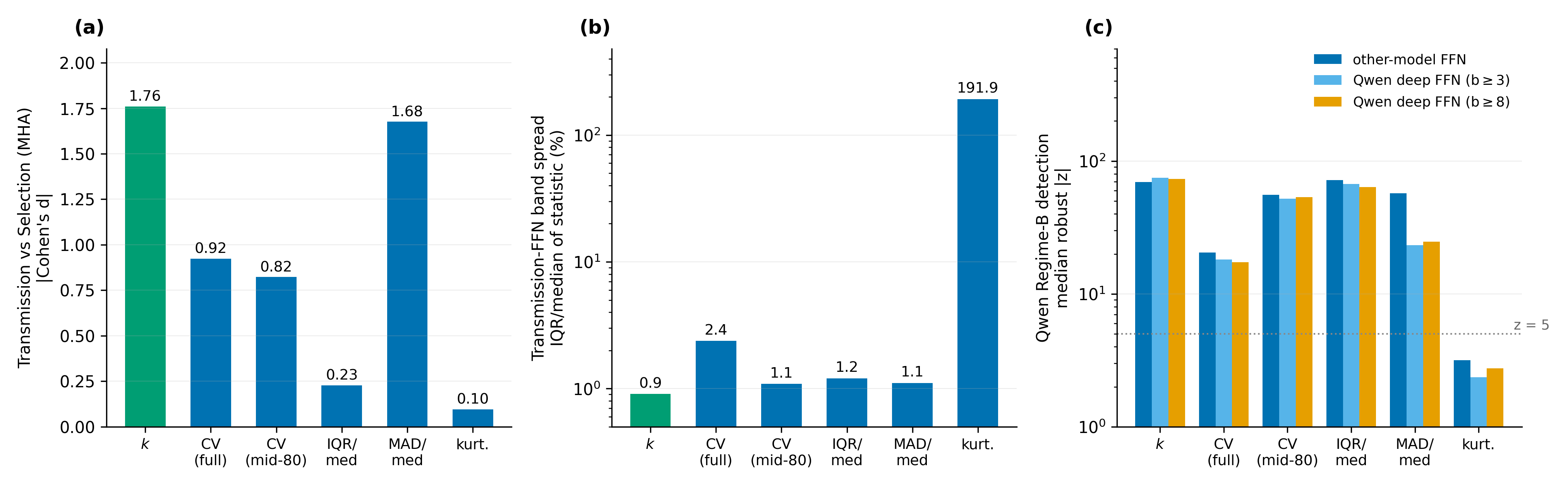}
\caption{\textbf{Fitted $k$ versus simple scale-free summaries}
(populations as in Table~\ref{tab:moment-compare}).
(a)~Transmission/Selection separation: $k$ leads ($|d| = 1.76$); kurtosis
separates nothing. (b)~Transmission-FFN band spread (log axis): $k$ forms the
tightest reference band (IQR/median $0.9\%$); kurtosis spreads over
$190\%$ and admits no band. (c)~Regime-B detection (log axis): median robust
$|z|$ against three reference populations; $k$ exceeds $|z| = 69$ under every
reference, while kurtosis stays below the $z = 5$ line --- the detection that
motivates the case study of Section~\ref{sec:within-entry-qwen} is invisible to the one summary that
carries genuinely non-body information.}
\label{fig:moment-compare}
\end{figure}

\paragraph{Comparison with moment-based summaries.}
A related question is what the fitted $k$ adds over simple scale-free
summaries of $|W|$. For an exact Weibull distribution, $k$ determines every
standardized moment, so strong correlation is expected --- and observed:
across the cohort, robust body summaries (trimmed CV, IQR/median, MAD/median)
track $k$ with Spearman $|\rho| = 0.86$--$0.996$. The question is therefore
not whether these summaries correlate with $k$, but which quantity expresses
the empirical findings most clearly. Table~\ref{tab:moment-compare} and
Figure~\ref{fig:moment-compare} compare them on the three uses this work
makes of $k$: class separation, reference-band tightness, and Regime-B
detection. $k$ is the only statistic in the leading tier on all three
simultaneously. The summaries that come close do so where they are monotone
transforms of the same body shape, while excess kurtosis --- the one summary
dominated by tail information --- separates the classes not at all, forms no
usable band, and misses the Regime-B detection entirely. Beyond these
comparisons, $k$ is read jointly with $\lambda$, $R^2$, and the
moment check from a single fit: one fitted object supplies shape, scale,
fit adequacy, and body-to-tail extrapolation together, with the analytic
initialization anchor $k_0(P)$ of Appendix~\ref{sec:appendix-init}.
Reference values under the half-Normal initialization target can equally be
computed for the simple summaries; the practical advantage of the fitted $k$
is this joint reading and its leading performance on all three tasks above,
not exclusive access to a calibrated anchor. These comparisons all concern the normalized body shape;
magnitude-bearing summaries such as RMS or a Frobenius norm estimate scale
--- the quantity carried by $\lambda$ --- and are not substitutes for $k$,
nor $k$ for them.

\begin{table}[t]
\centering
\small
\caption{Body fit under truncated maximum likelihood (rank-$[10,90]$ window,
CDF renormalized): per-matrix win shares and median per-observation truncated
negative-log-likelihood gaps relative to Weibull (positive $=$ Weibull better)
for lognormal (LN) and gamma (G). The 9 flagged Regime-B matrices are reported
separately, not discarded.}
\label{tab:dist-body}
\begin{tabular}{lcccc}
\toprule
Component class & $n$ & Win share W\,/\,LN\,/\,G (\%) &
$\Delta$NLL/obs $\times 10^{3}$ (LN$-$W, G$-$W) \\
\midrule
Transmission FFN (flagged set aside) & 475 & \textbf{100}\,/\,0\,/\,0 & $+1.92$, $+0.21$ \\
Transmission $\Wv$\,/\,$\Wo$ & 304 & \textbf{98}\,/\,0\,/\,2 & $+1.95$, $+0.20$ \\
Selection $\Wq$\,/\,$\Wk$ (GQA) & 184 & \textbf{92}\,/\,0\,/\,8 & $+1.78$, $+0.09$ \\
Selection $\Wq$\,/\,$\Wk$ (MHA) & 64 & \textbf{62}\,/\,3\,/\,34 & $+2.44$, $+0.15$ \\
Mixed fused $\Wqkv$ (terminal) & 56 & \textbf{93}\,/\,5\,/\,2 & $+2.05$, $+0.20$ \\
Regime-B FFN (flagged) & 9 & 0\,/\,11\,/\,\textbf{89} & $+0.14$, $-0.07$ \\
Initialization (step 1) & 128 & \textbf{100}\,/\,0\,/\,0 & $+1.90$, $+0.22$ \\
\bottomrule
\end{tabular}

\end{table}

\begin{table}[t]
\centering
\small
\caption{Held-out tail prediction: median $|\log_{10}(\text{predicted}/
\text{observed})|$ at $q99$ and $q99.9$, extrapolated from the body fit of each
family. Bold marks the best family per class. All quantiles shown are excluded
from the fitting window.}
\label{tab:dist-tail}
\begin{tabular}{lcc}
\toprule
Component class & $q99$: W\,/\,LN\,/\,G (dex) & $q99.9$: W\,/\,LN\,/\,G (dex) \\
\midrule
Transmission FFN (flagged set aside) & \textbf{0.100}\,/\,0.759\,/\,0.160 & \textbf{0.124}\,/\,1.023\,/\,0.201 \\
Transmission $\Wv$\,/\,$\Wo$ & \textbf{0.087}\,/\,0.735\,/\,0.144 & \textbf{0.105}\,/\,0.989\,/\,0.178 \\
Selection $\Wq$\,/\,$\Wk$ (GQA) & \textbf{0.026}\,/\,0.595\,/\,0.064 & \textbf{0.028}\,/\,0.810\,/\,0.068 \\
Selection $\Wq$\,/\,$\Wk$ (MHA) & \textbf{0.125}\,/\,1.026\,/\,0.163 & \textbf{0.147}\,/\,1.352\,/\,0.195 \\
Mixed fused $\Wqkv$ (terminal) & \textbf{0.061}\,/\,0.586\,/\,0.104 & \textbf{0.089}\,/\,0.815\,/\,0.130 \\
Regime-B FFN (flagged) & 1.896\,/\,8.478\,/\,\textbf{0.834} & 2.267\,/\,10.05\,/\,\textbf{0.951} \\
Initialization (step 1) & \textbf{0.115}\,/\,0.792\,/\,0.178 & \textbf{0.151}\,/\,1.070\,/\,0.232 \\
\bottomrule
\end{tabular}

\end{table}

\begin{table}[t]
\centering
\small
\caption{Fitted $k$ versus simple scale-free summaries on the three tasks the
framework uses $k$ for. \emph{Separation}: $|$Cohen's $d|$ between the 779
training-state Transmission matrices (the 9 flagged Regime-B matrices set
aside) and the 64 separately-stored MHA $\Wq/\Wk$ matrices. \emph{Band}:
per-matrix spread across the 475 training-state Transmission-FFN matrices
(coefficient of variation and IQR/median).
\emph{Detection}: median robust $|z|$ of the 9 flagged matrices against three
reference populations (non-Qwen FFN; Qwen FFN at blocks $\geq 3$; $\geq 8$);
the range spans the three references, and $k$ attains the highest worst-case
$|z|$. Bold marks the best statistic per column; $z < 5$ (kurtosis) fails
detection under any conventional threshold. Excess kurtosis is computed
from raw moments as $E[|W|^4]/E[|W|^2]^2 - 3$, i.e.\ the excess kurtosis
of the signed weights under the zero-mean convention, not a central-moment
kurtosis of $|W|$.}
\label{tab:moment-compare}
\begin{tabular}{lccc}
\toprule
Statistic & \shortstack{Class separation\\Transmission vs Selection\\(MHA), $|$Cohen's $d|$}
          & \shortstack{Transmission-FFN band\\per-matrix CV\,/\,IQR-med.\\(\%, $n{=}475$)}
          & \shortstack{Regime-B detection\\median robust $|z|$\\(range over 3 refs)} \\
\midrule
fitted $k$ (middle-80\%) & \textbf{1.76} & \textbf{0.61}\,/\,\textbf{0.91} & \textbf{69--75} \\
CV (full)                & 0.92          & 1.90\,/\,2.38                   & 17--20 \\
CV (middle-80\%)         & 0.82          & 0.79\,/\,1.09                   & 52--56 \\
IQR\,/\,median           & 0.23          & 0.89\,/\,1.21                   & 64--72 \\
MAD\,/\,median           & 1.68          & 1.04\,/\,1.11                   & 23--57 \\
excess kurtosis          & 0.10          & 231\,/\,192                     & 2.4--3.2 \\
\bottomrule
\end{tabular}

\end{table}

\begin{table}[t]
\centering
\small
\caption{\textbf{Weibull body fit quality by component group (1{,}220 matrices from a six-model subset of the cohort, middle-80\% protocol).} This battery is assembled from the raw-histogram re-fits of six cohort models (LLaMA-3-8B, Mistral-7B, OLMo-2-7B, Qwen2.5-7B, Pythia-410m, Pythia-6.9b; the Pythia-6.9b matrices are additionally re-fitted at step~1, which supplies an early-training reference and is not a terminal fit) and is therefore a different denominator from the $\Gamma$ closure check of Section~\ref{sec:ffn}, which is a full-moment extrapolation check run on the 900 FFN fits of the full cohort (not a goodness-of-fit test); within this table the FFN row is further split into 539 matrices outside the flagged region and the 9 flagged ones. $^{*}$The flagged Regime-B site, i.e.\ the Qwen2.5-7B shallow blocks that the $k$ screen flags. Fit quality is not uniform, and its structure is informative. Among Transmission FFN matrices outside the flagged region, essentially every fit is excellent. The FFN fits that do fall below $R^2 = 0.99$ are exactly the Qwen2.5-7B shallow blocks that the $k$ screen flags as Regime-B (Appendix~\ref{sec:appendix:qwen-diagnostic:channels}): a degraded body fit co-occurs with the shape departure rather than appearing at random. Separately-stored Selection projections show the widest spread. The poorest fit in this battery ($R^2 = 0.735$, $k_{80\%} = 0.281$) occurs at OLMo-2-7B, block~0, $\Wq$. The strongest shape departure among the separately-stored entries ($k_{80\%} = 0.249$, OLMo-2-7B, block~2, $\Wk$) is a different matrix and is fitted with $R^2 = 0.866$: a low $k$ does not by itself imply a poor fit. Entries outside this subset reach more extreme values still: OLMo-1-7B, not covered here, has minimum $R^2 = 0.718$ and minimum $k_{80\%} = 0.283$ among its Selection projections (Section~\ref{sec:qk}). The spread reported in this table is therefore conservative with respect to the full cohort. A reduced $R^2$ is therefore read as part of the departure signal, not as a failure of the protocol.}
\label{tab:fit-quality}
\begin{tabular}{lcccc}
\toprule
Group & $n$ & Median $R^2$ & Min $R^2$ & Fits with $R^2 < 0.99$ \\
\midrule
Transmission FFN (excl.\ flagged Qwen shallow) & 539 & 0.9980 & 0.9977 & 0 \\
Transmission FFN, Qwen2.5-7B blk 0--2$^{*}$ & 9 & 0.9914 & 0.9732 & 4 \\
Transmission $\Wv$/$\Wo$ & 336 & 0.9983 & 0.9894 & 1 \\
Selection $\Wq$/$\Wk$ (separate) & 248 & 0.9993 & 0.7352 & 33 \\
Mixed $\Wqkv$ (merged) & 88 & 0.9984 & 0.9690 & 3 \\
\bottomrule
\end{tabular}

\end{table}

\subsection{Initialization Reference for Pythia}
\label{sec:appendix-init-pythia}

Together, the anchor $(k_0, \lambda_0) \approx (1.205,\, 0.8875\,\sigma_{\text{init}})$ --- analytically specified and numerically evaluated --- provides a complete reference: both depend only on $\sigma_{\text{init}}$ and the fit protocol. The shape $k_0 \approx 1.20$ is independent of the initialization scale for any i.i.d.\ Gaussian initializer under this protocol; the scale $\lambda_0$ is initialization-scheme-specific and depends on which component of the model the weight matrix belongs to.

\paragraph{Component-specific $\sigma_{\text{init}}$ in Pythia.} Within Pythia, $\sigma_{\text{init}}$ varies by component because input-side and output-side projections use different initializers in the GPT-NeoX codebase \citep{black2022gptneox}. Two distinct recipes appear across the five sizes:

\textbf{Recipe A (70m / 160m / 410m / 1B).} Input-side projections ($\Wqkv$, $W_{\text{FFN\_in}}$) use \texttt{small\_init} with $\sigma_{\text{small\_init}} = \sqrt{2/(5d)}$ \citep{nguyen2019transformers}; output-side projections ($\Wo$, $W_{\text{FFN\_out}}$) use \texttt{wang\_init} with $\sigma_{\text{wang\_init}} = 2/(L\sqrt{d})$ \citep{black2022gptneox}. The input-to-output ratio simplifies to $L/\sqrt{10}$.

\textbf{Recipe B (6.9B).} Falls back to GPT-NeoX defaults: \texttt{normal} ($\sigma_{\text{in}} = 0.02$) input-side and \texttt{scaled\_normal} ($\sigma_{\text{out}} = 0.02/\sqrt{2L}$) output-side. The ratio simplifies to $\sqrt{2L}$.

\paragraph{Step-0 verification across 5 sizes.} Since the Weibull fit gives $\lambda = c(k) \cdot \sigma$ with a kind-independent constant $c(k)$ at fixed $k_0 \approx 1.20$, the measured $\lambda$ ratio at the step-0 checkpoint should match the predicted $\sigma$ ratio. Table~\ref{tab:lambda-init-verify} reports the verification across all 5 Pythia sizes.

\begin{table}[h]
\centering
\small
\caption{Component-specific $\lambda_{\text{init}}$ ratio verification across the Pythia family. ``Measured ratio'' is the mean of the two input-to-output pairs at the step-0 checkpoint: attention-side $\lambda_{\Wqkv}/\lambda_O$ and FFN-side $\lambda_{W_{\text{FFN\_in}}}/\lambda_{W_{\text{FFN\_out}}}$. All 5 sizes agree with the analytical prediction to within 0.13\%.}
\label{tab:lambda-init-verify}
\begin{tabular}{lcccccc}
\toprule
Size & $L$ & $d$ & Recipe & Predicted ratio & Measured ratio & Error \\
\midrule
70m  & 6  & 512  & A & $6/\sqrt{10}  = 1.897$ & 1.900 & $0.12\%$ \\
160m & 12 & 768  & A & $12/\sqrt{10} = 3.795$ & 3.798 & $0.07\%$ \\
410m & 24 & 1024 & A & $24/\sqrt{10} = 7.589$ & 7.589 & $0.00\%$ \\
1B   & 16 & 2048 & A & $16/\sqrt{10} = 5.060$ & 5.053 & $0.13\%$ \\
6.9B & 32 & 4096 & B & $\sqrt{64}   = 8.000$ & 8.000 & $0.00\%$ \\
\bottomrule
\end{tabular}

\end{table}

The two recipes produce distinct initial scaling laws: Recipe A scales the ratio linearly in $L$, Recipe B as $\sqrt{L}$. During training the input/output ratio collapses from these recipe-specific initial values toward $\sim 1.2\times$ at the terminal checkpoint, consistent with the component-paired growth documented in Section~\ref{sec:lambda} (Pearson $r = 0.9967$ between $\lambda_O$ and $\lambda_{\text{FFN\_out}}$).

\subsection{Candidate Mechanisms for Selection Evolution}
\label{sec:driving-forces}

Five mechanisms have been hypothesized in prior work that may contribute to the observed reference-relative $k$ departures of $\Wq/\Wk$. None of them is isolated by the present cohort, which is observational; the wording below is deliberately conditional:

\textbf{D1 --- Functional necessity.} \citet{elhage2021circuits} describe the QK circuit as determining the attention pattern. If forming selective attention patterns requires $\Wq$ and $\Wk$ to amplify some embedding directions and suppress others, the resulting configuration could present as heavier tails (lower $k$).

\textbf{D2 --- AdamW sign-descent dynamics.} Adam can behave similarly to sign descent with momentum \citep{kunstner2023noise}; we hypothesize that such element-wise sign-like updates may move weight elements away from zero, which could amplify a heavy-tail signal of the kind D1 describes. Adaptive optimizers are empirically implicated in activation-outlier formation \citep{kaul2025attention}; whether the same mechanism contributes to weight tails is untested here.

\textbf{D3 --- Softmax saturation feedback.} \citet{bondarenko2023quantizable} documented ``no-op'' attention heads associated with increasingly extreme softmax inputs; we hypothesize that the associated gradients could affect $\Wk$, a link not tested here.

Two further candidate forces --- D4: residual-stream coupling \citep{elhage2021circuits}; D5: cumulative training signal $T/\tau$ (in the Pythia entries, $k$ drift is descriptively associated with $T/\tau$; the ordering is not strict, see Section~\ref{sec:qk}) --- may modulate the above; neither has controlled-experiment isolation.

\subsection{K-Head Architecture Constraint}
\label{sec:appendix-khead}

In multi-head attention (MHA), each $(Q_i, K_i)$ pair is independently stored. In grouped-query attention (GQA) \citep{ainslie2023gqa}, a single $K_j$ head serves several query heads simultaneously (4-to-1 for LLaMA-3/Mistral and 7-to-1 for Qwen2.5-7B, per the released model configurations), which may constrain how selectively $\Wk$ can respond to any single query direction. This architectural constraint is consistent with the observed MHA/GQA dichotomy in $k$ values. The sharing argument concerns parameter dimensions only: it does not predict the magnitude or direction of the $k$ departure, and the observed MHA/GQA difference remains confounded with family and recipe (Section~\ref{sec:qk}).

\subsection{Weight Displacement During Training}
\label{sec:appendix-displacement}

\begin{figure}[!htbp]
\centering
\includegraphics[width=0.95\linewidth]{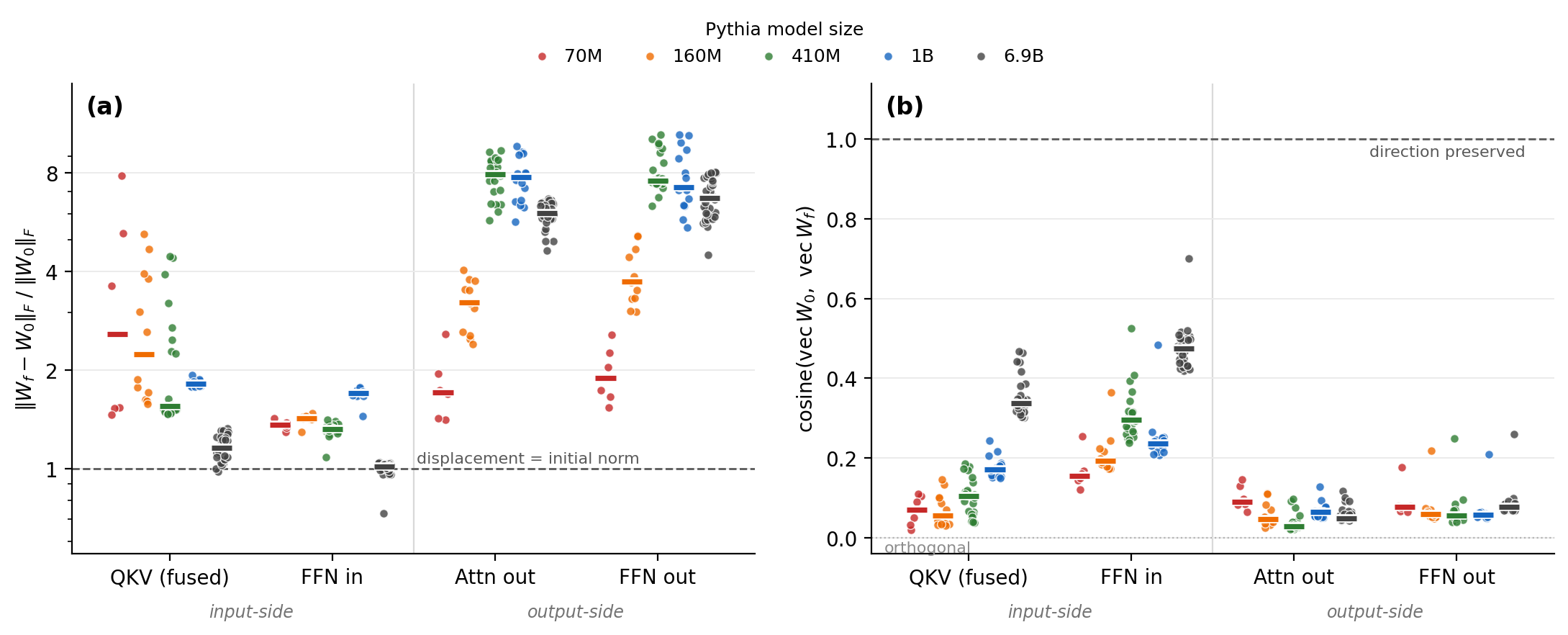}
\caption{\textbf{Weight displacement from initialization to the terminal checkpoint across five Pythia sizes (360 matrices).} (a)~Relative Frobenius displacement $\|\Delta W\|_F/\|W_0\|_F$ by stored-weight kind; values above 1 indicate displacement exceeding the initial norm. (b)~Cosine similarity between initial and terminal weights. Within each block, the five sub-columns are ordered by model size; jitter is applied only to separate overlapping points. White-outlined bars mark per-group medians. The Transmission-kind shape parameter stays in or near the reference band while the underlying weights are displaced far from initialization and largely reoriented; because the cosine is scale-invariant, the asymmetry is not a rescaling artifact; whether the initialization scale drives it dynamically remains untested.}
\label{fig:fr4-displacement}
\end{figure}

The invariance of the Transmission-class shape is macroscopic only. Comparing initial and terminal weights across the five Pythia sizes (360 matrices, step 0 versus step 143{,}000), the relative displacement $\|\Delta W\|_F / \|W_0\|_F$ has per-size medians of $1.0$--$7.9$, and the cosine similarity between initial and terminal weights has medians of only $0.03$--$0.47$: the matrices are displaced far from initialization in both magnitude and direction, while the Transmission-kind $k$ remains in or near the band at both endpoints (Figure~\ref{fig:fr4-displacement}). The displacement exceeds the initial norm in 347 of the 360 matrices. The shape stability is therefore not inherited from microscopic proximity to initialization: the measurements establish substantial matrix-level displacement and directional reorientation together with a stable distributional shape. Why training preserves this shape while displacing the weights this far is a question this paper does not answer; Appendix~\ref{sec:driving-forces} lists candidate mechanisms and labels them as hypotheses.

\begin{figure}[!htbp]
\centering
\includegraphics[width=0.95\linewidth]{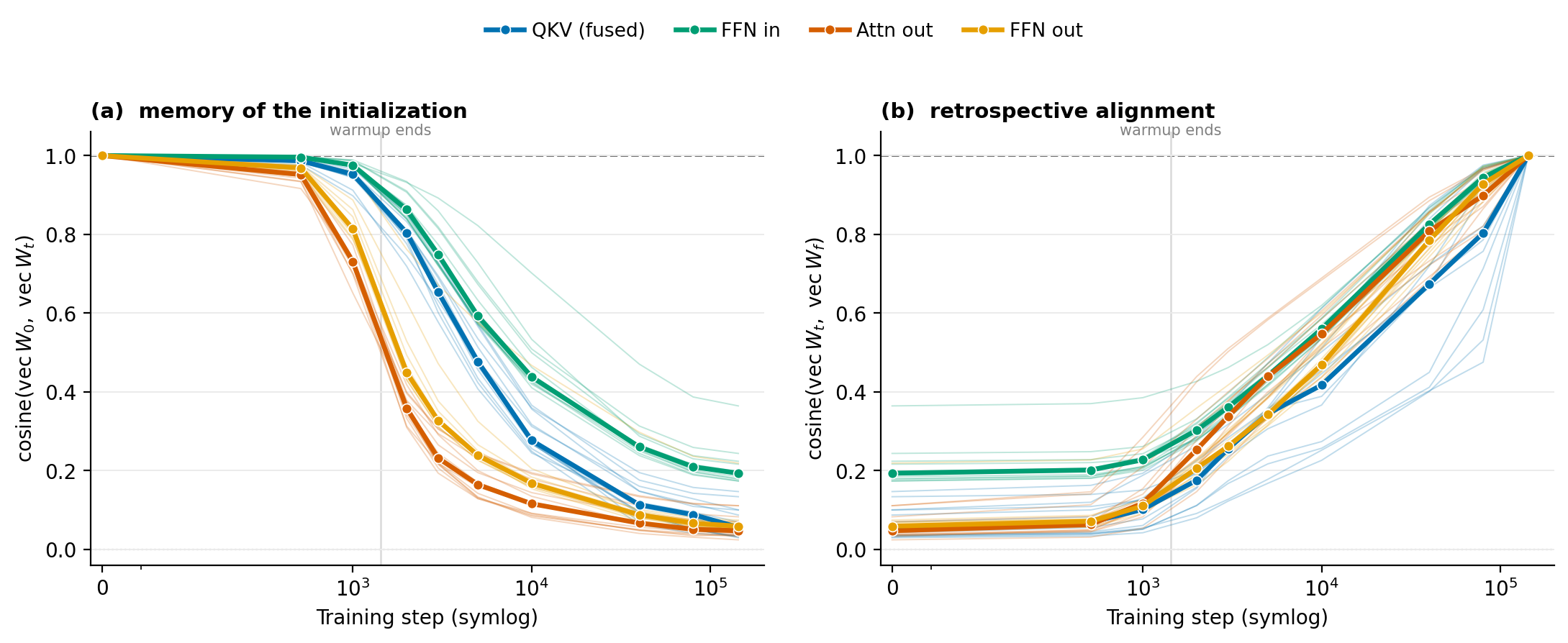}
\caption{\textbf{Time-resolved direction trajectory, Pythia-160m official checkpoints (10 checkpoints, step 0 to 143k; medians per kind over 12 blocks).} (a)~Cosine to the initial weights: memory of the initialization direction decays smoothly across the whole run, with no early one-shot scramble. (b)~Cosine to the final weights (retrospective alignment): directions keep reorganizing well into the annealing phase. The Transmission-kind median $k$ moves by only about $1\%$ across training (Figure~\ref{fig:f156}a), so the shape invariance is not a coincidence between two endpoint snapshots.}
\label{fig:fr6-trajectory}
\end{figure}

The reorientation is progressive rather than an endpoint effect. On the official Pythia-160m intermediate checkpoints (10 checkpoints from step 0 to step 143{,}000; medians per kind over 12 blocks), the cosine to the initial weights decays smoothly to $0.05$--$0.19$ for the Transmission kinds, and the cosine to the final weights is still approximately $0.80$--$0.94$ at 56\% of training (Figure~\ref{fig:fr6-trajectory}); the Transmission-kind median $k$ moves by only about $1\%$ across training (extracted from the $k$ trajectories of Figure~\ref{fig:f156}a). Over the ten checkpoints the retrospective alignment rises monotonically in 48 of 48 matrices, and the cosine to initialization falls monotonically in 47 of 48.

The input/output-side asymmetry is consistent across scale: at the endpoint, the output-side projections are nearly orthogonal to initialization at every size (median cosine $0.03$--$0.09$ across all five Pythia models), while input-side retention is generally larger in the larger models, with local exceptions; the 160m trajectory shows the same asymmetry emerging early in training, with the output-side kinds decorrelating first. The low cosine shows that the change is not a rescaling of the initial matrix; whether the initialization scale also drives this asymmetry through the optimizer's relative step size remains to be tested.

\FloatBarrier
\section{Experiment Setup}
\label{sec:appendix-experiments}

Table~\ref{tab:experiment-summary} states, for each experiment reported in the paper, what it targets, the cohort it uses, and its key takeaway. Grouping thresholds are listed in the final column; all of them are \emph{post hoc illustrative bins} used for exposition, not quantities derived from theory. Analysis-internal thresholds (such as the channel valley threshold and the $90\%$ criterion of Appendix~\ref{sec:appendix:qwen-diagnostic:channels}) are stated where used. The entry-level composition of every cohort referenced below --- sizes, corpora, and checkpoints --- is specified once in Table~\ref{tab:cohort-setup} and not repeated here.

\begin{table}[!htbp]
\centering
\footnotesize
\renewcommand{\arraystretch}{1.0}
\caption{\textbf{Summary of experiments: target, cohort, and key takeaway.} Cohort composition (sizes, corpora, checkpoints): Table~\ref{tab:cohort-setup}. $^{\P}$The summaries compared are trimmed CV, IQR/median, MAD/median, and excess kurtosis (Table~\ref{tab:moment-compare}, Figure~\ref{fig:moment-compare}). $^{\ddagger}$The Qwen regime split is insensitive to its threshold: the two groups are separated by a $0.50$-wide gap in shallow-minimum $k$ (Regime A lowest $1.12$, Regime B highest $0.62$), so \emph{any} threshold in $[0.62, 1.12]$ produces the identical partition. The $T/\tau$ state bins (Saturated / Near-saturated / Approaching / Transition) are descriptive labels used for grouping and exposition; the scaling analysis of Section~\ref{sec:lambda-scaling} uses the continuous $T/\tau$ value, not the bins.}
\label{tab:experiment-summary}
\begin{tabular}{p{2.2cm}p{0.6cm}p{3.4cm}p{2.0cm}p{3.5cm}p{1.9cm}}
\toprule
Experiment & \S & Target & Cohort & Key takeaway & Grouping thresholds \\
\midrule
$k$ band &
\ref{sec:ffn} &
Width of the terminal Transmission-class $k$ band across architectures &
12 entries, 7 families &
Median terminal $k \in [1.186, 1.204]$; CV $=0.51\%$; leave-one-family-out $[0.47\%, 0.57\%]$ &
--- \\

Body--tail ablation &
\ref{sec:bodytail} &
Re-fit under three trim protocols to test the protocol dependence of the fitted $k$ &
12 entries &
12/12 in band at 80\%, 0/12 untrimmed; gap $=$ the analytic anchor shift (protocol geometry) &
trim $\in \{80, 90, 100\}\%$ \\

Super-weight observation &
\ref{sec:superweight} &
Quantify isolated extreme weights and test whether the fitted body survives their presence &
8 entries with per-block access &
Body stays Weibull-conforming outside the flagged Regime-B blocks; a few extreme elements per matrix detach &
$\max/q_{99}$ \\

Anomaly detection case study &
\ref{sec:within-entry-qwen}, \ref{sec:appendix:qwen-diagnostic} &
Trace a cohort-level $k$ alarm to blocks and the responsible raw-weight structure &
11-entry Qwen cohort &
Alarm $\rightarrow$ shallow blocks $\rightarrow$ 5{,}031 channels ($26.6\%$) suppressed across all three FFN projections, $9.4\times$ enrichment &
Regime A $\geq 1.0$, B $\leq 0.7$$^{\ddagger}$ \\

Selection drift &
\ref{sec:qk} &
Characterize $\Wq$/$\Wk$ departure along depth, training budget, and attention storage architecture &
12 entries &
Separately-stored MHA deepest, GQA milder, merged transitional --- observational; confounded with family and recipe &
--- \\

$\lambda$ evolution and scaling &
\ref{sec:lambda} &
Track terminal $\lambda$ and its trajectories against the optimizer control variables &
Pythia, 5 sizes &
Terminal $\lambda$ is directionally consistent with $\sqrt{\eta/\lamwd}$ ($r=0.94$, per-size deviations $7$--$36\%$); trajectories phase-separate by $T\cdot\eta\cdot\lamwd$ &
$T/\tau$ state bins \\

Initialization anchor verification &
\ref{sec:appendix-init}, \ref{sec:appendix-init-pythia} &
Verify the analytic anchor $(k_0, \lambda_0)$ directly on step-0 checkpoints &
Pythia, 5 sizes, step 0; 128 init matrices &
Per-block step-0 $k = 1.205 \pm 0.001$; recipe $\lambda$ ratio within $0.13\%$; free generalized-gamma fit lands on the half-Normal point &
--- \\

Reference-family comparison &
\ref{sec:appendix-dist-compare} &
Compare Weibull with gamma and lognormal under the paper fitting protocol &
1{,}220 matrices (six-model subset) &
Weibull wins the body fit in $\geq 92\%$ of every class except Regime-B and separate-MHA $\Wq$/$\Wk$ &
--- \\

Moment-summary comparison &
\ref{sec:appendix-dist-compare} &
Test what the fitted $k$ adds over simple scale-free summaries of $|W|$$^{\P}$ &
1{,}220 matrices (six-model subset) &
$k$ alone is in the leading tier on all three uses: class separation, band tightness, Regime-B detection &
--- \\

Weight displacement &
\ref{sec:appendix-displacement} &
How far weights move between initialization and the terminal checkpoint &
5 sizes, 360 matrices &
Displacement exceeds the initial norm in 347/360 matrices and cosine to initialization falls to $0.03$--$0.47$, while Transmission-kind $k$ stays in or near the band &
--- \\

Direction trajectory &
\ref{sec:appendix-displacement} &
Time-resolve the displacement: cosine to initial and final weights across checkpoints &
Pythia-160m, 10 checkpoints &
Directional memory decays smoothly (no early one-shot scramble); reorganization continues into annealing &
--- \\
\bottomrule
\end{tabular}

\end{table}

\section{Software and Data}
\label{sec:appendix-api}

\subsection{\texttt{npm-weibull-py v0.4}}

The released \texttt{npm-weibull-py} library exports seven core diagnostics for fitting and benchmarking Weibull parameters on transformer weight matrices:

\begin{table}[h]
\centering
\small
\caption{The seven exported core diagnostics of \texttt{npm-weibull-py v0.4}, retaining the historical F-numbering. F4 and F7 were merged before the public release; F6e denotes the distribution-free extension of F6.}
\label{tab:api}
\begin{tabular}{llp{0.50\linewidth}}
\toprule
ID & Function & Description \\
\midrule
F1 & \texttt{weibull\_fit} & Fit Weibull $(k, \lambda)$ with $R^2$ and KS from a histogram \\
F2 & \texttt{classify\_k} & Classify $k$ into transmission / mild / strong / init regimes \\
F3 & \texttt{sigma\_decompose} & $\sigma$ growth attribution: $\lambda$-dominated vs.\ $k$-dominated vs.\ mixed \\
F5 & \texttt{k\_drift\_severity} & Severity label for $k_{\text{init}} \rightarrow k_{\text{trained}}$ drift \\
F6 & \texttt{compute\_T\_tau} & Wang--Aitchison $\tau_{\text{iter}}$ cycle ratio and physical state \\
F6e & \texttt{per\_block\_metrics} & Distribution-free heaviness (Q90/Q10, P999/P50, Gini) \\
F8 & \texttt{classify\_attention\_arch} & MHA / GQA / MQA classifier from head counts \\
\bottomrule
\end{tabular}

\end{table}

The library is pip-installable; source is available in the \href{https://github.com/tiexinding/NPM-Weibull-public}{public repository}.

\subsection{DATABASE\_v9\_1}

The companion benchmark database covers the 12 cohort entries of Table~\ref{tab:cohort-setup} and is released in two layers:

\begin{itemize}
\item \textbf{Summary tables} (public repository): \texttt{DATABASE\_v9\_1} proper --- one row per cohort entry, with per-kind medians of $(k, \lambda, R^2)$ and sub-threshold counts for every stored-weight kind present in the entry ($\Wq$, $\Wk$, $\Wv$, $\Wqkv$, $\Wo$, $\Wgate$, $\Wup$, $\Wdown$, $W_{\text{FFN\_in}}$, $W_{\text{FFN\_out}}$), alongside the training metadata used in Section~5 ($\eta_{\text{peak}}$, $\lamwd$, $T/\tau$, Physical State); provided as CSV and Markdown with the generating scripts, together with the 11-entry Qwen-cohort table of Appendix~\ref{sec:appendix:qwen-diagnostic}.
\item \textbf{Per-matrix fits} (supplementary material): JSON, one file per model/checkpoint, covering the five Pythia sizes at step~0 and step~143{,}000, OLMo-1, and the 11-entry Qwen cohort. Each record carries the stored parameter name and kind with $k_{80\%}$, $k_{90\%}$, $k_{100\%}$, $\lambda$, $R^2$, $\langle|W|\rangle$, $\operatorname{sd}(|W|)$, and the first-moment ratio $\lambda/\langle|W|\rangle$ that feeds the $\Gamma$ closure check of Section~\ref{sec:ffn}. Per-matrix results for the remaining entries (OLMo-2, LLaMA-3, Mistral) are contained in the reference-family battery data of Appendix~\ref{sec:appendix-dist-compare}, shipped in the same supplement.
\end{itemize}

The \href{https://github.com/tiexinding/NPM-Weibull-public/tree/main/database_v9_1}{database directory} provides the summary tables and generating scripts; per-matrix fit records accompany the paper in the supplementary material.

\paragraph{Model sources.} All weights analyzed in this work were extracted from publicly released open-weight checkpoints on the Hugging Face Hub under their original licenses:
\begin{itemize}
\item Pythia 70m/160m/410m/1B/6.9B: \texttt{EleutherAI/pythia-\{70m,160m,410m,1b,6.9b\}}
\item OLMo-1-7B: \texttt{allenai/OLMo-7B}
\item OLMo-2-7B: \texttt{allenai/OLMo-2-1124-7B}
\item LLaMA-3-8B: \texttt{meta-llama/Meta-Llama-3-8B}
\item Mistral-7B: \texttt{mistralai/Mistral-7B-v0.1}
\item Qwen2.5-7B/14B: \texttt{Qwen/Qwen2.5-7B}, \texttt{Qwen/Qwen2.5-14B}
\item Qwen3-8B: \texttt{Qwen/Qwen3-8B}
\end{itemize}
Pythia releases all training checkpoints from step 0 to step 143000; we use 14 log-spaced revisions for trajectory analyses and the terminal checkpoint for cross-family comparisons. No model was retrained or fine-tuned in this study; all analyses are run on the released weights without modification.

\section{Diagnostic Framework Application Cases}
\label{sec:appendix:application-cases}

This appendix documents the Qwen-family shallow FFN case, expanding the within-entry deviation flagged in \S\ref{sec:within-entry-qwen} into a diagnostic chain from cohort screening to raw-weight channel verification.

\label{sec:appendix:qwen-diagnostic}

\subsection{Cohort scope}
\label{sec:appendix:qwen-diagnostic:scope}

This appendix expands the diagnostic case study (\S\ref{sec:within-entry-qwen}). The 11-entry Qwen cohort comprises the three Qwen entries from our main cohort plus eight additional Qwen-family entries (Qwen2-1.5B/7B, Qwen2.5-1.5B/3B, four base--Math-CPT pairs), spanning three generations (Qwen2, Qwen2.5, Qwen3), five sizes (1.5B/3B/7B/8B/14B), and three depths (28L/36L/48L); the full enumeration is in Table~\ref{tab:qwen-cohort}.

\subsection{Anomaly detection and regime partition}
\label{sec:appendix:qwen-diagnostic:detection}

We classify each entry by the minimum shallow-block $k$ across $W_{\text{gate}}, W_{\text{up}}, W_{\text{down}}$ ($\ell \in [0, 5]$): \emph{Regime A (clean)} if $\geq 1.0$, \emph{Regime B (bimodal)} if $\leq 0.7$; ``bimodal'' here denotes the presence of a secondary small-magnitude population --- an explicitly separated second mode is resolved in 4 of the 9 flagged matrices, the rest showing an unseparated shoulder (Appendix~\ref{sec:appendix:qwen-diagnostic:channels}). The partition is clean: 4 Regime A entries (all 1.5B variants, independent of Math-CPT) and 7 Regime B entries (all 3B/7B/8B/14B; Table~\ref{tab:qwen-cohort}). The same protocol applied to the 9 non-Qwen entries in our main cohort (Pythia $\times 5$, OLMo-1, OLMo-2, LLaMA-3, Mistral; for the 2-matrix GeLU Pythia entries the available FFN components $W_{\text{FFN\_in}}, W_{\text{FFN\_out}}$ substitute for the SwiGLU triple) yields zero Regime B detections; the deviation is Qwen-family-specific within the cohort surveyed.

\begin{table}[h]
\centering
\footnotesize
\setlength{\tabcolsep}{3pt}
\caption{11-entry Qwen-family cohort: shallow vs deep block $k$ medians, regime classification. Shal.\ = shallow (blocks 0--5); Deep = blocks 10 onwards; $k_\text{g} \equiv k$ for $W_\text{gate}$. ``Shal.\ min $k$'' is the minimum block-level $k$ across $W_\text{gate}, W_\text{up}, W_\text{down}$ within shallow blocks. Tokens: base = pretraining; ``+ Math CPT'' / ``+ 200B Math'' = math-domain continued pretraining. Regime A: shallow min $k \geq 1.0$ (clean). Regime B: shallow min $k \leq 0.7$ (bimodal).}
\label{tab:qwen-cohort}
\begin{tabular}{lrrlrrlc}
\toprule
Entry & Layers & Params & Tokens & Shal.\ med.\ $k_\text{g}$ & Shal.\ min $k$ & Deep med.\ $k_\text{g}$ & Regime \\
\midrule
Qwen2-1.5B-base    & 28 & 1.5B & 7T               & 1.19 & 1.18 & 1.18 & A \\
Qwen2.5-1.5B       & 28 & 1.5B & 18T              & 1.18 & 1.17 & 1.17 & A \\
Qwen2-Math-1.5B    & 28 & 1.5B & 7T + Math CPT    & 1.19 & 1.19 & 1.19 & A \\
Qwen2.5-Math-1.5B  & 28 & 1.5B & 18T + Math CPT   & 1.19 & 1.12 & 1.19 & A \\
Qwen2-7B           & 28 & 7B   & 7T               & 0.96 & 0.53 & 1.19 & B \\
Qwen2.5-7B         & 28 & 7B   & 18T              & 0.97 & 0.57 & 1.19 & B \\
Qwen2-Math-7B      & 28 & 7B   & 7T + 200B Math   & 0.96 & 0.52 & 1.20 & B \\
Qwen2.5-Math-7B    & 28 & 7B   & 18T + 700B Math  & 1.01 & 0.62 & 1.20 & B \\
Qwen2.5-3B         & 36 & 3B   & 18T              & 0.65 & 0.57 & 1.18 & B \\
Qwen3-8B           & 36 & 8B   & 36T              & 0.69 & 0.34 & 1.19 & B \\
Qwen2.5-14B        & 48 & 14B  & 18T              & 0.44 & 0.40 & 1.19 & B \\
\bottomrule
\end{tabular}

\end{table}

\subsection{Family-wide phenomenology}
\label{sec:appendix:qwen-diagnostic:phenomenology}

\begin{figure}[!t]
\centering
\includegraphics[width=0.88\linewidth]{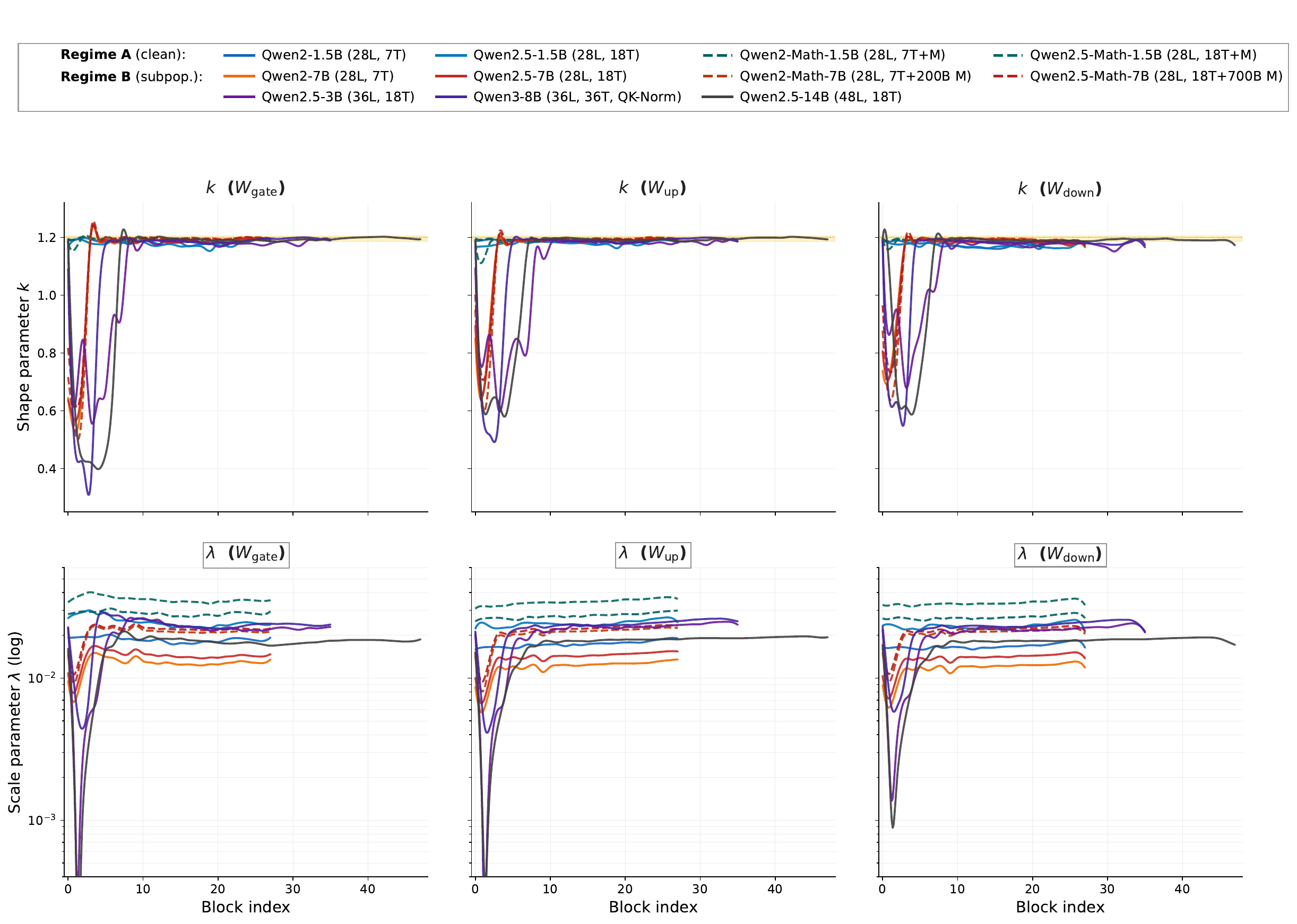}
\caption{Per-block FFN shape parameter $k$ (top row) and scale parameter $\lambda$ (bottom row) across the 11-entry Qwen-family cohort, for the three SwiGLU FFN sub-components $W_\text{gate}$, $W_\text{up}$, $W_\text{down}$. The amber shaded band in the $k$ panels marks the Transmission band $k \in [1.186, 1.204]$ established in Section~\ref{sec:ffn} across the 12-entry main cohort. \textbf{Regime A} (1.5B class, $N{=}4$, cool blue/teal lines): every block remains near the Transmission band, with none of the Regime-B shallow collapse; $\lambda$ profiles are comparatively smooth across depth. \textbf{Regime B} (3B+ class, $N{=}7$, warm red/orange/purple/gray lines): shallow blocks ($\ell \leq 6$; the exact affected range varies by entry) show paired $k$ drop and $\lambda$ collapse across all three FFN sub-components, while deep blocks ($\ell \geq 10$) return to near the Transmission band. The Regime A/B partition aligns with model parameter count in this cohort (size is confounded with generation, depth, and recipe) and is independent of base vs.\ Math-CPT distinction (Table~\ref{tab:qwen-cohort}).}
\label{fig:qwen-perblock-k}
\end{figure}

\begin{figure}[!t]
\centering
\includegraphics[width=0.78\linewidth]{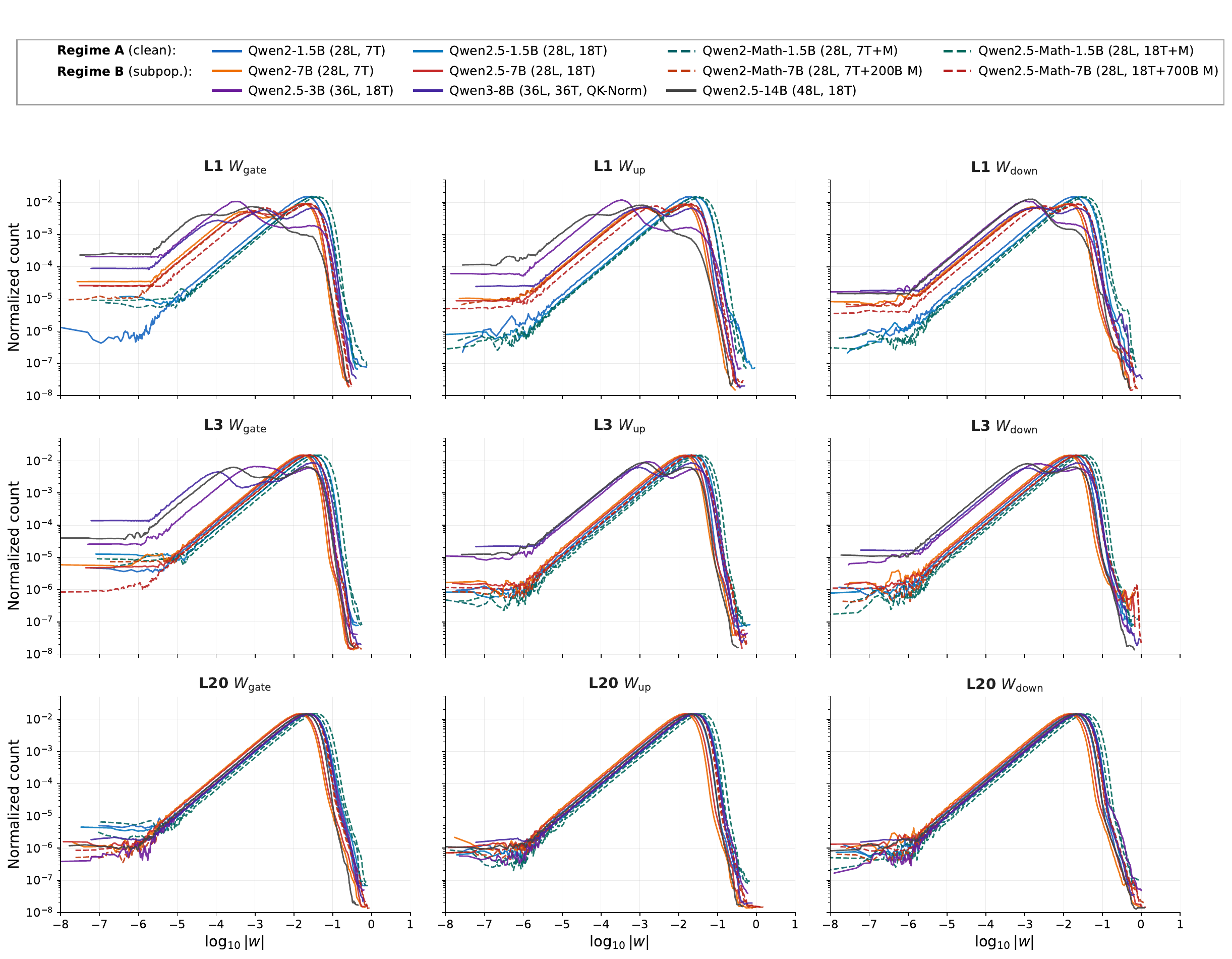}
\caption{FFN weight magnitude $\log_{10}|w|$ distributions across the 11-entry Qwen-family cohort at three representative layer depths (rows: L1 shallow, L3 shallow, L20 deep) and three SwiGLU sub-components (columns: $W_\text{gate}$, $W_\text{up}$, $W_\text{down}$). \textbf{L1 / L3 rows (shallow):} Regime A entries (cool blue/teal, 4 entries, all 1.5B) show a single unimodal main body at $\log_{10}|w| \approx -2$; Regime B entries (warm colors, 7 entries, 3B+) show the main body plus a secondary small-magnitude population extending into $\log_{10}|w| \in [-7, -3]$. \textbf{L20 row (deep):} all 11 entries collapse onto a single unimodal main body --- the regime-B secondary population is absent. This demonstrates the secondary-population signature is spatially localized to shallow blocks rather than a global property of the Regime B entries.}
\label{fig:qwen-multilayer-hist}
\end{figure}

Figure~\ref{fig:qwen-perblock-k} decomposes the cohort along the block axis; Figure~\ref{fig:qwen-multilayer-hist} drills down into representative blocks. The two views together characterize where the deviation occurs and what it looks like at the weight level. The block-axis view exposes spatial localization: in all 7 Regime B entries, the $k$ drop and paired $\lambda$ collapse are confined to shallow blocks ($\ell \leq 6$; blocks 0--2 for Qwen2.5-7B, blocks 1--6 for Qwen2.5-14B) and appear simultaneously in all three SwiGLU sub-components ($W_{\text{gate}}, W_{\text{up}}, W_{\text{down}}$); deep blocks ($\ell \geq 10$) return to near the Transmission band. The three sub-components are affected co-locally at the per-block level, which disfavors a single-component anomaly story.

The drill-down view (Figure~\ref{fig:qwen-multilayer-hist}) reveals the microstructure underlying the per-block $k$ drop: in shallow blocks, Regime B entries show a main body at $\log_{10}|w| \approx -2$ plus a secondary small-magnitude population extending into $\log_{10}|w| \in [-7, -3]$. In deep blocks, all 11 entries --- including those with severe shallow drift --- collapse onto a unimodal main body, confirming the deviation is a genuine secondary-population microstructure spatially confined to shallow blocks rather than a global property of the Regime B entries.

This secondary-population microstructure is structurally distinct from the super-weight phenomenon \citep{yu2024superweight}, which describes a small number of large-magnitude outliers. Per-entry $\max/q_{99}$ ratios (Table~\ref{tab:superweight}) show the two Regime-B entries covered there --- Qwen2.5-14B (median $17.3\times$, extreme $23.7\times$) and Qwen3-8B (median $17.3\times$, extreme $40.4\times$) --- sitting well below the OLMo-1 SwiGLU control (median $27.7\times$, extreme $107.2\times$), which nevertheless maintains a single-mode body. By this max/q99 metric, the Qwen signature is a body-level secondary population (its mass centered near $|w| \approx 10^{-3}$, with excess relative to in-band references extending down to $10^{-5}$) rather than a tail-level outlier --- a distinction the diagnostic framework surfaces directly through the $k$ value rather than through ad-hoc outlier counting.

\subsection{Related observations and deferred mechanism}
\label{sec:appendix:qwen-diagnostic:future}

We situate the finding against related published observations in adjacent modalities; the evidence that the flagged structure is not a measurement artifact comes from the raw-weight channel verification of Appendix~\ref{sec:appendix:qwen-diagnostic:channels}. \citet{wong2025secondarysinks} reported secondary attention sinks only in the Qwen family among the model families they tested (an activation-side match) and documented emergence following math-mid-training (Qwen2 $\to$ Qwen2-Math), noting that LLaMA-3.1, Phi-4, Mathstral, and CodeLlama do not exhibit the signature. \citet{zhang2026domlora} reported localized Projected Adapter Gradient Energy (PAGE) peaks at shallow FFN down-projection layers across multiple model families. \citet{voita2023dead} documented dead neurons in the first half of large language models. Each of these observations is adjacent to the present finding in one respect --- family (Qwen-specific sinks), location (shallow FFN), or modality (early-layer dead neurons) --- but none covers the same family, location, and time-scale simultaneously; they are related observations, not independent replications of the same structure. Disentangling the causal mechanism requires controlled experiments --- observational cohort analysis cannot isolate causes --- and is reserved for a focused follow-up study (e.g., matched-architecture pretraining with vs.\ without late-stage recipe shocks).

\subsection{Channel-level verification on raw weights}
\label{sec:appendix:qwen-diagnostic:channels}

\begin{figure}[!htbp]
\centering
\includegraphics[width=0.9\linewidth]{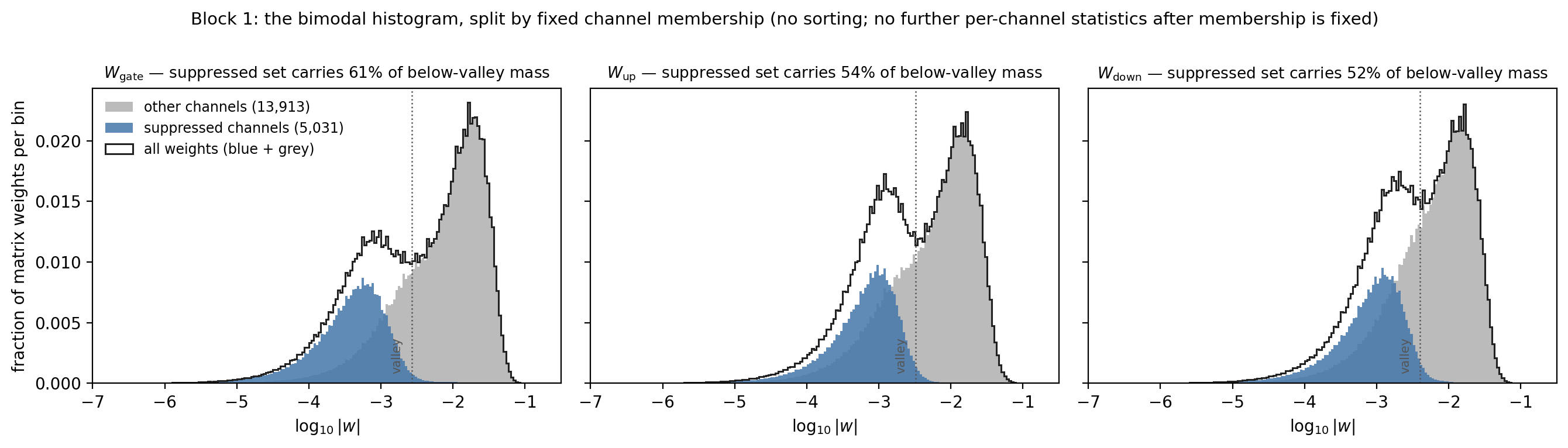}
\caption{\textbf{The secondary population decomposed by channel membership (Qwen2.5-7B, block~1, raw \texttt{bfloat16}).} The channel set is first identified by a statistical criterion (per-channel below-valley fraction $\geq 90\%$ in each of the three projections); with that membership then held fixed, each matrix's raw $\log_{10}|w|$ histogram is split by membership alone, with no further reordering and no per-channel statistic. Blue: all weights of the suppressed channels, forming a single population centered near $10^{-3}$. Grey: all remaining weights, forming the main body near $10^{-2}$. Their sum is the observed bimodal histogram of this block --- the Qwen2.5-7B curve in the L1 row of Figure~\ref{fig:qwen-multilayer-hist} --- so the mixture is reproduced by channel membership alone. The suppressed set is the dominant carrier of the below-valley mass ($52$--$61\%$), with the remainder from partially suppressed channels and the overlap tail of the main body.}
\label{fig:c5-decomposition}
\end{figure}

The preceding subsections characterized the Regime-B signature through fitted parameters and histograms. This subsection verifies the underlying microstructure directly on raw weight tensors, asking what the secondary small-magnitude population physically consists of. We fetch 15 tensors from the released Qwen2.5-7B checkpoint in their stored \texttt{bfloat16} precision (blocks 0--2, flagged; blocks 14 and 21, in-band controls; $W_{\text{gate}}$, $W_{\text{up}}$, $W_{\text{down}}$ each) and compute the channel statistics directly on the raw stored values, with no trimming and no mean subtraction; histograms enter only for display and for the decomposition of Figure~\ref{fig:c5-decomposition}.

\paragraph{Channel localization.} For each matrix we place a threshold at the valley of its $|w|$ distribution where a separated secondary mode exists ($2.7$--$4.0\times10^{-3}$ for the block-1 matrices; the secondary population is centered near $|w|\approx10^{-3}$, roughly an order of magnitude below the main body at $|w|\approx10^{-2}$), and at a fixed $10^{-4}$ for matrices without a separated mode, including all controls; we then measure, for every intermediate channel, the fraction of that channel's weights falling below the threshold. In control blocks, no channel exceeds $3.3\%$ below-threshold on either axis. In block~1, entire channels fall below threshold: the most extreme single channel reaches $100\%$ ($W_{\text{up}}$ rows), $99.5\%$ ($W_{\text{gate}}$ rows), and $100\%$ ($W_{\text{down}}$ columns). Across all nine flagged matrices, the below-threshold mass concentrates on few channels: the participation ratio over intermediate channels drops to $0.29$--$0.61$ for $W_{\text{gate}}/W_{\text{up}}$ rows (controls: $0.92$--$0.94$) and $0.39$--$0.66$ for $W_{\text{down}}$ columns (controls: $0.93$--$0.94$).

\paragraph{Cross-projection alignment.} The affected channels coincide across all three projections of the same block. In block~1, $5{,}341$ gate rows, $6{,}636$ up rows, and $5{,}402$ down columns are $\geq 90\%$ below their matrix's valley threshold; $5{,}031$ channel indices --- $26.6\%$ of the $18{,}944$ intermediate channels --- appear in all three sets simultaneously (gate--down pairwise overlap: $97.9\%$ of the smaller set, versus $\sim 530$ channels expected under independence, a $9.4\times$ enrichment). For roughly one quarter of the intermediate neurons in this block, the entire fan-in (rows of $W_{\text{gate}}$, $W_{\text{up}}$) and fan-out (columns of $W_{\text{down}}$) sit an order of magnitude below the main body. Splitting each matrix's raw histogram by this fixed channel membership reproduces the mixture directly (Figure~\ref{fig:c5-decomposition}): the suppressed set forms a unimodal population centered near $|w|\approx10^{-3}$ ($97$--$98\%$ of its weights lie below the valley) and carries $52$--$61\%$ of the total below-valley mass, while the remaining channels form the unimodal main body near $|w|\approx10^{-2}$; the residual below-valley mass comes from partially suppressed channels and the overlap tail of the main body. The secondary population is therefore not diffuse mass spread across the matrix --- its dominant carrier is this specific channel set.

\paragraph{Statistical characterization.} Hartigan's dip test separates an explicit secondary mode in 4 of the 9 flagged matrices (dip statistic $0.0035$--$0.0101$, $p \leq 10^{-5}$, surviving Bonferroni correction over the 15 tests at family-wise $\alpha = 0.05$); the six control matrices have dip statistics $0.0020$--$0.0022$ with nominal $p$ values $0.019$--$0.060$, none surviving the same correction; the remaining five flagged matrices show excess small-magnitude mass without a separated mode. A two-component Gaussian mixture in $\log_{10}|w|$ assigns the small-magnitude component a weight of $0.39$--$0.63$ in flagged matrices versus $0.27$--$0.28$ in controls.

\paragraph{Artifact exclusions.} The analysis operates on raw stored values: no mean subtraction (ruling out centering artifacts) and no re-quantization: the stored \texttt{bfloat16} values are losslessly represented in \texttt{float32} for computation. The low-magnitude population spans $1{,}949$--$3{,}141$ distinct \texttt{bfloat16} codes per matrix with no single code carrying more than $0.29\%$ of it --- a continuous population, not quantization pile-up. Exact zeros account for $\sim 3\times10^{-6}$ of entries in flagged and control matrices alike.

\paragraph{What the diagnostic chain delivers.} The chain runs: a cohort-level $k$ departure flags three entries (Section~\ref{sec:within-entry-qwen}); per-block disaggregation localizes the deviation to shallow blocks (Appendices~\ref{sec:appendix:qwen-diagnostic:detection}--\ref{sec:appendix:qwen-diagnostic:phenomenology}); channel-level analysis identifies the responsible structure --- a specific, enumerable set of intermediate channels suppressed coherently across all three FFN projections. Each step is verifiable from the released checkpoints alone, and no Weibull fitting enters the verification. This is the intended use of the framework, realized in this case: here the $k$ departure did not merely label a component as different --- it led to a concrete structural configuration that could be identified and confirmed at the weight level (related observations in Appendix~\ref{sec:appendix:qwen-diagnostic:future}). Whether other reference-band departures resolve into similarly enumerable structure is a question this single case cannot settle.

\end{document}